\documentclass[a4paper,11pt,twoside]{article}

\usepackage{listings}

\lstdefinestyle{mystyle}{
    language=Python,
    basicstyle=\ttfamily,
    keywordstyle=\color{blue},
    stringstyle=\color{red},
    commentstyle=\color{green},
    morecomment=[l][\color{magenta}]{\#}
}

\usepackage[ruled,vlined]{algorithm2e}

\usepackage{graphicx}
\usepackage{caption}
\usepackage{subcaption}
\usepackage{multirow}
\usepackage[utf8]{inputenc}
\usepackage[T1]{fontenc}
\usepackage{ dsfont }
\usepackage[english]{babel}

\usepackage{charter}

\usepackage{indentfirst}
\usepackage{xcolor}
\usepackage{verbatim}

\usepackage{hyperref}

\usepackage{pifont}

\usepackage[top=3.cm, bottom=4.0cm, left=2.2cm, right=2.2cm]{geometry}
\usepackage{amsmath}
\usepackage{amsthm}	
\usepackage{amsfonts}	
\usepackage{amssymb	
            ,bbm
            ,units 
            ,stmaryrd
           }
\usepackage{enumerate}

\usepackage[square,numbers,sort&compress]{natbib}
\SetSymbolFont{stmry}{bold}{U}{stmry}{b}{n}

\usepackage{
            ulem		
           ,soul		
} 

\numberwithin{equation}{section}
\numberwithin{figure}{section}
 \usepackage[nodayofweek]{datetime}

\renewcommand*{\thefootnote}{\fnsymbol{footnote}}

\title{  A Mean Field Ansatz for Zero-Shot Weight Transfer  }

\author{
\normalsize Xingyuan Chen\textit{$^{a}$} \\
        \small  Chenxingyuan4@huawei.com   
\and
\normalsize Wenwei Kuang\textit{$^{a}$} \\
        \small    Kuang.wenwei@huawei.com
\and
\normalsize Lei Deng\textit{$^{a}$} \\
        \small    Deng.lei2@huawei.com
\and
\normalsize Wei Han\textit{$^{a}$} \\
        \small    Harvey.hanwei@huawei.com
\and
\normalsize Bo Bai\textit{$^{a}$} \\
        \small   Baibo8@huawei.com
\and
 \normalsize Gon\c calo dos Reis\textit{$^{b,}$}\footnote{G.d.R.~acknowledges support from the FCT – Fundação para a Ciência e a Tecnologia, I.P., under the scope of the projects UIDB/00297/2020 (https://doi.org/10.54499/UIDB/00297/2020) and UIDP/00297/2020 (https://doi.org/10.54499/UIDP/00297/2020) (Center for Mathematics and Applications, NOVA Math).} \\
        \small  G.dosReis@ed.ac.uk
}

\date{%
    \footnotesize 
    $^{a}$~Theory Lab, 2012 Labs, Huawei Technologies Co., Ltd.
    \\
    $^{b}$~School of Mathematics, University of Edinburgh, The King's Buildings, Edinburgh, UK
    \\
    \longdate \today  
    \vspace{-1.0cm}
}

\usepackage{wrapfig}
\usepackage{graphicx}
\usepackage{epsfig,epstopdf}
\usepackage{tikz,pgfplots}
\usepackage{epsfig}
\tikzset{
-> 
}

\usepackage{algpseudocode}

\usepackage{amssymb}
\usepackage{pifont}
\usepackage{tikz}
\usetikzlibrary{automata, positioning, arrows}  

\theoremstyle{plain}
\newtheorem{theorem}{Theorem}[section]

\newtheorem{definition}[theorem]{Definition}

\newtheorem{example}[theorem]{Example}
\newtheorem{ansatz}[theorem]{Ansatz}

\newcommand{\bE}{\mathbb{E}}
\newcommand{\bF}{\mathbb{F}}

\newcommand{\bN}{\mathbb{N}}
\newcommand{\bP}{\mathbb{P}}

\newcommand{\bR}{\mathbb{R}}

\newcommand{\cC}{\mathcal{C}}

\newcommand{\cI}{\mathcal{I}}

\newcommand{\cL}{\mathcal{L}}

\newcommand{\cN}{\mathcal{N}}
\newcommand{\cO}{\mathcal{O}}
\newcommand{\cP}{\mathcal{P}}

\newcommand{\cR}{\mathcal{R}}

\newcommand{\cU}{\mathcal{U}}

\definecolor{darkgreen}{rgb}{0,0.35,0}

\newcommand{\1}{\mathbbm{1}}

\newcommand{\dd}{\mathrm{d}}

\newcommand{\lf}{ \overline{f} }

\newcommand{\lbtheta}{ \overline{\boldsymbol\theta} }

\newcommand{\bx}{ \mathbf{x}} 
\newcommand{\by}{ \mathbf{y}}

\newcommand{\btheta}{\boldsymbol\theta} 

\begin{document}

\selectlanguage{english}

\maketitle
\renewcommand*{\thefootnote}{\arabic{footnote}}


\begin{abstract}

The pre-training cost of large language models (LLMs) is prohibitive. One cutting-edge approach to reduce the cost is zero-shot weight transfer, also known as model growth for some cases, which magically transfers the weights trained in a small model to a large model. However, there are still some theoretical mysteries behind the weight transfer. In this paper, inspired by prior applications of mean field theory to neural network dynamics, we introduce a mean field ansatz to provide a theoretical explanation for weight transfer. Specifically, we propose the row-column (RC) ansatz under the mean field point of view, which describes the measure structure of the weights in the neural network (NN)   and admits a close measure dynamic. Thus, the weights of different sizes NN admit a common distribution under proper assumptions, and weight transfer methods can be viewed as sampling methods. We empirically validate the RC ansatz by exploring simple MLP examples and LLMs such as GPT-3 and Llama-3.1. We show  the mean-field point of view is adequate under suitable assumptions which can provide theoretical support for zero-shot weight transfer.

\end{abstract}
\tableofcontents

\section{Introduction}

Since the emergence of the scaling laws \cite{kaplanScalingLawsNeural2020}, the pre-training cost for Large Language Model (LLM) is exponentially increasing,  from training the Roberta \cite{liuRoBERTaRobustlyOptimized2019}  with 1024 V100 GPUs for 1 day, to training the PanGu-$\Sigma$ \cite{ren2023pangu} with 512 Ascend 910 GPUs over 100 days. 
Much research has focused on reducing the high training costs, such as exploring optimal scaling laws in \cite{deepmind}. In this work, we consider the model growth approach \cite{fujie2024}, which involves transferring weights from smaller pre-trained models to larger models, enabling the larger model to skip the early training stage and achieve similar performance. However, the theory behind it needs more clarification. We aim to address the mean field point of view to theoretically support the model growth and other similar weight transfer methods by viewing the training of a NN as samplings, which may lead to new insights.

This work mainly focuses on the following 2 points:
\begin{enumerate}
    \item We provide the row and column (RC) ansatz to support the mean field point of view for different architectures NNs and develop the corresponding distribution structure. 
    \item Under the mean-field point of view, the weights in a NN shall admit some joint distribution under proper assumptions. Thus, the weight transfer methods (model growth or pruning) are equivalent to sampling methods from the corresponding empirical distribution.
\end{enumerate}

Before introducing the mean-field point of view and the idea of samplings, we first provide the following simple 3-layer fully connected NN to clarify different NN settings:

\begin{example}[Simple 3-layer NN]
  \label{example: differnt setups}
  Following   the standard parametrizations (SP) and Maximal Update parametrizations ($\mu$P) in \cite{tp4}, with the mean-field  parametrizations (MFP) in \cite{mf-2022-path2,nguyen2023rigorous}, we consider the simple 3-layer NN $f:\bR \rightarrow \bR$  with width $N$ without bias:  
  
  \begin{align}
      \label{eq: 3lnn - sp}
      \textit{SP:}\ f (x)&= \frac{1}{\sqrt{N}} \sum_{ {\gamma_2}=1}^{N}
         v_{{\gamma_2}} \psi  \Big(    
          \frac{1}{\sqrt{N}} \sum_{ {\gamma_1}=1}^{N}    w^{(1)} _{{\gamma_2},{\gamma_1}} \psi
          \big(   {u}_{\gamma_1} x  \big)
          \Big), 
          \\
          \label{eq: 3lnn - mup}
     \textit{$\mu$P:}\  f (x)&= \frac{1}{{N}} \sum_{ {\gamma_2}=1}^{N}
         v_{{\gamma_2}} \psi  \Big(    
          \frac{1}{\sqrt{N}} \sum_{ {\gamma_1}=1}^{N}    w^{(1)} _{{\gamma_2},{\gamma_1}} \psi
          \big(   {u}_{\gamma_1} x  \big)
          \Big),
          \\
          \label{eq: 3lnn - mf}
      \textit{MFP:} \ f (x)&= \frac{1}{{N}} \sum_{ {\gamma_2}=1}^{N}
         v_{{\gamma_2}} \psi  \Big(    
          \frac{1}{{N}} \sum_{ {\gamma_1}=1}^{N}    w^{(1)} _{{\gamma_2},{\gamma_1}} \psi
          \big(   {u}_{\gamma_1} x  \big)
          \Big), 
  \end{align}
  where   the weights $u,v \in \bR^N,~w^{(1)}\in \bR^{N \times N}$ are independent of $N$, $\psi :\bR \rightarrow \bR$ is the  activation function, the learning rate may connected with $N$ for different parametrizations to match some   requirements in \cite{tp4,nguyen2023rigorous}.     
\end{example}

One of the main differences in Example \ref{example: differnt setups} is on the scalar, where it takes $1/\sqrt{N}$ or $1/N$, which leads to different dynamic at limitation as $N\rightarrow \infty$. Roughly speaking, assuming the weights are identical, the weights with scalar $1/\sqrt{N}$ can only admit a zero-mean distribution and can only change negligibly which leads to kernel behaviour (e.g. neural tangent kernel). In contrast, the weights with scalar $1/N$ is able to admit a non-zero mean distribution which can change massively and lead to measure movement. It is known that kernel behaviour may not be desired in contrast to the feature learning \cite{tp4}, thus we focus on the mean filed neural network (MFNN) using MFP in this work. 
 
The mean filed point of view for the dynamic of NN is first introduced in \cite{2018-mf-1}, we proposed the row and column (RC) ansatz provide further support to the mean filed view for modern NN architecture,   where the weight transfer is natural from this view. We show a simple 2-layer example to demonstrate the weight transfer methods as sampling methods.

\begin{example}[ 2-layer MFNN from  \cite{2019-mf-lln}  ]
\label{example: 2-layer}
For a MFNN $f:\bR \rightarrow \bR$ with an embedding layer and an output layer without bias,  we have  
    \begin{align}
        \label{eq: 2-layer NN t}
        f(x,\btheta^N_t) =\frac{1}{N} \sum_{i=1}^N \lf (x,\lbtheta_{t,i}) =\frac{1}{N} \sum_{i=1}^N v_i \psi ({u}_i,x ) ,
    \end{align}
    where $\bR^{2N }\ni \btheta^N_t:=(\btheta_{t,1},\dots,\btheta_{t,N})$ is the trainable weights at step $t$, with $\bR^{2}\ni \btheta_i=( {u}_i,v_i)$ and $\psi:\bR\rightarrow\bR$ is an activation function.  
    The corresponding empirical measure of the weights admit  the following form  
    \begin{align}
        \label{eq: intro 2l emprical measure}
        \mu^{N}_t := \frac{1}{N} \sum_{i=1}^N \delta_{ v_{t,i},      {u}_{t,i} }.
    \end{align}
    Under some   proper assumptions, \cite{2019-mf-lln} shows that the empirical measure $\mu^{N}_t$  converges to some limit measure $\mu_t^*$ in probability as $N\rightarrow \infty$.  
\end{example}

  The 2-layer NN \eqref{eq: 2-layer NN t} works  similar to the Monte Carlo estimation, where $\{v_t,u_t\}_{i}$ admit some joint distribution. The NN function $f$ thus acts like a test function applying to the sample $ \btheta_t^N$.  From this point of view,  for a  given trained NN with width $N$, one way to make a new NN with  $\tilde{N}$ is to generate $\tilde{N}$ samples from the empirical measure $\mu^{N}_t$  in \eqref{eq: intro 2l emprical measure}, see Example \ref{example: wt}  and Section \ref{sec: more on wt} for more detailed discussions.

The weight transfer is equivalent to samplings under the mean filed point of view for the MFNN, however, there is no guarantee that the MFNN using MFP admits this view, thus, we develop the RC ansatz to theoretically support this view.   By analysing LLMs such as Llama-3.1 in the experiment section, though the neural tangent kernel (NTK) introduced in \cite{2018ntkorigin} is closer to SP in \eqref{eq: 3lnn - mf}, it is still adequate to consider the mean field view for LLM using SP or $\mu$P  at initialisation.


We summarize, the  main contributions are as follows:

\begin{enumerate}
    
    \item \textbf{Numerical validation of the feasibility for the mean-field parametrizations}: Using the mean-field parametrizations(MFP), we train MLP examples and GPT-3, which work properly.  The MFP setting is adequate for NN and provides a higher capacity for the initialisation and training process of NN. 
    
    \item \textbf{The RC ansatz for the mean filed neural network}: We proposed the row and column (RC) ansatz to depict the measure structure for the MFNN that can deal with different NN architectures and admit close measure dynamics under proper assumptions.
    
    \item \textbf{The weight transfer for MFNN and LLMs}:  We provide theoretical support for the weight transfer of the MFNN, the numerical results not only validate the RC ansatz and weight transfer for the MFNN but also support a similar mechanism for the trained LLMs. 
\end{enumerate}

We refer to the Appendix \ref{sec: appendix} for  a detailed discussions such as the existence and uniqueness of the limit measure.

\section{Related work}
\subsection{Model growth and Model pruning}

Model growth corresponds to some upsampling methods from the mean-field point of view, which reduce the training cost of training a LLM as verified in \cite{fujie2024}. The \cite{net2net} is a starting work to study model growth in the deep NN, \cite{bert2bert} study a BERT-based model growth method. The FLM-101B \cite{li2023flm} is trained via the model growth method introduced in \cite{yao2024masked} which saves a huge training cost. We refer to \cite{fujie2024} for more detailed references and discussions.

Similarly, model pruning corresponds to some downsampling method from the mean-field point of view, which reduces the size and complexity of the NN by removing unnecessary parameters to improve efficiency. We refer \cite{mp1,mp2,mp3,mp4,mp5,mp6}  for more detailed references and discussions.

\subsection{Mean filed Neural Network} 
The mean filed point of view for the 2-layer NN is introduced in \cite{2018-mf-1,2019-mf-lln}, where the measure of the weights admits the Fokker-Plank equation with gradient flows in measure space. For multi-layer MFNN, \cite{mf-2022-path2} introduces the difficulty of developing a proper measure structure and develops an approach by sending the width of the weights into infinity step by step. \cite{mf-2019-path1} introduce an approach by fixing the first and the last layer, they show nice ideas that the weights are path-connected and the corresponding "ideal particles" admit independent measure structure. \cite{nguyen2023rigorous} is the first work to
obtain global convergence guarantees for the multi-layer MFNN with rigorous mathematical proof. By introducing the "neuronal embedding", they can nicely deal with MFNN using stochastic gradient descent (SGD) to admit a close dynamic. See more detailed discussion in \cite[Section 9]{nguyen2023rigorous}. In this paper, the RC ansatz has a similar measure structure to  \cite{nguyen2023rigorous} but with higher flexibility by introducing more random variables, we provide more discussions in the following text.

\section{Methodology}
\subsection{ The RC Ansatz }
\label{sec: the rc ansatz}
To introduce the RC ansatz precisely, we first provide the following auxiliary definitions to depict the NN: the  $\gamma$ notation for the lower indices in the NN computation and the $\Gamma$ set for the measure structure in the RC ansatz.

\begin{definition}[$\gamma$ notation]
    \label{def: gamma set}
    For a given NN  with  $L_w$ matrices $\{W^{(\ell)}\}_{\ell =1}^{L_w},~W^{(\ell)}\in \bR^{D^{\ell,1} \times D^{\ell,2}}$, and  $L_b$ vectors $\{B^{(\ell)}\}_{\ell =1}^{L_b}$, $B^{(\ell)}\in \bR^{D^{\ell,1} }$, with the embedding matrix $U\in \bR^{D^{u,1} \times D^{u,2}}$, the output matrix $V\in \bR^{D^{v,1} \times D^{v,2}}$ with  biases $B^{(U)} \in \bR^{ D^{u,1} } $ and $B^{(V)} \in \bR^{ D^{v,2} } $ and activation functions $\{ \psi \}_{i=1}^{L_\psi}$, where $L_w,L_b,  \{D^{\ell,1},D^{\ell,2}\}_{\ell = 1}^{L_w}, D^{u,1} , D^{u,2}, D^{v,1} , D^{v,2},L_\psi$ are positive integers.    
    The $\gamma:=\{\gamma_1,\dots\}$ in the lower indices of the matrices and vectors in the computation servers for the computation in the NN, we use $|\cdot |$ to denote the length of $\gamma$.  
\end{definition}
Notice that for the case $D^{u,2}=1$ ($D^{v,1}=1$) corresponding to the 1-dimension input (output), we let $U$($V$) be a vector for simplicity. We provide the 4-layer NN with skip connection in  Example \ref{example: gamma set}  to clarify the $\gamma$ notation, see the Appendix \ref{sec: appendix} for more examples.

The $\gamma$ notation depicts the computation order in a given NN and is crucial to developing the corresponding measure structure. Recall in  Example \ref{example: 2-layer}, there is only a single summation in Equation \ref{eq: 2-layer NN t} (the $i$ is equivalent to $\gamma_1$),  so that the empirical measure has only $N$ samples instead of $N^2$ samples. We need to consider $\{u,v\}_i$ in pair, which shows a glance at the importance of a proper measure structure,  see more detailed discussions in  Section \ref{section: 2layers}.

Now, to properly construct the measure structure, consider the $\gamma$ notation and the lower indices position for all the weights, we group the $\gamma=\{\gamma_1,\gamma_2,\dots\}$ into different sets $\Gamma_i$ as follow.   
\begin{definition}[$\Gamma$ set]
    \label{def: gamma}
    Under the settings in Definition \ref{def: gamma}, consider the lower indices position of all the weights $ \{W^{(\ell)}\}_{\ell =1}^{L_w},\{B^{(\ell)}\}_{\ell =1}^{L_b},U,V,B^{(U)},B^{(V)}$, we can then split the $\gamma=\{\gamma_1,\gamma_2,\dots\}$ into different disjoint sets of indices in $\Gamma:=\{\Gamma_1,\dots \}$,  such that for any lower index position of a given weight, there exist a set $\Gamma_i \in \Gamma$  contains all the indices showing up at that position in the computation of the NN using the $\gamma$ notation. We let such $\Gamma$ set of sets with minimal elements to be the $\Gamma$ set of the NN.
\end{definition}
Notice that the $\Gamma$ set of a NN is a set of indices sets, and each element contains a few distinct elements in $\gamma$. The notations and definition above may be mathematically non-rigorous, where we try to avoid too complicated notations but just to show the main idea instead. We provide the following example for clarification.

\begin{example}[$\gamma$ notation and $\Gamma$ set]  
  \label{example: gamma set} 
  We  consider the 4-layer NN $f:\bR\rightarrow \bR$ with width $N$  and skip connection architecture to demonstrate the key idea of the   $\Gamma$ set where  we take $\phi:\bR^N\rightarrow\bR^N$ as an activation function 
  \begin{align}
   \nonumber 
      f(x)  =& 
         \frac{1}{N}\sum_{ {\gamma_3}=1}^{N}    V_{{\gamma_3}} \psi  
         \bigg(   \frac{1}{N}\sum_{ {\gamma_2}=1}^{N}  W^{(2)} _{{\gamma_3},{\gamma_2}}
          \psi  \Big(   \frac{1}{N} 
       \sum_{ {\gamma_1}=1}^{N}    W^{(1)} _{{\gamma_2},{\gamma_1}} 
        \cdot 
       \psi
      \big(   U_{ {\gamma_1}} x  +  B^{(U)} _{ {\gamma_1}} \big) + B^{(1)} _{ {\gamma_2}}
      \Big) 
       \\
      \label{eq: mlp nn function - 4-layer 1d }
      &\  
      + B^{(2)} _{ {\gamma_2}} + \frac{1}{N}
       \sum_{ {\gamma_1}=1}^{N}     
     \cdot 
       W^{(1)} _{{\gamma_3},{\gamma_1}} \psi
      \big(   U_{ {\gamma_1}} x  +  B^{(U)} _{ {\gamma_1}} \big) + B^{(1)} _{ {\gamma_3}}
      \bigg) 
      +  B^{(V)} _{ {\gamma_4}},
\end{align}
with   $\gamma_4 = 1$, $D^{u,1}=N,~D^{u,2}=1,~D^{1,1}=D^{1,2}=D^{2,1}=D^{2,2}=N,~D^{v,1}=1,~D^{v,2}=N$ and $N\in \bN$.
We have the $\gamma$ set of 4 elements: $\gamma=\{\gamma_1,\gamma_2,\gamma_3,\gamma_4\}$, $|\gamma|=4$, and the $\Gamma$ set with 3 sets: $\Gamma=\big\{   \{\gamma_1\},  \{\gamma_2,\gamma_3\}, \{\gamma_4\}    \big\}$, $|\Gamma|=3$, where that the row index of $W^{(1)}$ has   $\gamma_2$ and $\gamma_3$ and leads to the set $\{\gamma_2,\gamma_3\}$ in $\Gamma$.   One can also write in Equation \ref{eq: mlp nn function - 4-layer 1d } that the last summation over $\gamma_5$ instead of $\gamma_1$ again by introducing an extra index, but we suggest not adding an index unless necessary for clarity.  
\end{example}

The skip connection in Example \ref{example: gamma set} demonstrates the development of the $\Gamma$ set, the $\Gamma$ set is to separate the dependency across different weights. Applying the first-order gradient descent based training method, the updates of   $V$ shall be dependent on $W^{(2)}, W^{(1)}$, and the updates of $W^{(2)}W^{(1)}$ shall dependent of $V$, we thus need to consider them together via $\Gamma$, see Section \ref{section: other nn structure} for more detailed discussions. Now, based on the definitions and notations above, we can introduce the row and column (RC) ansatz

\begin{ansatz}[RC ansatz for the MFNN]
\label{ansatz: rc ans}
For a MFNN under the notations in Definition \ref{def: gamma}, assume further that there exist a set of functions $\{ \phi^{(\ell)} \}_{\ell =1}^{L_w}$ such that the weight $w_{i,j}^{(\ell)}$  in $W^{(\ell)}$ with $i\in\{1,\dots,D^{\ell,1}  \},~j\in\{1,\dots,D^{\ell,2}  \}, \ell \in \{1,\dots,L_w\}$ admit the form $w_{i,j}^{(\ell)}=\phi^{(\ell)}(\cR^{(\ell)}_i,\cC^{(\ell)}_j)$, for some vectors $     \cR^{(\ell)}=\{\cR^{(\ell)}_1,\dots, \cR^{(\ell)}_{D^{\ell,1}}\} \in \bR^{D^{\ell,1}}$ and  $   \cC^{(\ell)}=\{\cC^{(\ell)}_1,\dots, \cC^{(\ell)}_{D^{\ell,2}}\}\in \bR^{D^{\ell,2}}$. Then  the weights at initialisation  and at any training step $T$ using  the first-order gradient based method can be depicted by an empirical measure $\mu_t^{\cR\cC,N} \in \cP(\bR^{  D^{v,1}+ 3 L_w + D^{u,2}+2}),~ t \in [0,T]$. The empirical measure $\mu_t^{\cR\cC,N}$ converge weakly to some limit measure $\mu_t^{\cR\cC}$ as $N\rightarrow \infty$. 

Further, the $\mu_t^{\cR\cC}$ admits a product form $\mu_t^{\cR\cC} = \prod_{i=1}^{|\Gamma|} \mu_{i,t}^{\cR\cC}$ where $\mu_{i,t}^{\cR\cC} \in \cP(\bR^{ L_{i}^ {\cR\cC} })$ for some positive integers $\{ L_{i}^{\cR\cC} \}_{i=1}^{|\Gamma|}$   satisfy  $ \sum_iL_{i}^{\cR\cC} =  D^{v,1}+ 3 L_w + D^{u,2}+2$.  
\end{ansatz}

The RC ansatz provides a description of the measure structure for the corresponding NN, it describes a matrix-type weight with 2 random variables (RVs) instead of a single RV, where they correspond to some row and column features. The product form of the RC ansatz splits the whole measure into different parts based on $\Gamma$ set, this separation comes from the nature of the first-order gradient based method which enables us to apply weight transfer step by step in Algorithm \ref{alg: RC wt}. This measure structure also suggests dependency across different layers, roughly speaking, the adjacency weight matrices would be dependent in a way of rows followed by column as shown in the Experiment section.

The RC ansatz provides a closed measure dynamic for the problem discussed in \cite[Section 4.3]{mf-2022-path2}, we present a detailed analysis with the motivation of developing the RC ansatz in  Section \ref{sec: mlp ffn case} by going through a simplified 5-layer NN example.

Notice that the RC ansatz can properly adapt to the case when the weight matrix can be written in the form of $w_{i,j}^{(\ell)}=\phi^{(\ell)}(\cR^{(\ell)}_i,\cC^{(\ell)}_j)$ which correspond to the RC initialisation in Section \ref{sec: rc ini}.  For independent and identically distributed (i.i.d) initialisation of the weights matrix shall have $N^2$ degree of freedom and thus cannot be represented by $2N$ random variables in general, and the RC ansatz serves as a proper approximation. We show in the experiment section that this approximation is still good, the analysis in Section \ref{section: mlp 5-layer} for the update dynamics also supports the RC ansatz even for i.i.d initialisation. 

\subsection{ Framework for weight transfer}

Under the RC ansatz, which provides theoretical support for the mean filed point of view for the MFNN with different architectures. We now provide the following framework in Algorithm \ref{alg: RC wt} for the weights transfer, notice that weight transfer methods across different sizes NN correspond to up-sampling and down-sampling methods. Our framework and methods here are not new, but we seek to provide the mean field point of view for the existing methods which may support new methods. From the mean-field point of view, we want some target measure for the weights, the training processes of the corresponding NN are part of the many methods, and there exist more ways to approximate the target measure, for example, by taking advantage of the empirical measure update trajectory. See more detailed discussion in Section \ref{sec: more on wt}.   

\begin{algorithm}[htb!]
    \SetAlgoLined
    \KwIn{ 
          The weights set $ \btheta^N :=\big \{ \{W^{(\ell)}\}_{\ell =1}^{L_w} \cup \{B^{(\ell)}\}_{\ell =1}^{L_b},U,V,B^{(U)},B^{(V)} \big \}$ in the NN with corresponding  $\gamma$ notation and $\Gamma$ set, functions $ F^{\cR }, F^{\cC }$ to generate the row and column information, functions $ G^{\cR }, G^{\cC }$ to generate the new matrix based on row and column indices. 
          }
          \   
          \
          \For{weights in each $\Gamma$}{
              1.  Get the row or column charismatic via $ F^{\cR },~F^{\cC }$ for each weight matrix / column 
              \;
              2.  Generate the new index set $\cI_i \in \bR^{N_i}$, for  $i \in \{1,\dots,|\Gamma |\}$ based on previous step.
              \;
              3. Substitute  the corresponding weights in $ \btheta^N$ by sampling    on $\cI_i$ and $ G^{\cR }, G^{\cC }$. 
          
          }
          \KwOut{The transferred weights  set $ \btheta^N$.    }   
    \caption{ Weight transfer framework }
    \label{alg: RC wt}
  \end{algorithm}

  The key idea above is the measure structure $\mu_t^{\cR\cC} = \prod_{i=1}^{|\Gamma|} \mu_{i,t}^{\cR\cC}$ using the RC ansatz, where the weights connected to $\mu_{i,t}^{\cR\cC}$ need to be considered jointly via $\cI_i$, and the product form enable us to sampling step by step.  We provide a more detailed form in Algorithm \ref{alg: RC wt detialed}. We show the following example for weight transfer from width $N$ to $\tilde{N} \in \bN$:

  \begin{example}[Weight transfer for Example \ref{example: gamma set} ]  
      \label{example: wt} 
      To attend width $\tilde{N} $, using Algorithm \ref{alg: RC wt} with $ G^{\cR }, G^{\cC } $, we have   
  \begin{enumerate}
      \item the row of $U, B^{(U)}$ and the column of  $W^{(1)}$ are sampling via the index set $\cI_1 \in \bR^{ \tilde{N} }$;
      \item the row of $V, W^{(2)},W^{(1)},B^{(1)},B^{(2)}$ and the column of  $W^{(2)}$ are sampling via $\cI_2 \in \bR^{ \tilde{N} }$;
      \item the row of $ B^{(V)}$ are sampling via $\cI_3 = \{1\}$.  
  \end{enumerate}

  \end{example}
  
  Notice that if $ \tilde{N} = 2N$ in Example \ref{example: wt}, a naive way is just to double the corresponding rows and columns following the step above so that the new $W^{(1)}$ becomes a $2N \times 2N$ matrix, where we split it into 4 disjoint $N \times N$ block such that each block is the same as the origin $W^{(1)} \in \bR^{N \times N}$. Other weights can be transferred accordingly. Considering the scalar $1/\tilde{N}$, this naive method corresponds to the type (A) method in \cite[Section 3.1]{fujie2024} where the scalar change from $1/N$ to $1/2N$ and provides a $1/2$ scale parameter.  
\section{Experiments}

\subsection{MLP experiments on CIFRA-10}
We show simple MLP examples to give some important insights related to the RC ansatz and demonstrate weight transfer results. 
\begin{figure}[h!bt]
    \centering
    \begin{subfigure}{.33\columnwidth}
      \centering
      \includegraphics[width=\columnwidth,]{ 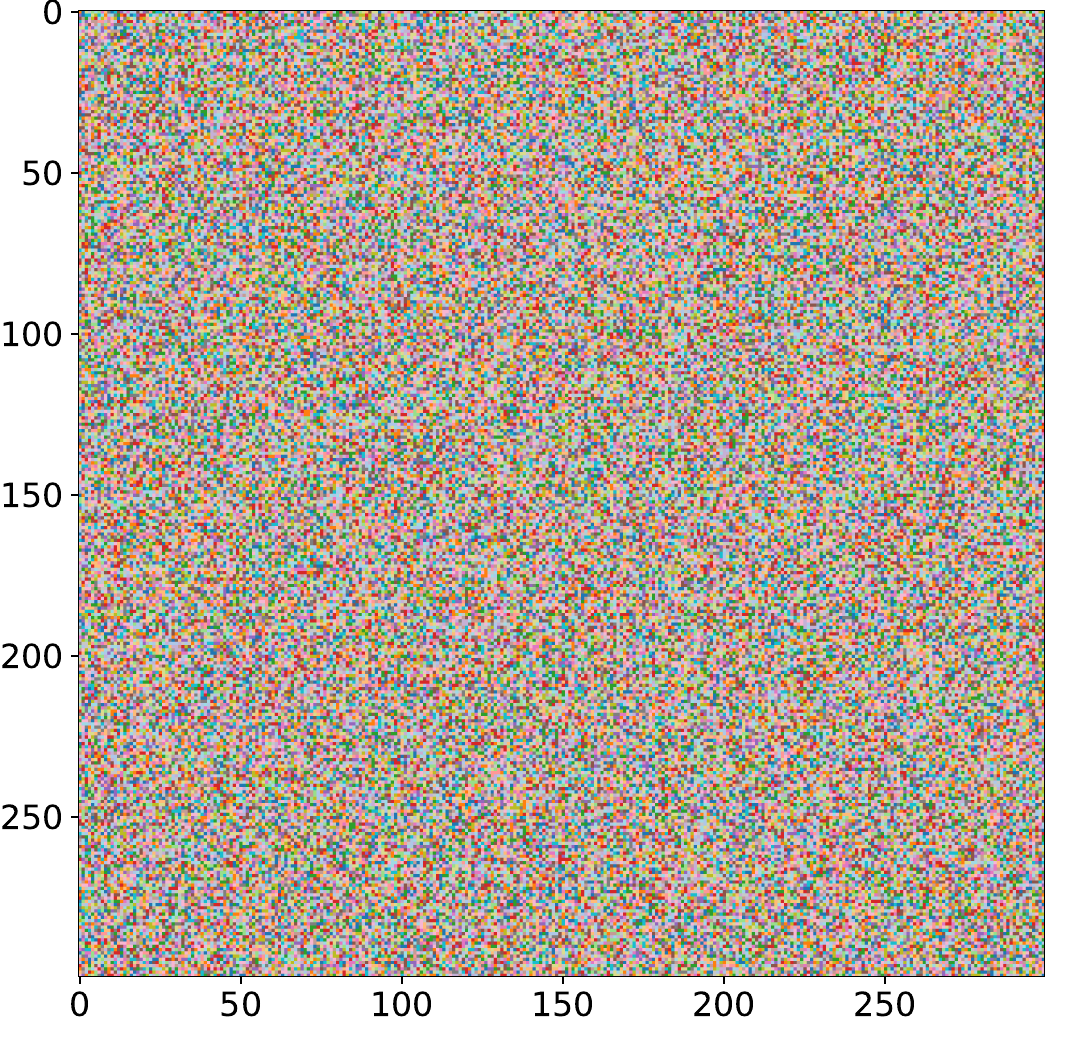}
      \caption{Init of MFP}
    \end{subfigure}%
  \begin{subfigure}{.33\columnwidth}
      \centering
      \includegraphics[width=\columnwidth,]{ 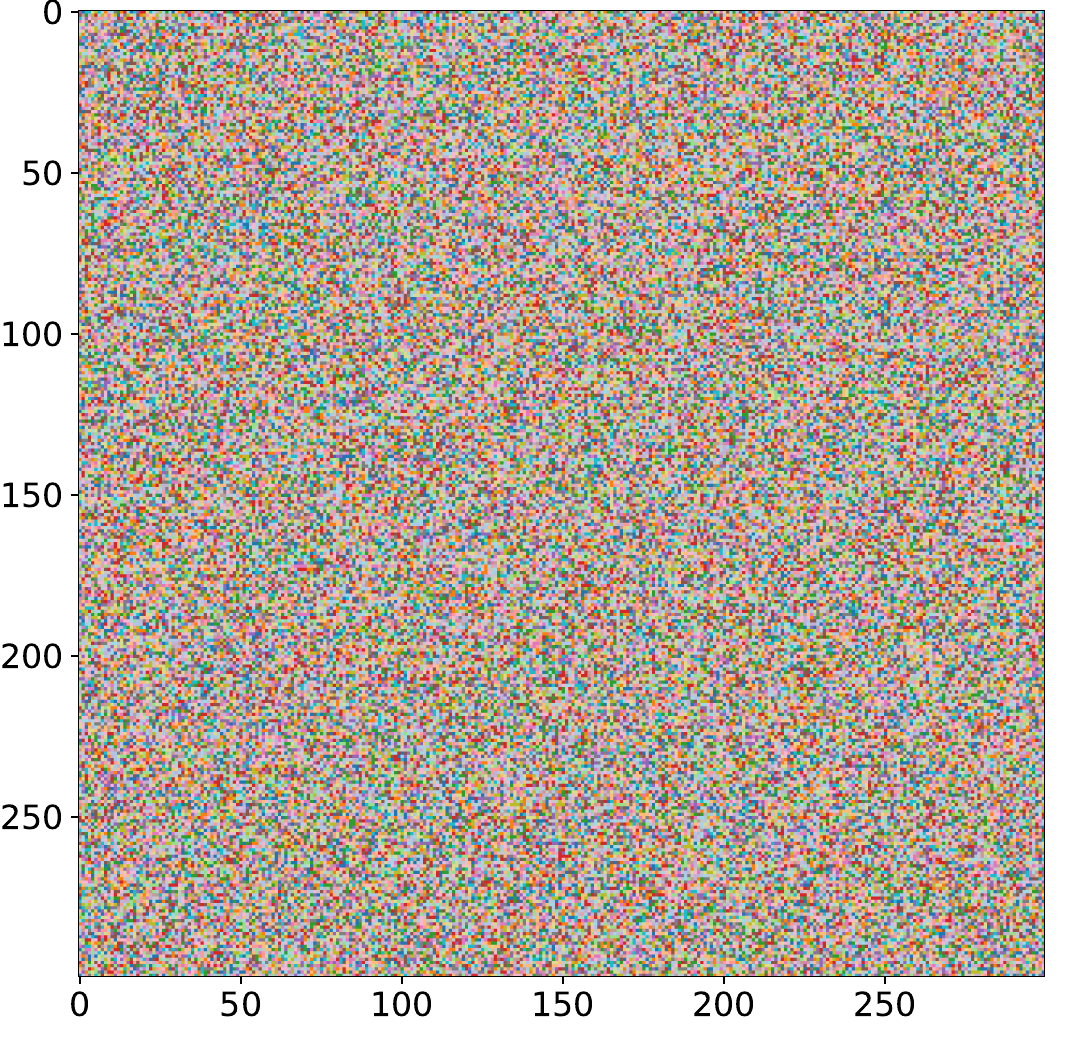}
      \caption{Init of SP }
    \end{subfigure}%
  \begin{subfigure}{.33\columnwidth}
      \centering
      \includegraphics[width=\columnwidth,]{ 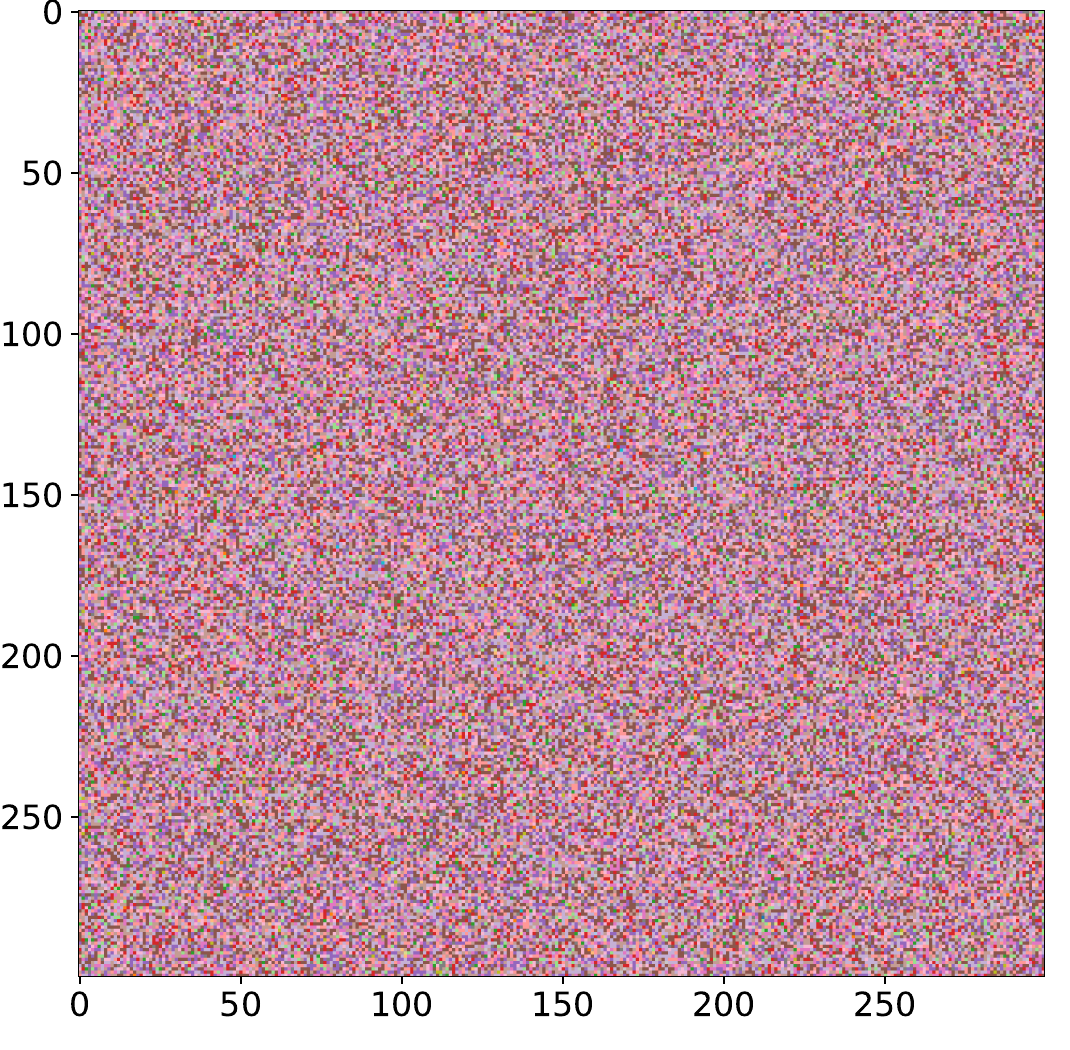}
      \caption{Init of $\mu P$}
    \end{subfigure}%
  \\
  \begin{subfigure}{.33\columnwidth}
      \centering
      \includegraphics[width=\columnwidth,]{ 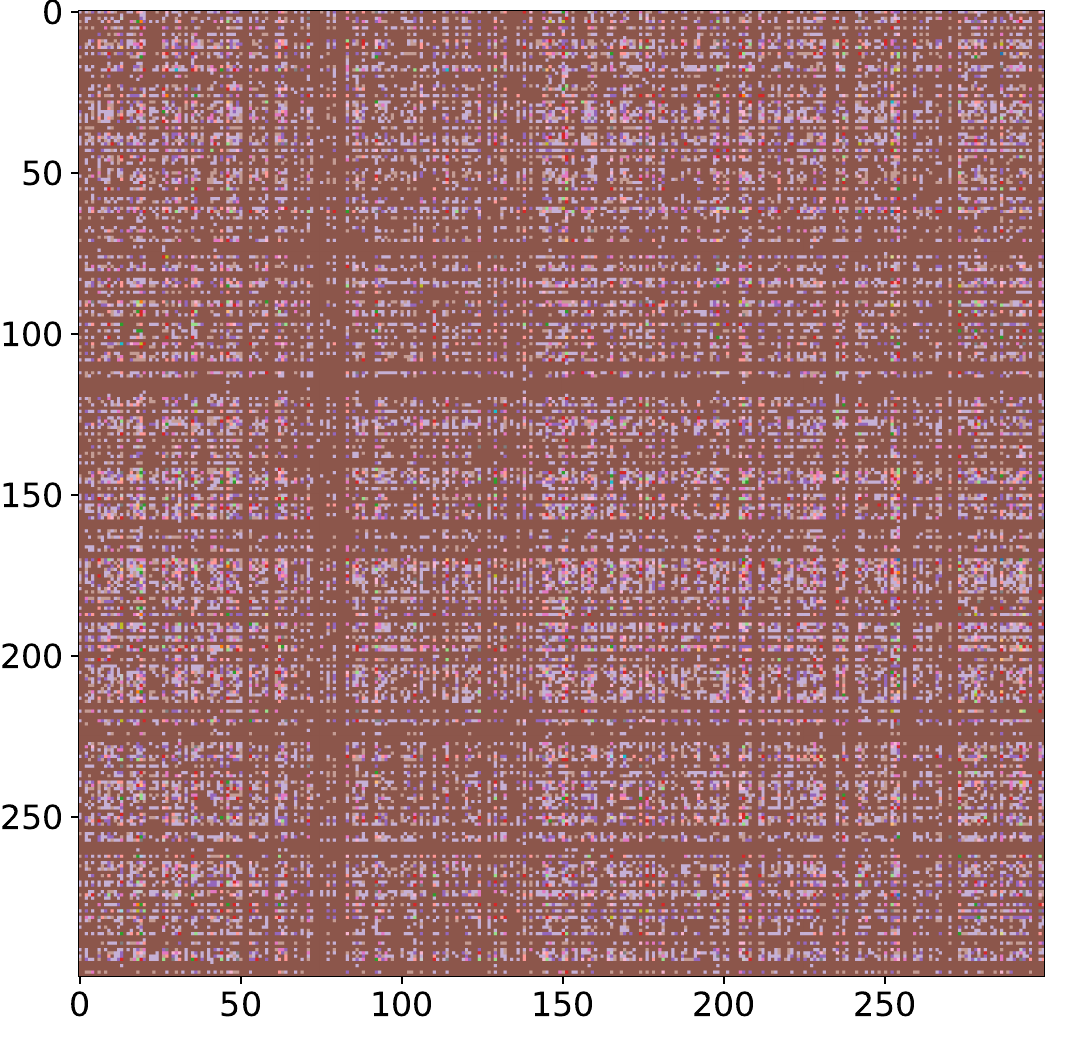}
      \caption{ Trained of MFP}
    \end{subfigure}%
  \begin{subfigure}{.33\columnwidth}
      \centering
      \includegraphics[width=\columnwidth,]{ 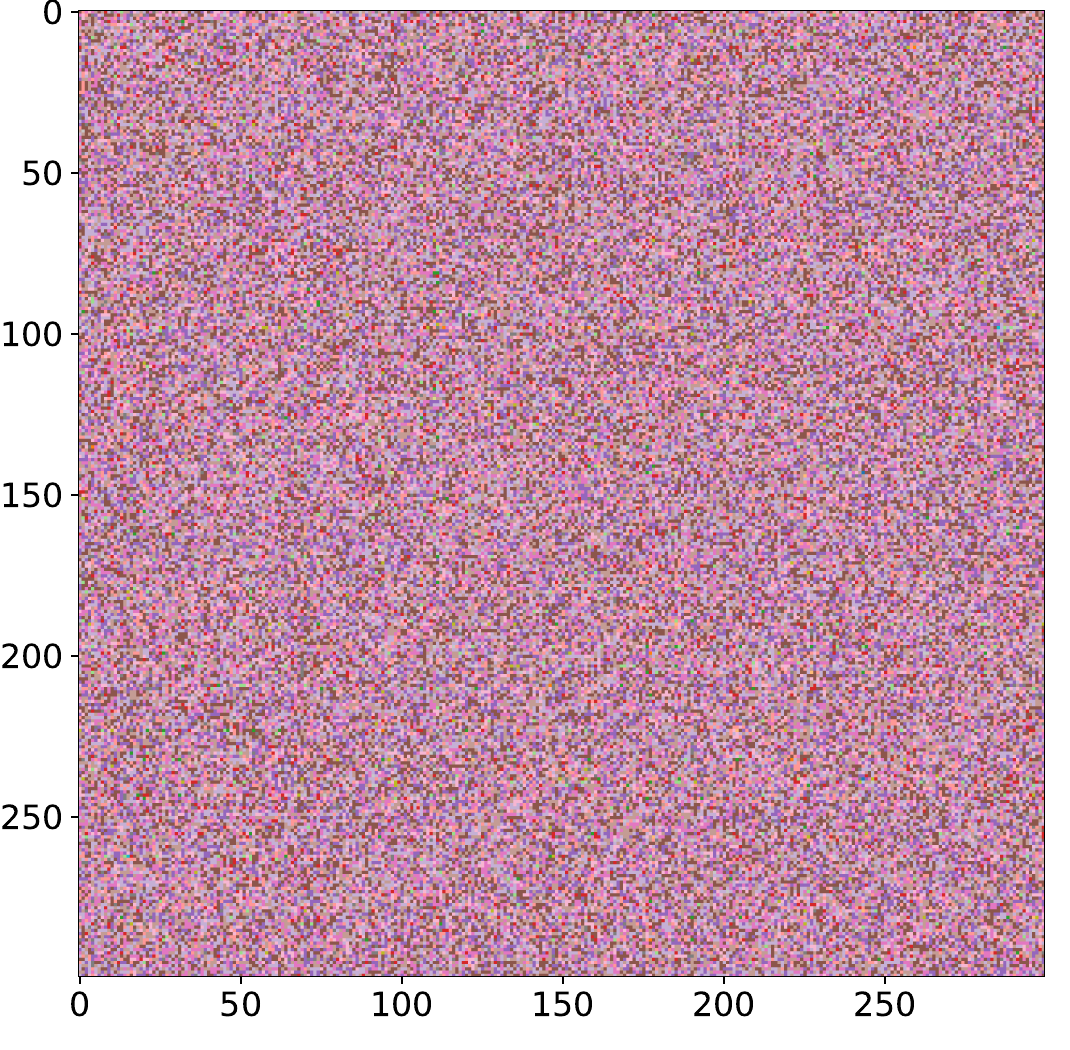}
      \caption{ Trained of  SP }
    \end{subfigure}%
  \begin{subfigure}{.33\columnwidth}
      \centering
      \includegraphics[width=\columnwidth,]{ 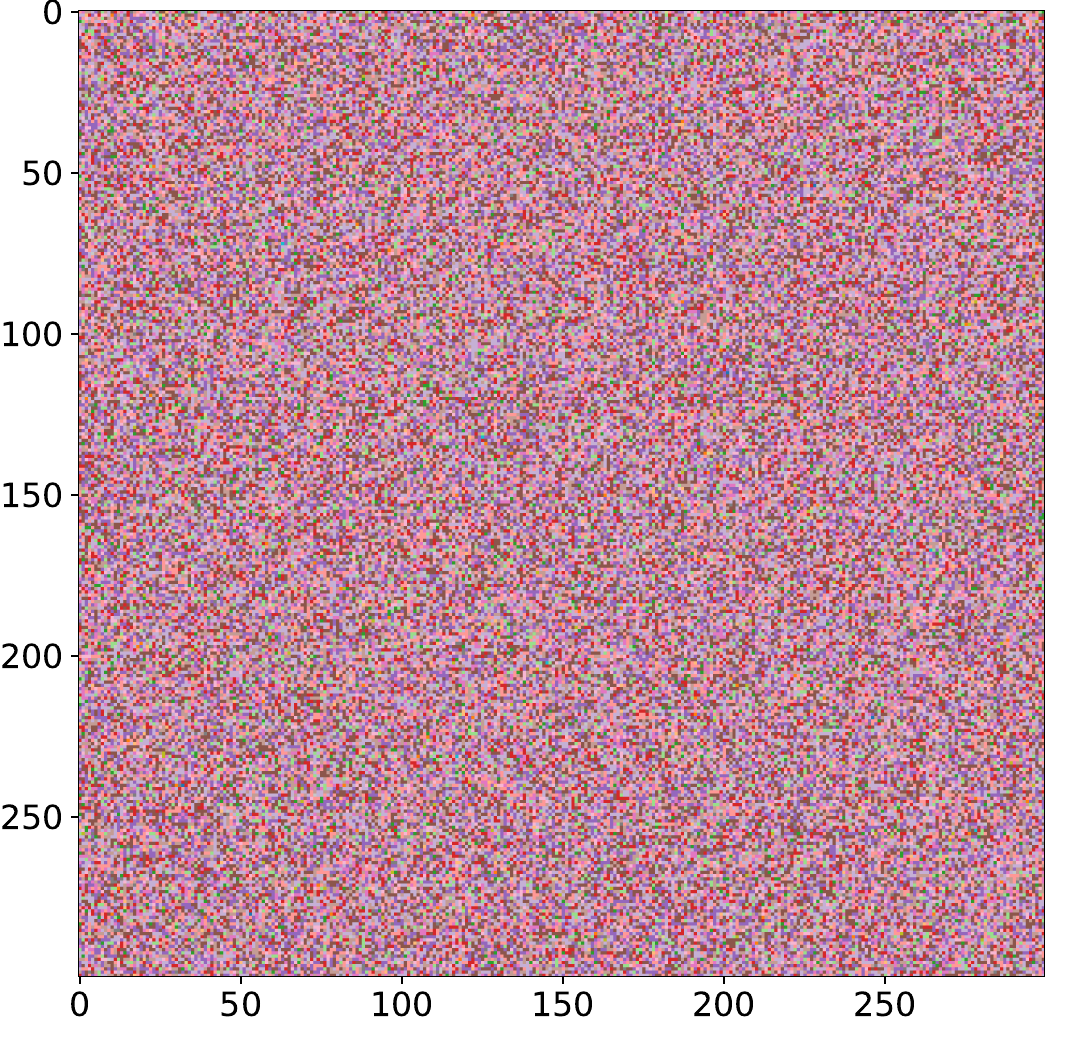}
      \caption{ Trained of $\mu P$ }
    \end{subfigure}%
    \caption{  Heat plots for the normalized value of the middle layer in a 3-layer NN with biases under 3 different setups with $N=300$, above are the results at initialization and below are the results after training for 10 epochs. Notice that at initialization, the SP and MF take a uniform distribution, and the $\mu$ P takes a Gaussian distribution.
 } 
    \label{fig:1: correlation}
\end{figure}
\subsubsection{For RC ansatz verification}
In Figure \ref{fig:1: correlation}, there is no particular pattern in $(a)-(c)$. After training for 10 epochs, where all 3 methods achieve above $45\%$ accuracy on the test dataset, there is a negligible pattern in $(e)$ and $(f)$ for SP and $\mu P$, but there is clear row and column correlation pattern in $(d)$ in the MFP which support the RC ansatz. 

The detailed reason is discussed in the Appendix, we provide the brief reason here: the one-step updates of the weights are row and column wise connected, however, for the middle layer in the 3-layer NN example, the one-step update of the weights have the same order as initialization (i.e $\cO(1)$ at initialization with each step update $\cO(1)$ ) under MFP, while the one-step update of the weights have a smaller rate compare to initialization (i.e $\cO(1)$ at initialization with update $\cO(1/\sqrt{N})$ ) for SP and $\mu P$.

In the later example,  we show that as the training steps $T$ increase (e.g.  $T\sim \cO(N)$), there exists a similar row and column dependency for SP and $\mu$P.

In Table \ref{table:1: correlation}, there is no dependence at initialisation across different layers, and after training for 10 epochs, the dependency between the M.column and the rows in the first layer increases under all 3 setups, where the MFP has the most significant increment. The dependency between M.col and the last layer also increases as expected.   

\begin{table}[h!bt]
    \centering 
    \resizebox{\columnwidth}{!}{%
    \begin{tabular}{|c|c|cc|cc|cc|}
    \hline
    \multirow{2}{*}{Step}    & Setup       & \multicolumn{2}{c|}{SP}                                       & \multicolumn{2}{c|}{$\mu$P}                                      & \multicolumn{2}{c|}{MFP}                                       \\ \cline{2-8} 
                             & Layer names & \multicolumn{1}{c|}{M.row} & M.column & \multicolumn{1}{c|}{M.row} & M.column & \multicolumn{1}{c|}{M.row} & M.column \\ \hline
    \multirow{2}{*}{Initial} & First layer & \multicolumn{1}{c|}{0.0204}            & 0.0333               & \multicolumn{1}{c|}{0.0189}            & 0.0202               & \multicolumn{1}{c|}{0.0042}            & 0.0020               \\ \cline{2-8} 
                             & Last layer  & \multicolumn{1}{c|}{0.0266}            & 0.0276               & \multicolumn{1}{c|}{0.0185}            & 0.0218               & \multicolumn{1}{c|}{0.0248}            & 0.0213              \\ \hline
    \multirow{2}{*}{Trained} & First layer & \multicolumn{1}{c|}{0.0192}            & \textbf{0.1002}               & \multicolumn{1}{c|}{0.0151}            & \textbf{0.0665}               & \multicolumn{1}{c|}{0.0215}            & \textbf{0.5274}                \\ \cline{2-8} 
                             & Last layer  & \multicolumn{1}{c|}{ \textbf{0.0768} }            & 0.0592               & \multicolumn{1}{c|}{\textbf{0.0358}}            & 0.0157               & \multicolumn{1}{c|}{ \textbf{0.2371 } }            & 0.02665               \\ \hline
    \end{tabular}
    }
     \caption{The absolute correlation coefficient between different layers of the 3-layer NN example under 3 different setups with $N=1000$. The first layer stands for the embedding matrix of size $N\times 3072$ and we take the mean value on columns, the M.row(column) stand for the column(row) mean value of the middle matrix of size $N\times N$, the last layer stands for the output matrix of size $10\times N$ where we take the summation on first 4 rows. Where all the weights are sampled i.i.d and each model is trained for 10 epochs.}
     \label{table:1: correlation}

\end{table}

  \begin{figure*}[h!bt]
    \centering
    \begin{subfigure}{.33\textwidth}
      \centering
      \includegraphics[width=\textwidth,]{ 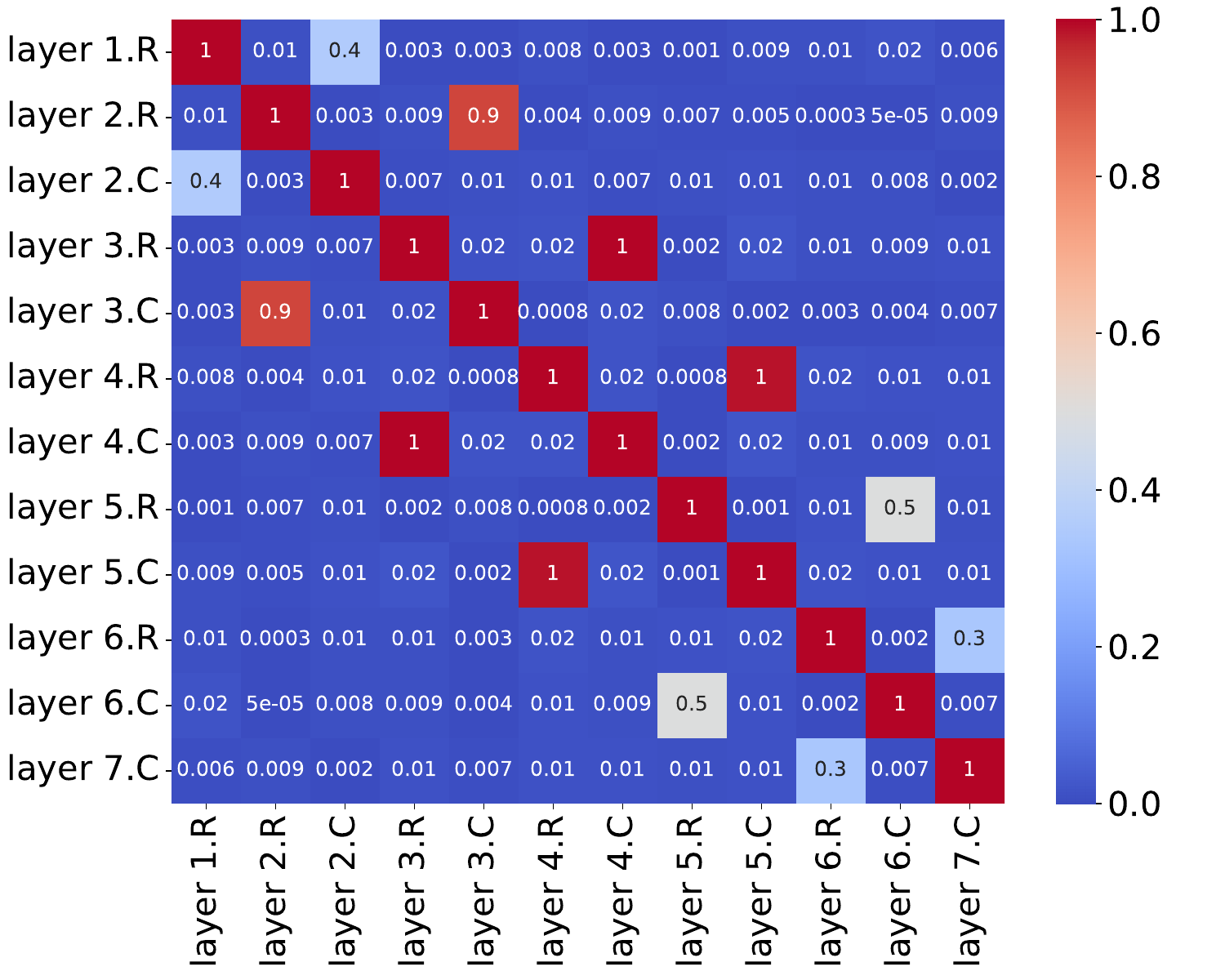}
      \caption{Non-constant trainable bias }
    \end{subfigure}%
  \begin{subfigure}{.33\textwidth}
      \centering
      \includegraphics[width=\textwidth,]{ 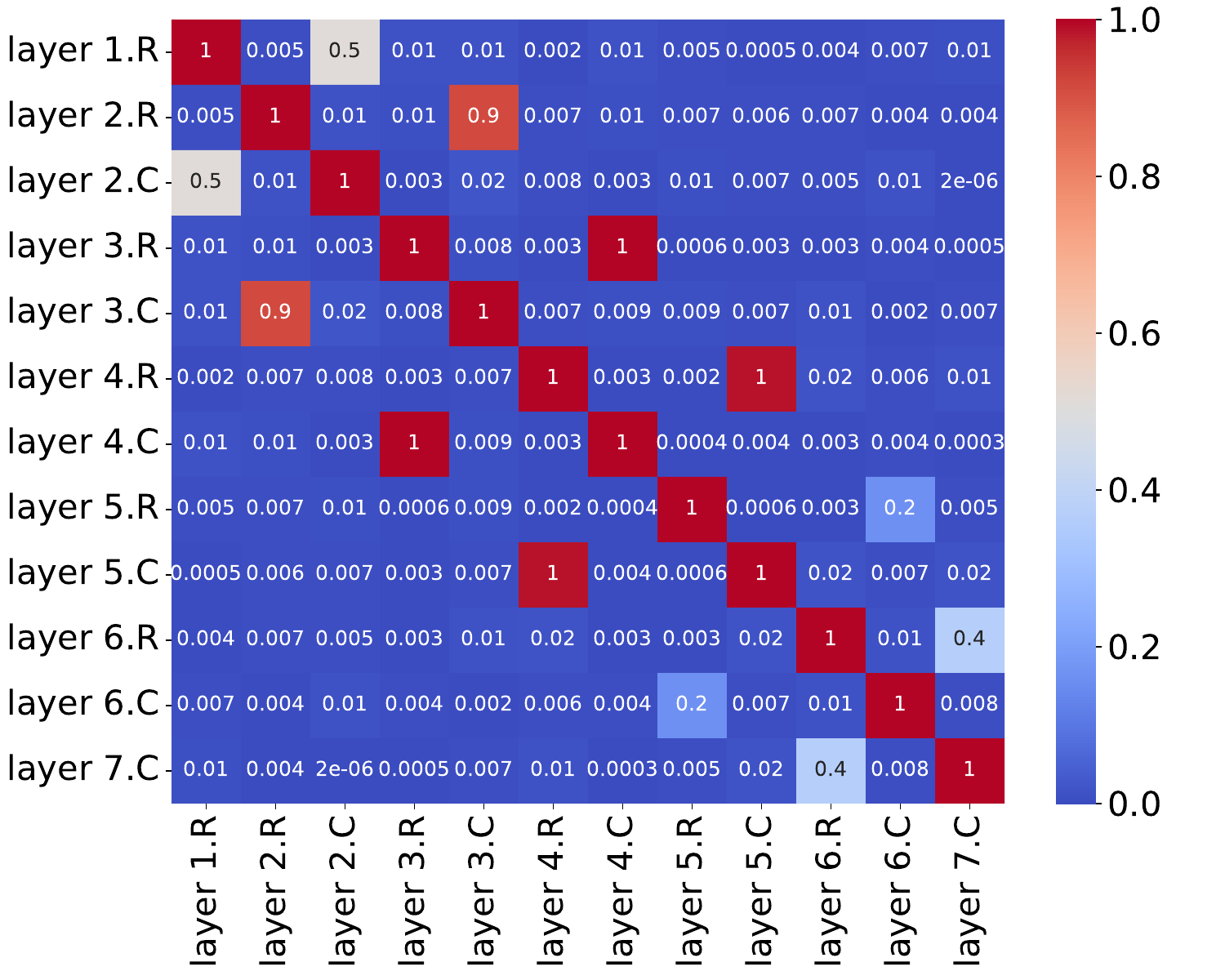}
      \caption{Constant trainable bias }
    \end{subfigure}%
  \begin{subfigure}{.33\textwidth}
      \centering
      \includegraphics[width=\textwidth,]{ 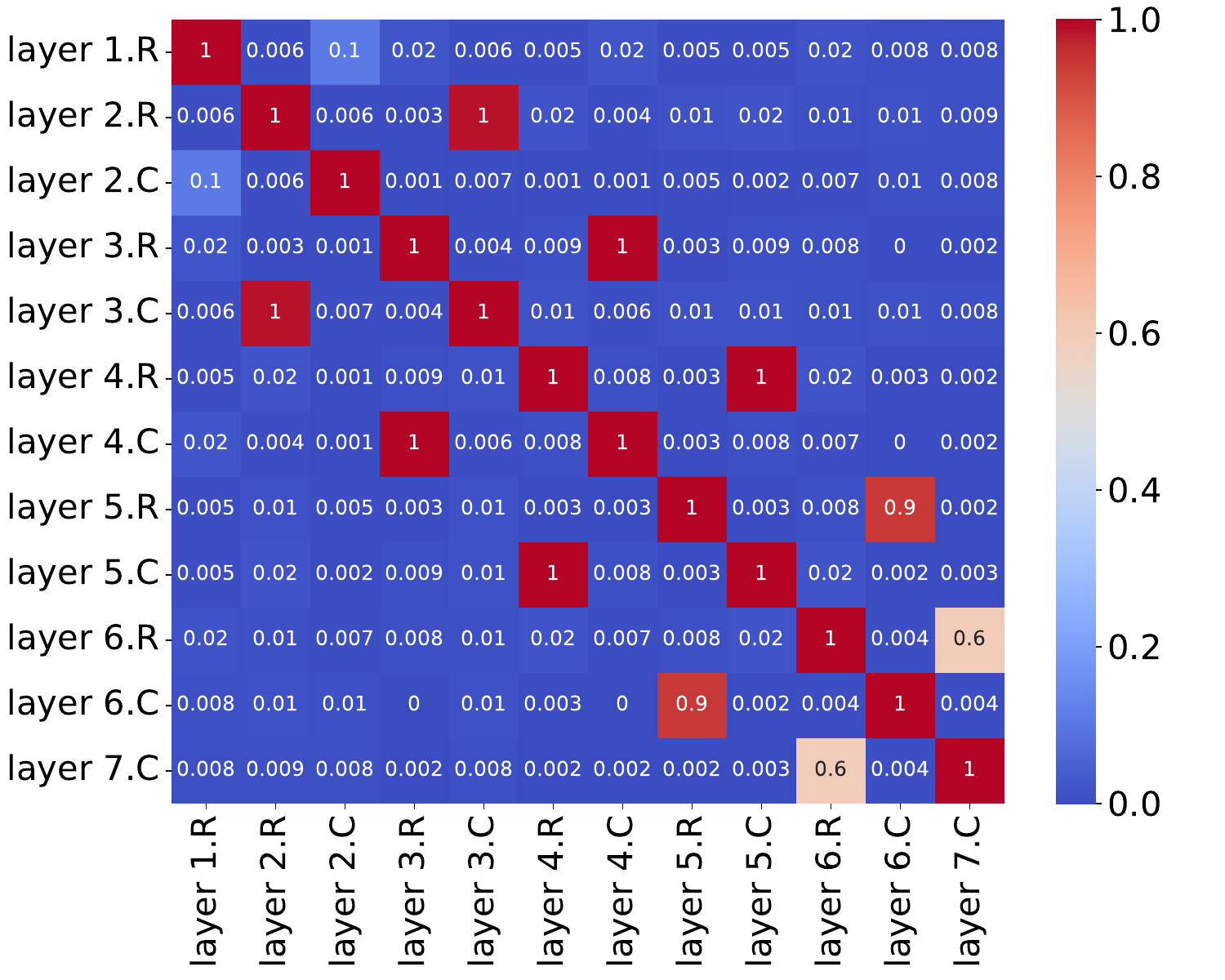}
      \caption{Constant fixed bias}
    \end{subfigure}%
    \caption{ Correlation matrices for a 7-layer MFNN example  under  different bias settings with $N=1000$. (a) Non-constant trainable bias; (b) Constant trainable bias; (c) no bias.    
 } 
    \label{fig:3: 7-layer correlation}
\end{figure*}

We also make a test on the 7-layer example under MFP in Figure \ref{fig:3: 7-layer correlation}, where the $.R$ ($.C$) denotes the corresponding row (column) mean value. The initial weights are i.i.d, and we can see the trained correlation result is as expected to the RC ansatz, the matrices are connected in a way rows follow by columns  (this may be in different order based on matrix calculation in a program). The middle weights of a MFNN shall be independent for $L\geq 5$ when taking constant bias initialisation in \cite{nguyen2023rigorous}, we would like to kindly amend this argument.

\begin{figure*}[h!bt]
    \centering
    \begin{subfigure}{.25\textwidth}
      \centering
      \includegraphics[width=\textwidth,]{ 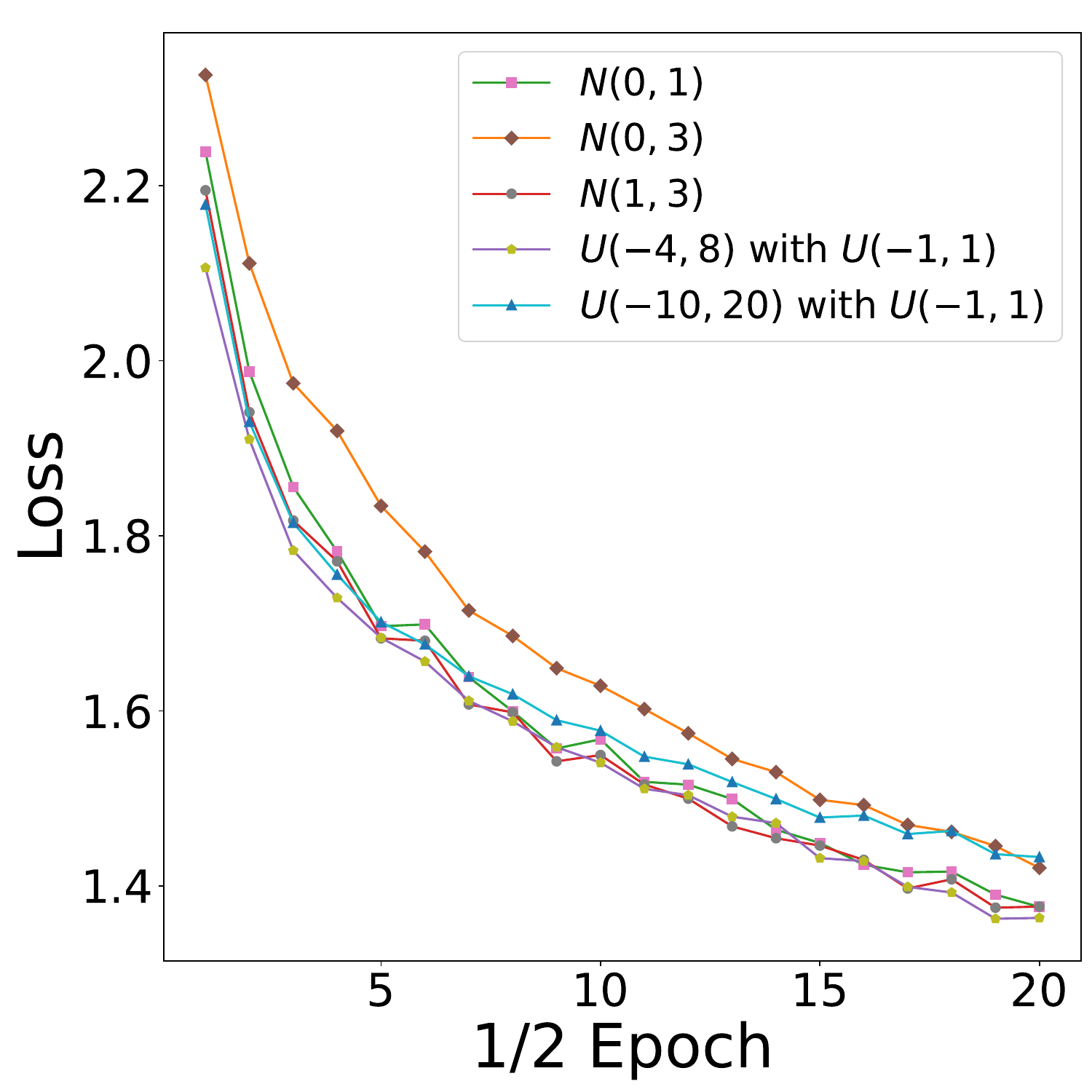}
      \caption{Different initialisation}
    \end{subfigure}%
  \begin{subfigure}{.375\textwidth}
      \centering
      \includegraphics[width=\textwidth,]{ 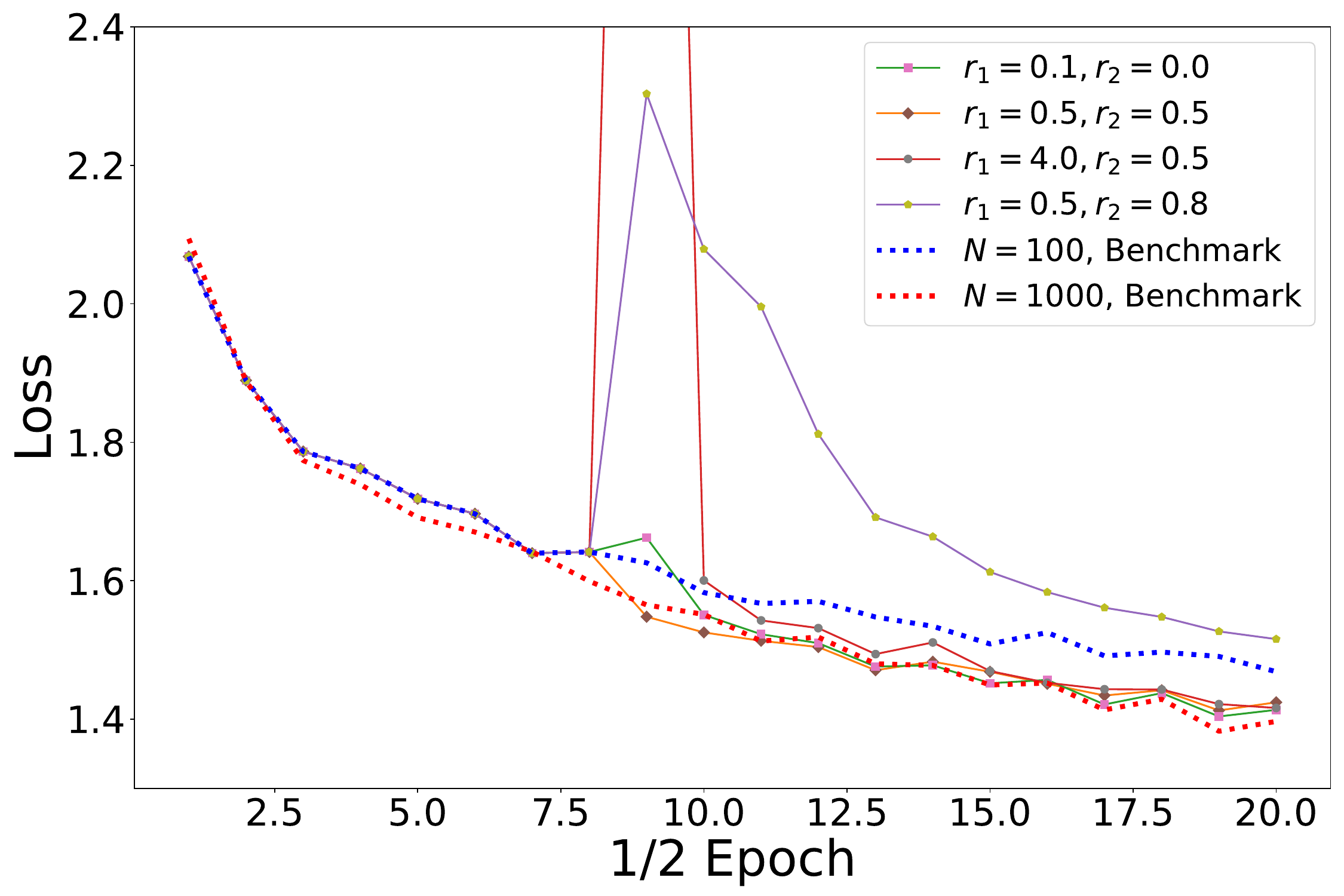}
      \caption{ Loss of small to large }
    \end{subfigure}%
  \begin{subfigure}{.375\textwidth}
      \centering
      \includegraphics[width=\textwidth,]{ 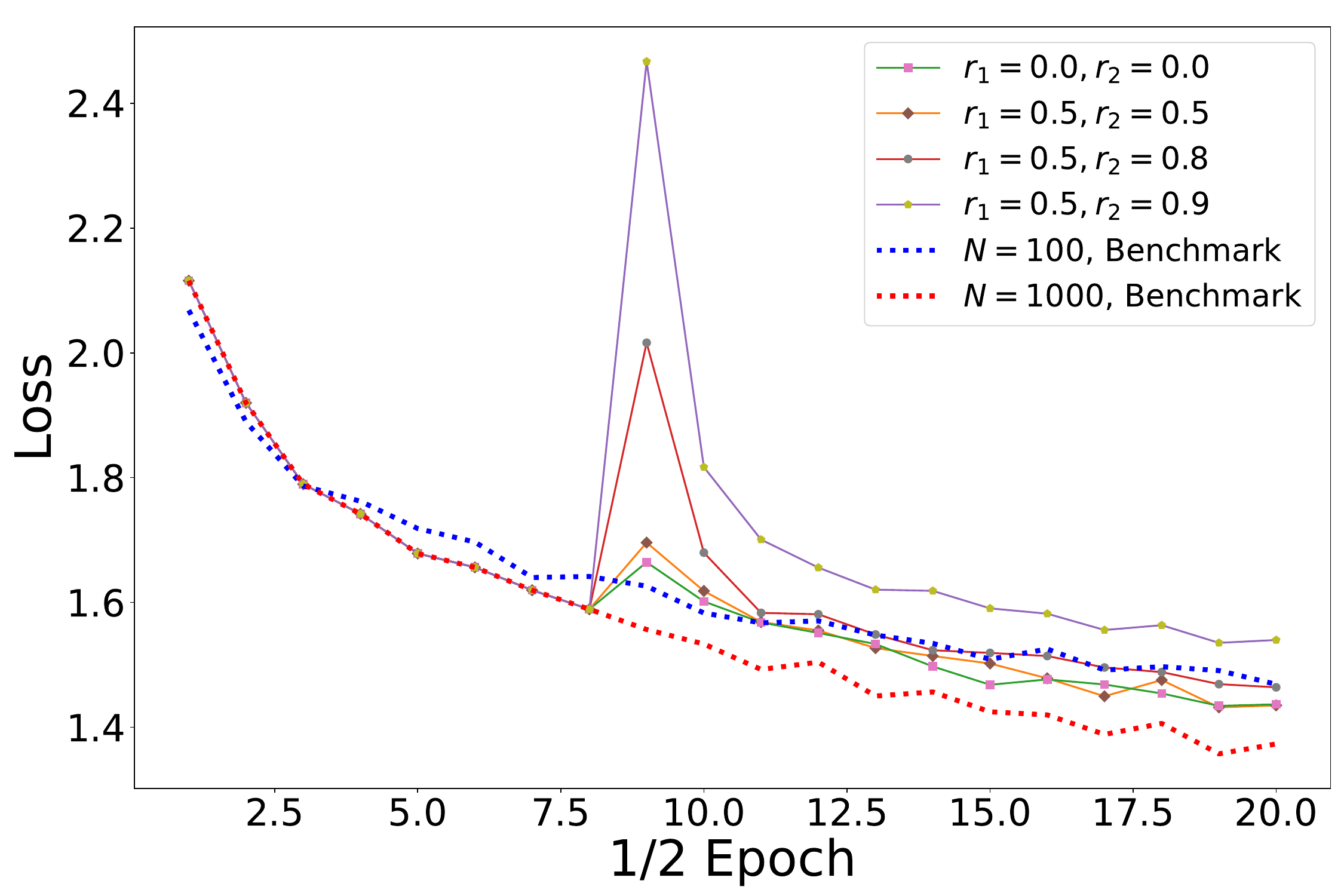}
      \caption{Loss of  large to small }
    \end{subfigure}%
  \\
  \begin{subfigure}{.25\textwidth}
      \centering
      \includegraphics[width=\textwidth,]{ 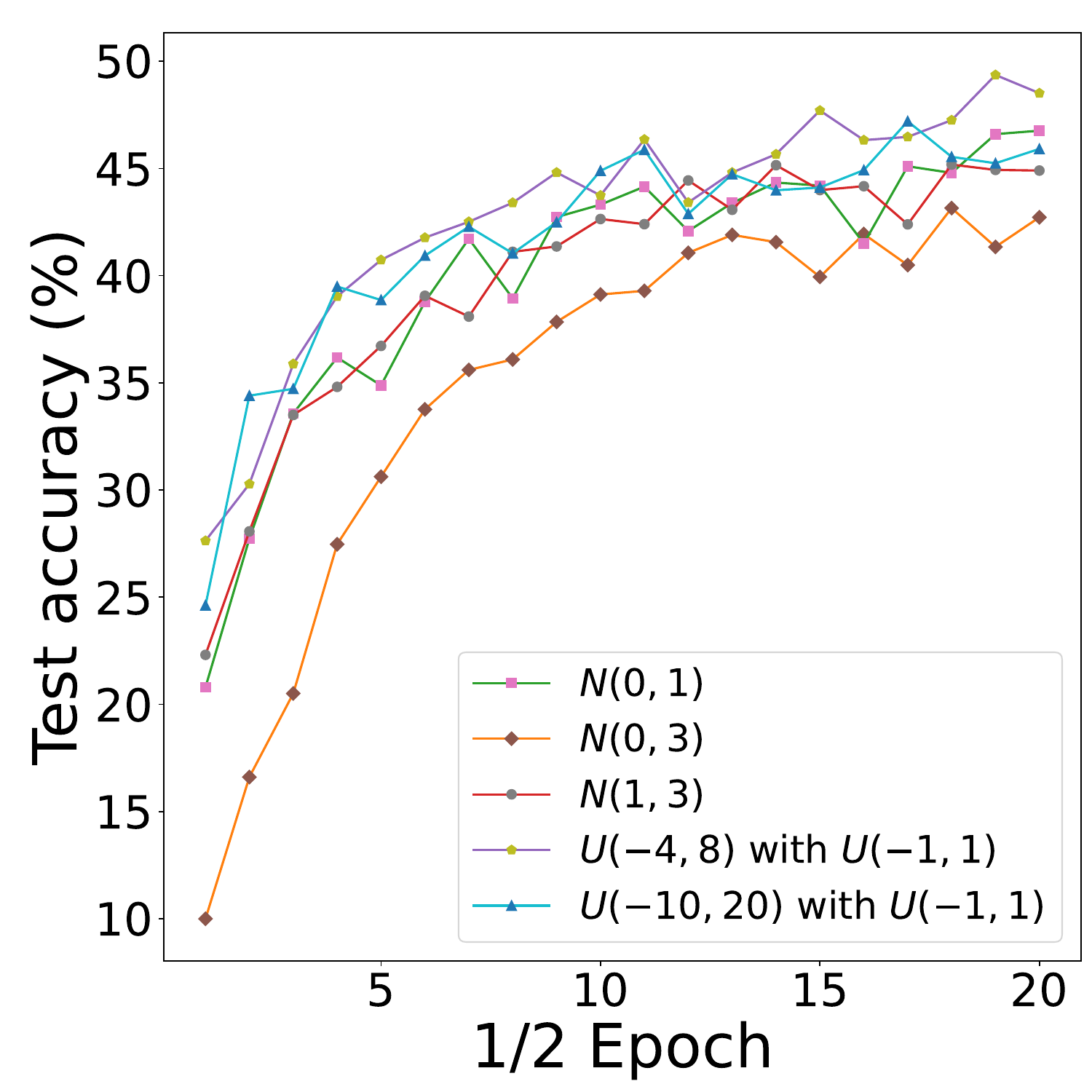}
      \caption{Different initialisation}
    \end{subfigure}%
  \begin{subfigure}{.375\textwidth}
      \centering
      \includegraphics[width=\textwidth,]{ 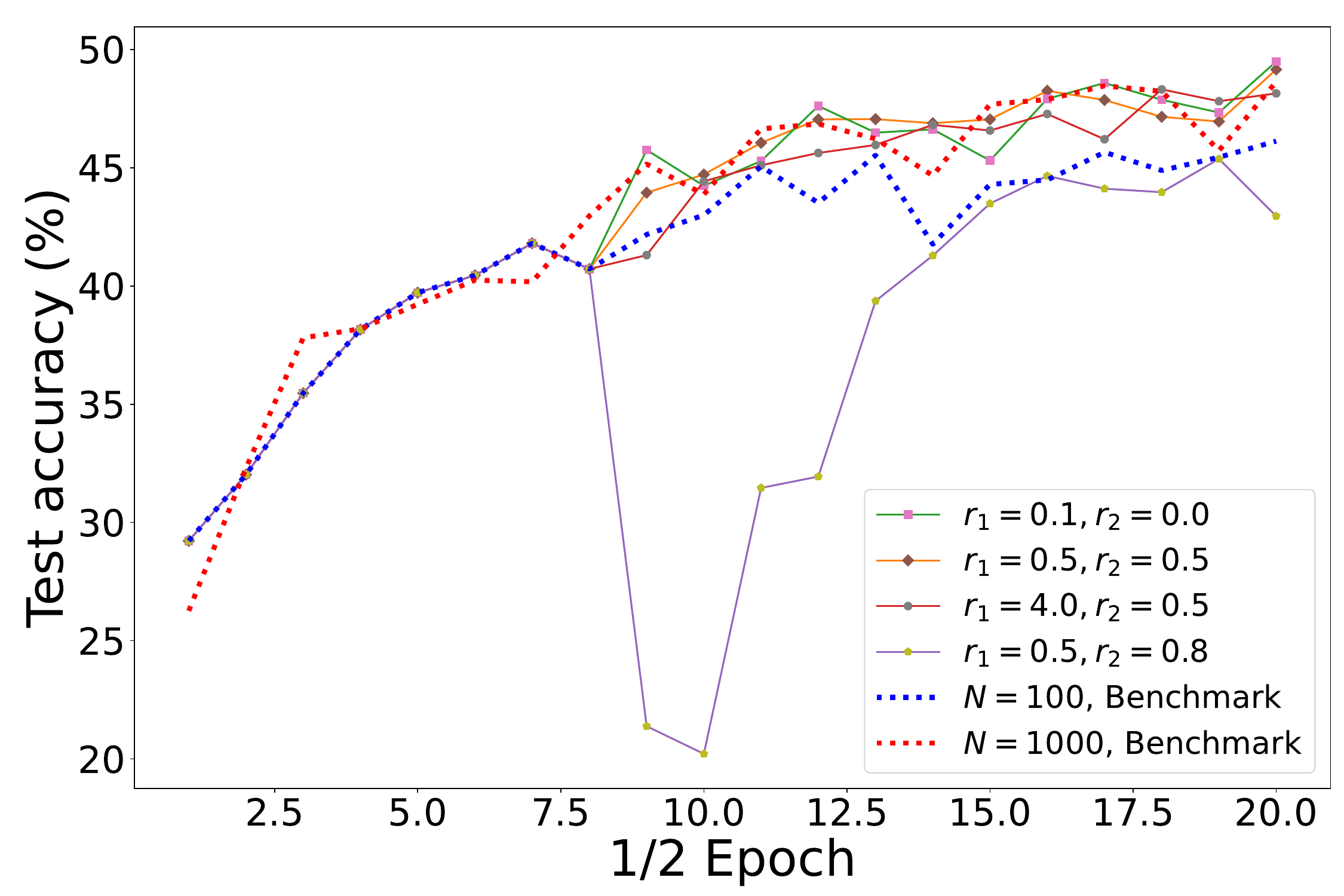}
      \caption{ Acc of small to large }
    \end{subfigure}%
  \begin{subfigure}{.375\textwidth}
      \centering
      \includegraphics[width=\textwidth,]{ 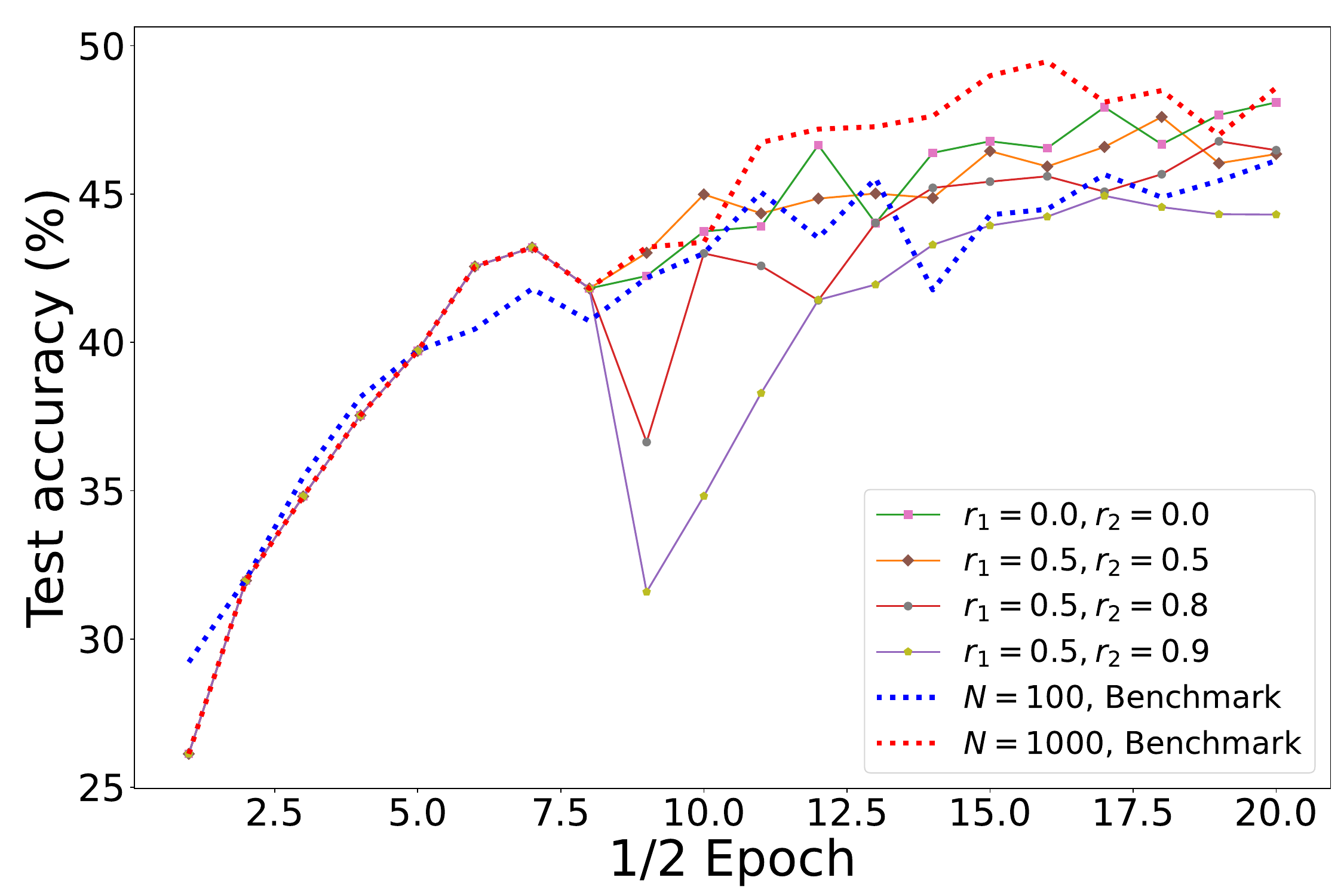}
      \caption{ Acc of  large  to  small }
    \end{subfigure}%
    \caption{  Training loss and test accuracy of the 3-layer NN example under different settings with $\eta = 0.1$. (a) and (d) are of different initial distributions. (b) and (e) are the weight transfer results from the $N=100$ model to the  $N=1000$ model at the 4-$th$ epoch with different random rates $r_1$ and norm rates $r_2$. (c) and (f) are the weight transfer from the the $N=1000$ model to the  $N=100$ model with different random rates $r_1$ and norm rates $r_2$.
 } 
    \label{fig:2: wt}
\end{figure*}
\subsubsection{For Zero-Shot Weight Transfer Verification} In  Figure \ref{fig:2: wt} (b) and (e), we show the weight transfer applying to the MFNN from small model to large model, which is also known as the model growth, $r_1$ is the random rates for the new $\hat{w}$ from $w$ such that $\hat{w}:=w\cdot u$ with $u\sim \cU(-r_1,r_1)$. $r_2$ is the norm rates for the sorted weights by norm via the  $\Gamma$ set where we take the first $1-r_2$ samples with a larger norm. The $r_1$ can be taken up to $4.0$ (blows up at $5.0$) and the $r_2$ can be taken up to $0.8$ in this example. 

In  Figure \ref{fig:2: wt} (c) and (f), we show the weight transfer from the large model to the small model, which is similar to the model pruning,    the $r_1$ can be up to $0.5$ and the $r_2$ can be up to $0.9$.  
In this example,   the weight transfer is quite robust  across different sizes, the performance converges to the benchmark within few epochs for $r_1\leq 4.0,~r_2\leq 0.8$.

Notice that sampling by norm and adding randomness may already change the measure, so that there is no guaranteed this can work for general cases, in fact, one can always develop example violate  this method, we suggest one to consider carefully to develop proper sampling methods in different situations.  

\subsection{Verified on LLMs}
In this section, we would like to first address some similar insights showing the matrices are also connected in a row and column way in the LLMs as in the  MLP examples, we then show the measure updates in GPT-3 using MFNN with a non-zero mean type initialization, and the corresponding weight transfer result. 
\begin{figure*}[h!bt]
    \centering
    \begin{subfigure}{.33\textwidth}
      \centering
      \includegraphics[width=\textwidth,]{ 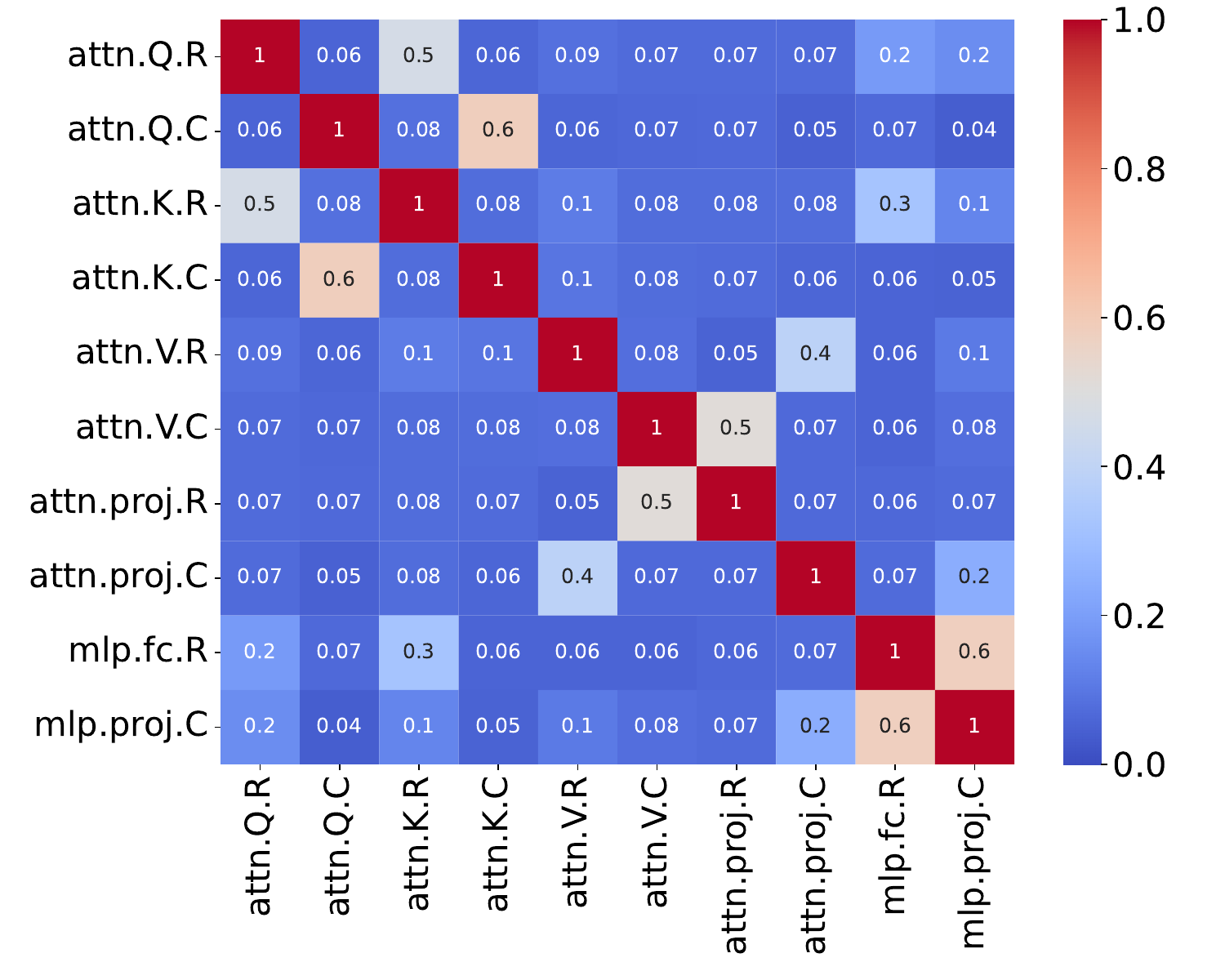}
      \caption{Correlations of GPT-2  }
    \end{subfigure}%
  \begin{subfigure}{.33\textwidth}
      \centering
      \includegraphics[width=\textwidth,]{ 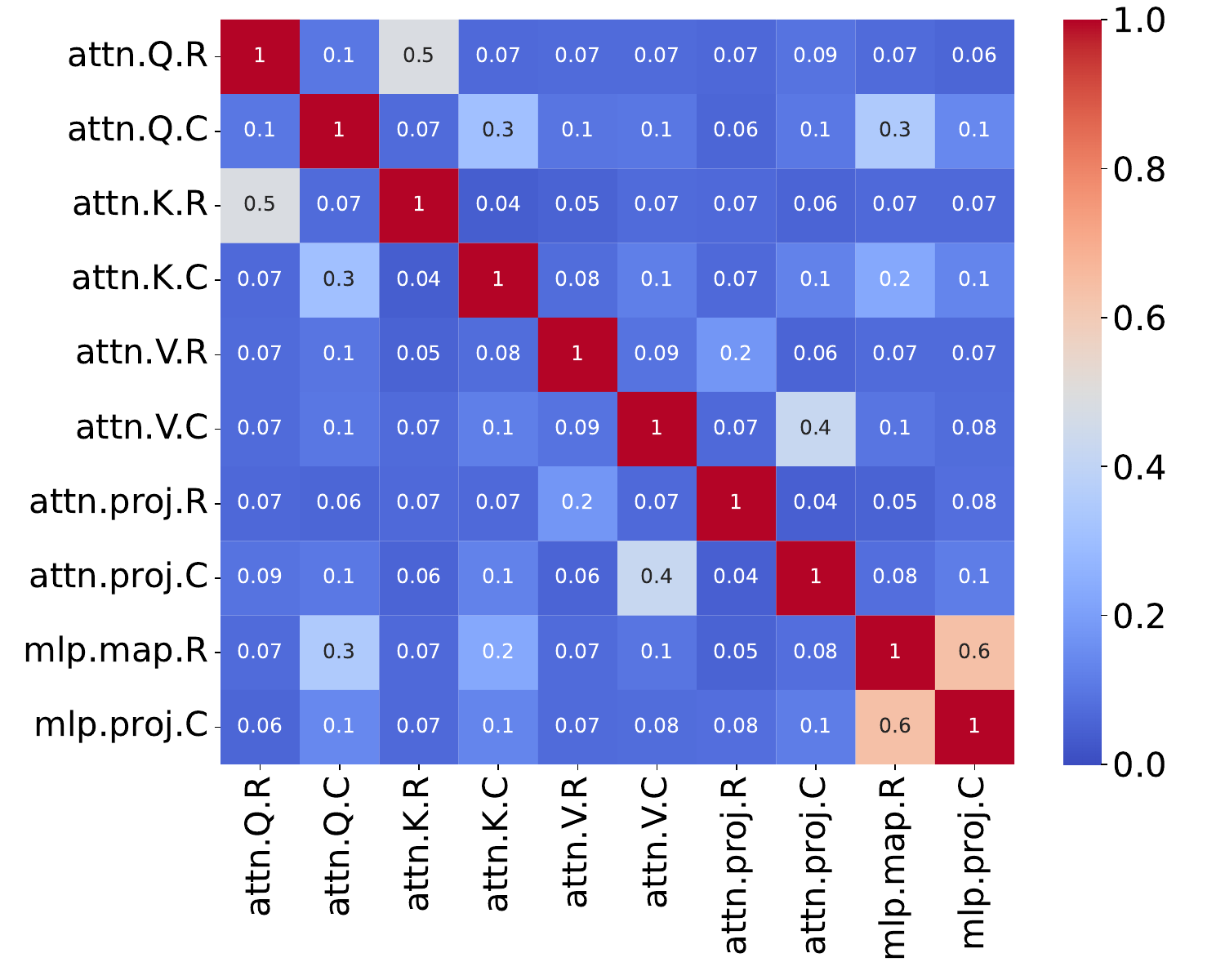}
      \caption{Correlations of GPT-3-MF }
    \end{subfigure}%
  \begin{subfigure}{.33\textwidth}
      \centering
      \includegraphics[width=\textwidth,]{ 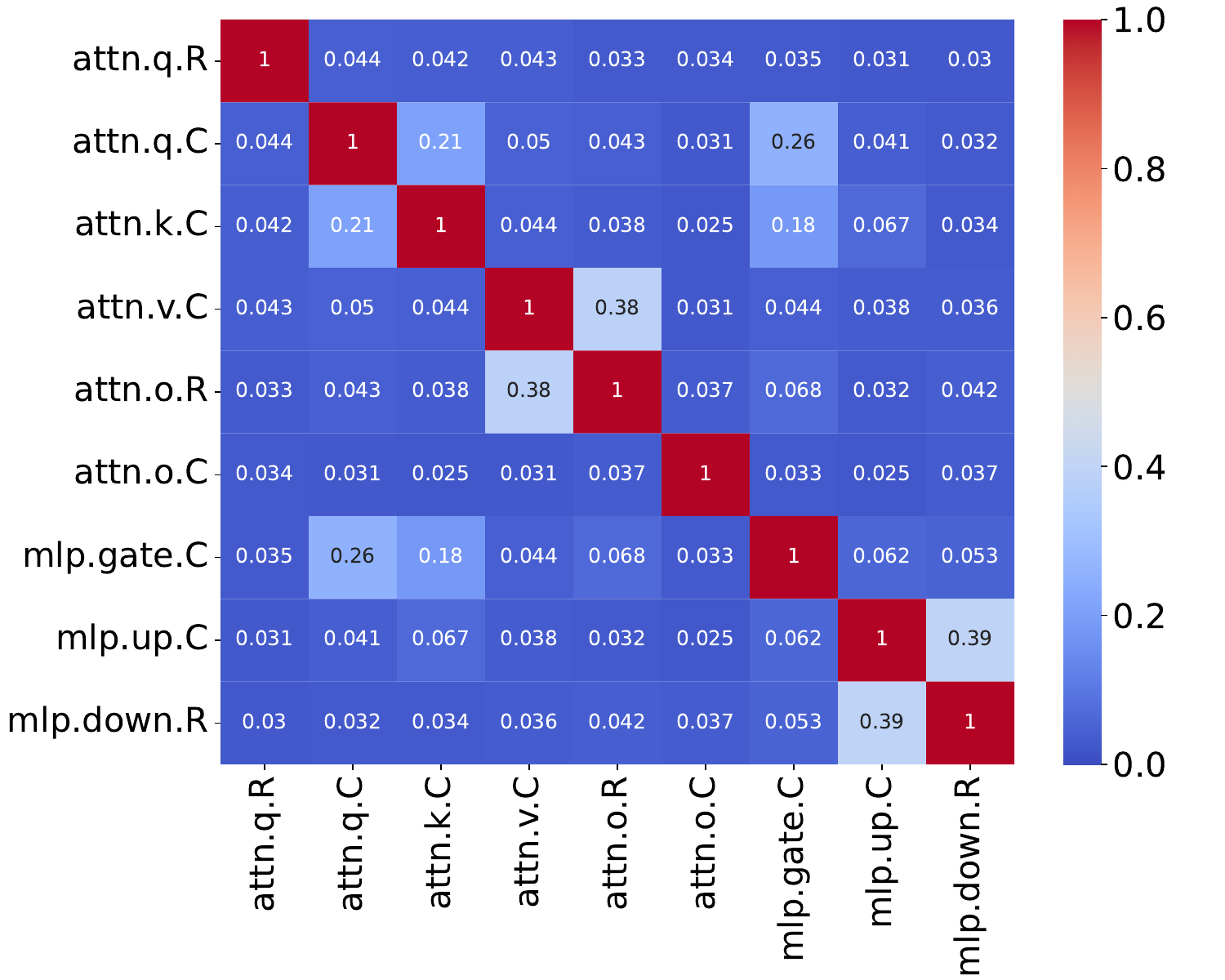}
      \caption{Correlations of Lamma-3.1-8b}
    \end{subfigure}%
    \caption{   Maximum absolute correlation coefficients for different LLMs, where the $.C$ denotes the mean value on rows, $.R$ denotes the mean value on columns, we show the maximum absolute value of the correlation over different blocks in each LLM. 
 } 
    \label{fig:4: LLM correlation}
\end{figure*}
\subsubsection{For RC ansatz verification} Figure \ref{fig:4: LLM correlation} shows the correlation matrices of different LLMs, where we find there exist interesting correlations across different layers as expected, notice that compare to the simple correlation structure in Figure \ref{fig:3: 7-layer correlation}, the skip connection and the attention architecture in LLM leads to more complicate correlation structure, which is discussed in Appendix. We would like to address that there are some slightly different correlation patterns across different LLM, for example, in GPT-2, the $mlp\_fc.weight.R$ is connected with $attn.weightK.R$ while in GPT-3  the $mlp\_fc.weight.R$ is connected with $attn.weightK.C$, this differences comes from the way how the matrix is calculated inside the code.

However, the correlation result  in GPT-2 and Lamma-3.1 are not expected from the theoretical analysis  under   SP, $\mu P$ and even NTK settings due to the $1/\sqrt{N}$ scalar as discussed before, where the i.i.d initialization shall leads to roughly independent result as in Table \ref{table:1: correlation} from theoretical point of view.  
The  correlation here shall comes from the fact that the training steps for LLMs are massively over $N$ (i.e the $T \gg N$), thus the updates of the weights can not be ignored (not as the case in the MLP example with $\cO(1/\sqrt{N})$ compare to $\cO(1)$).
We would like to address the weight of the LLMs   after sufficient training steps (e.g $T \sim  \cO(N)$) admits mean-filed dynamic, thus weight transfer  methods are applicable. 

\begin{figure}[hbt]
    \centering
    \begin{subfigure}{.32\columnwidth}
      \centering
      \includegraphics[width=\columnwidth,]{ 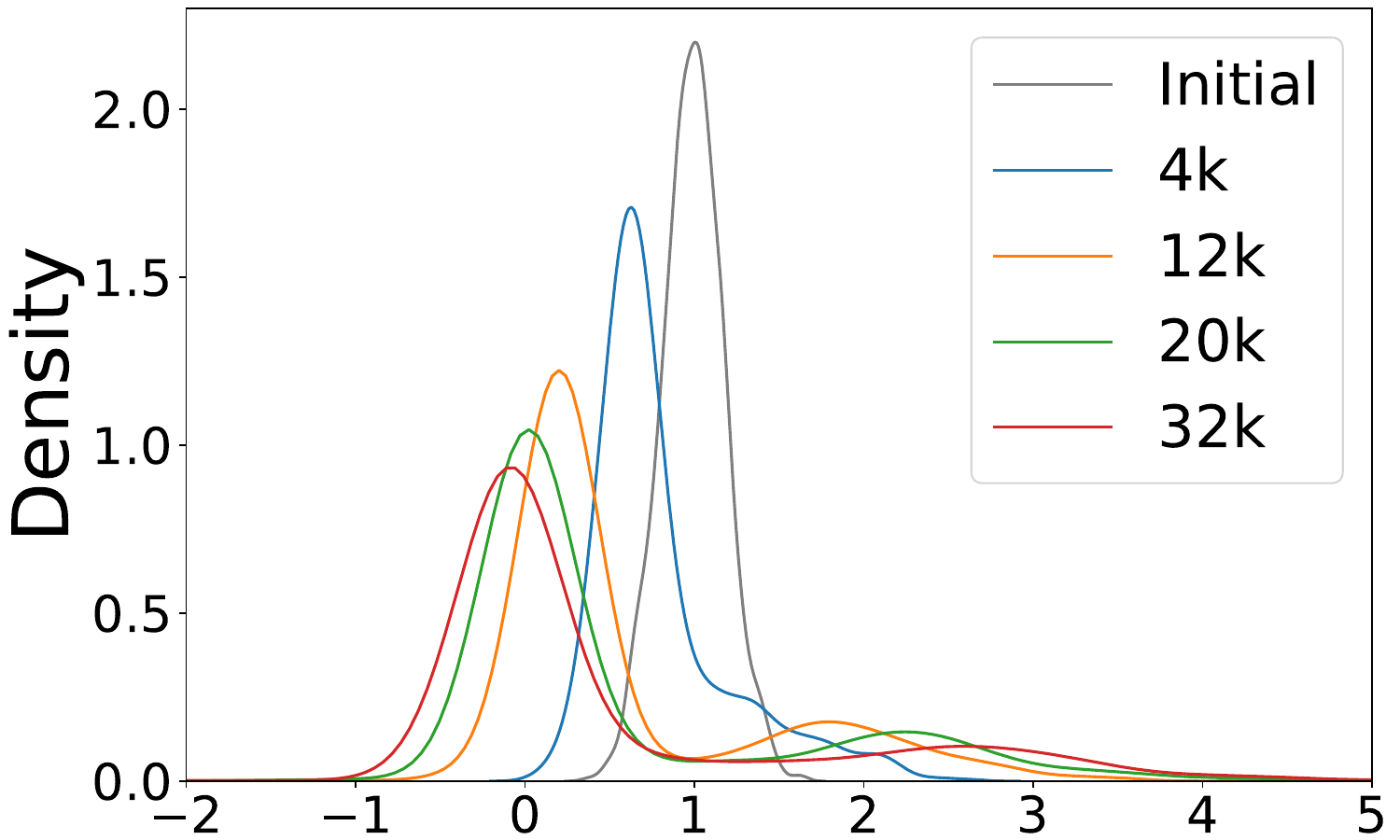}
      \caption{Block 5 in the small model  }
    \end{subfigure}%
      \begin{subfigure}{.32\columnwidth}
      \centering
      \includegraphics[width=\columnwidth,]{ 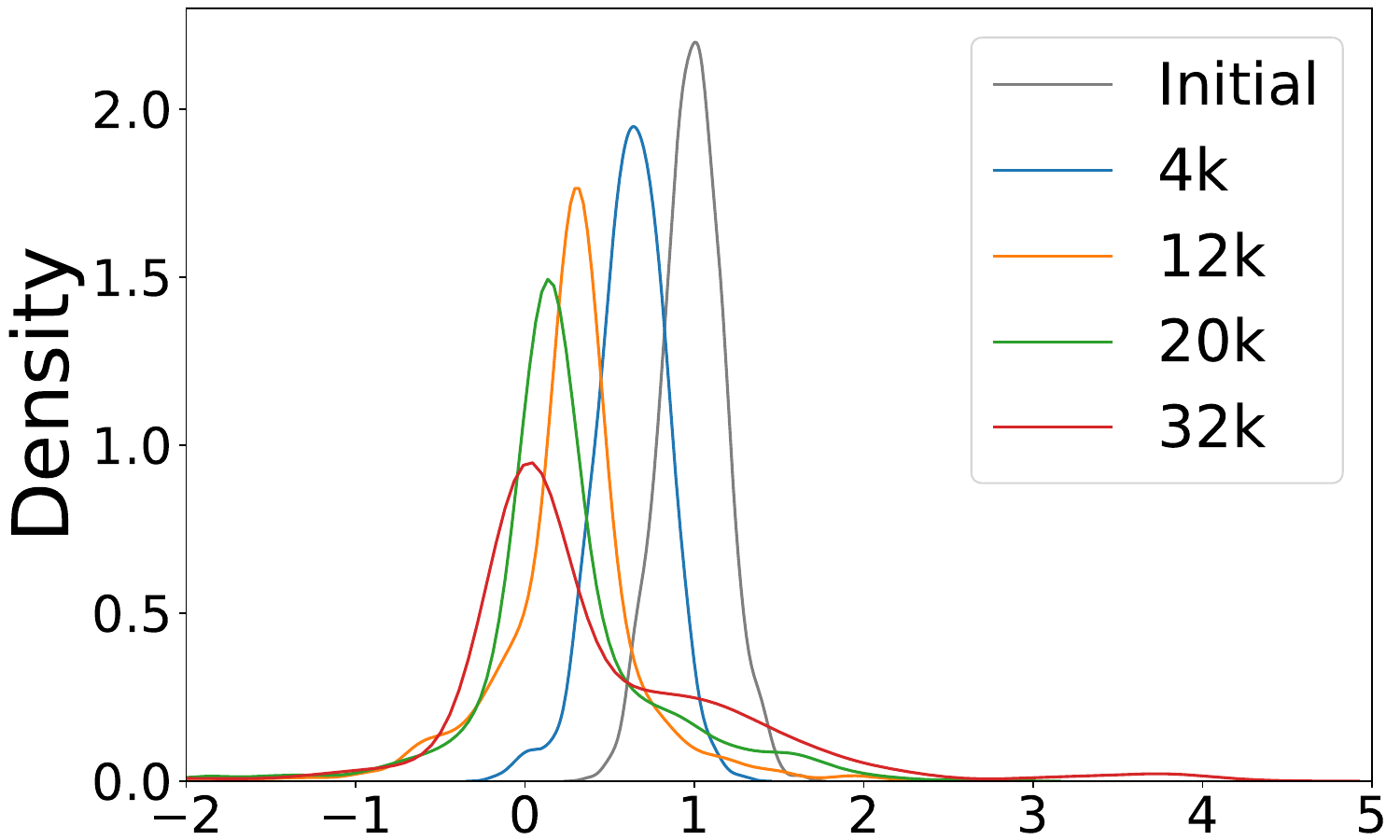}
      \caption{Block 23 in the small model  }
    \end{subfigure}%
    \begin{subfigure}{.32\columnwidth}
      \centering
      \includegraphics[width=\columnwidth,]{ 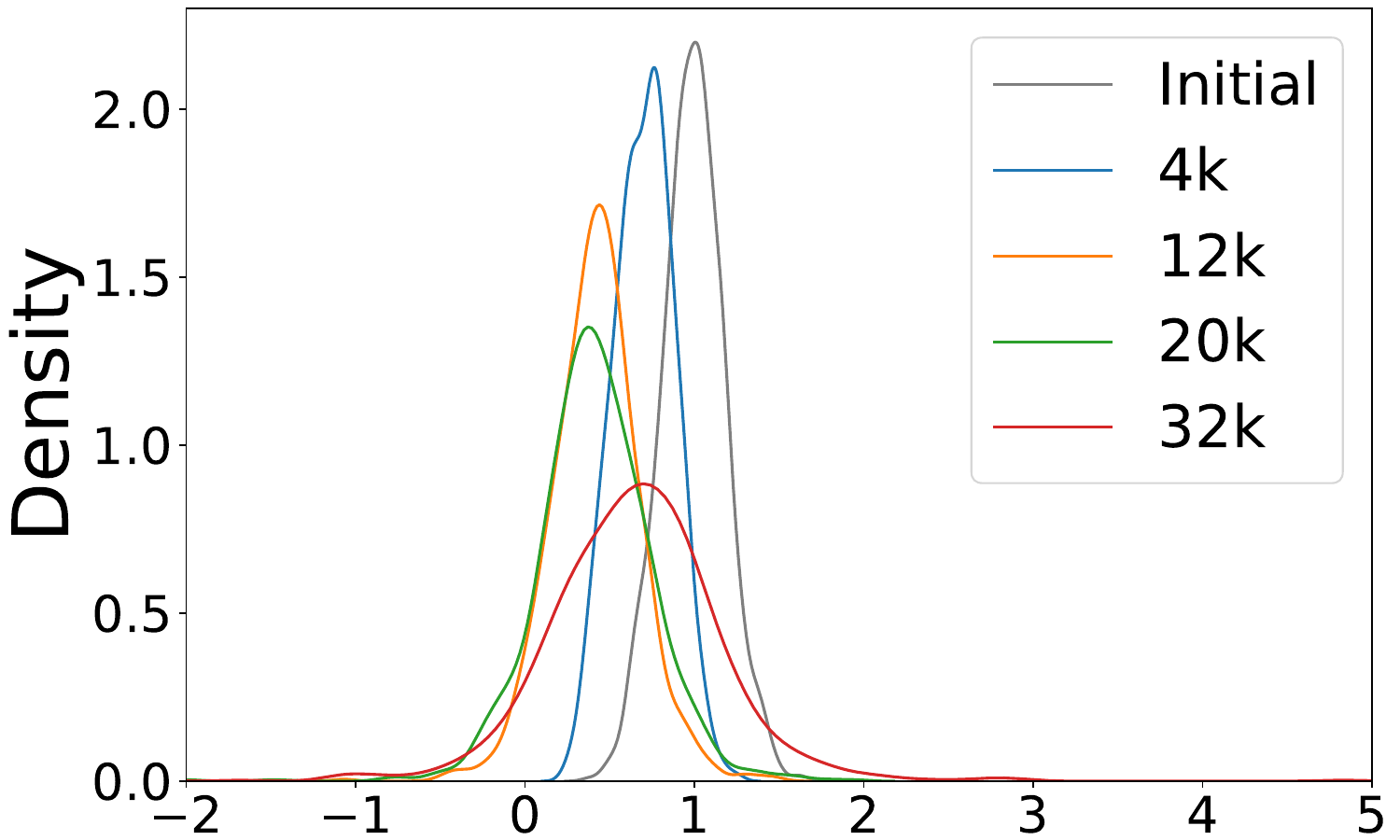}
      \caption{Block 0 in the small model  }
    \end{subfigure}%
      \\ 
       \begin{subfigure}{.32\columnwidth}
      \centering
      \includegraphics[width=\columnwidth,]{ 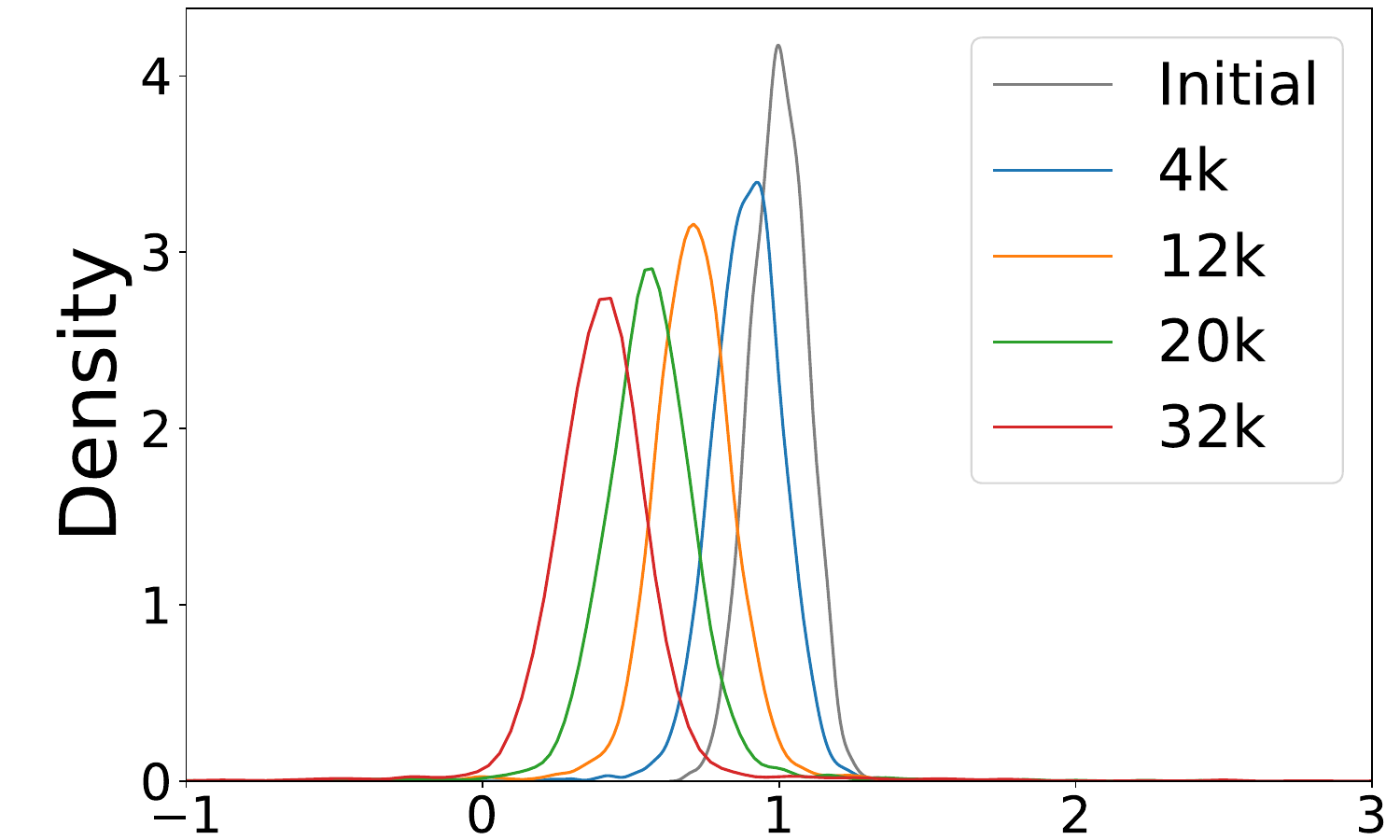}
      \caption{Block 5 in the large model  }
    \end{subfigure}%
  \begin{subfigure}{.32\columnwidth}
      \centering
      \includegraphics[width=\columnwidth,]{ 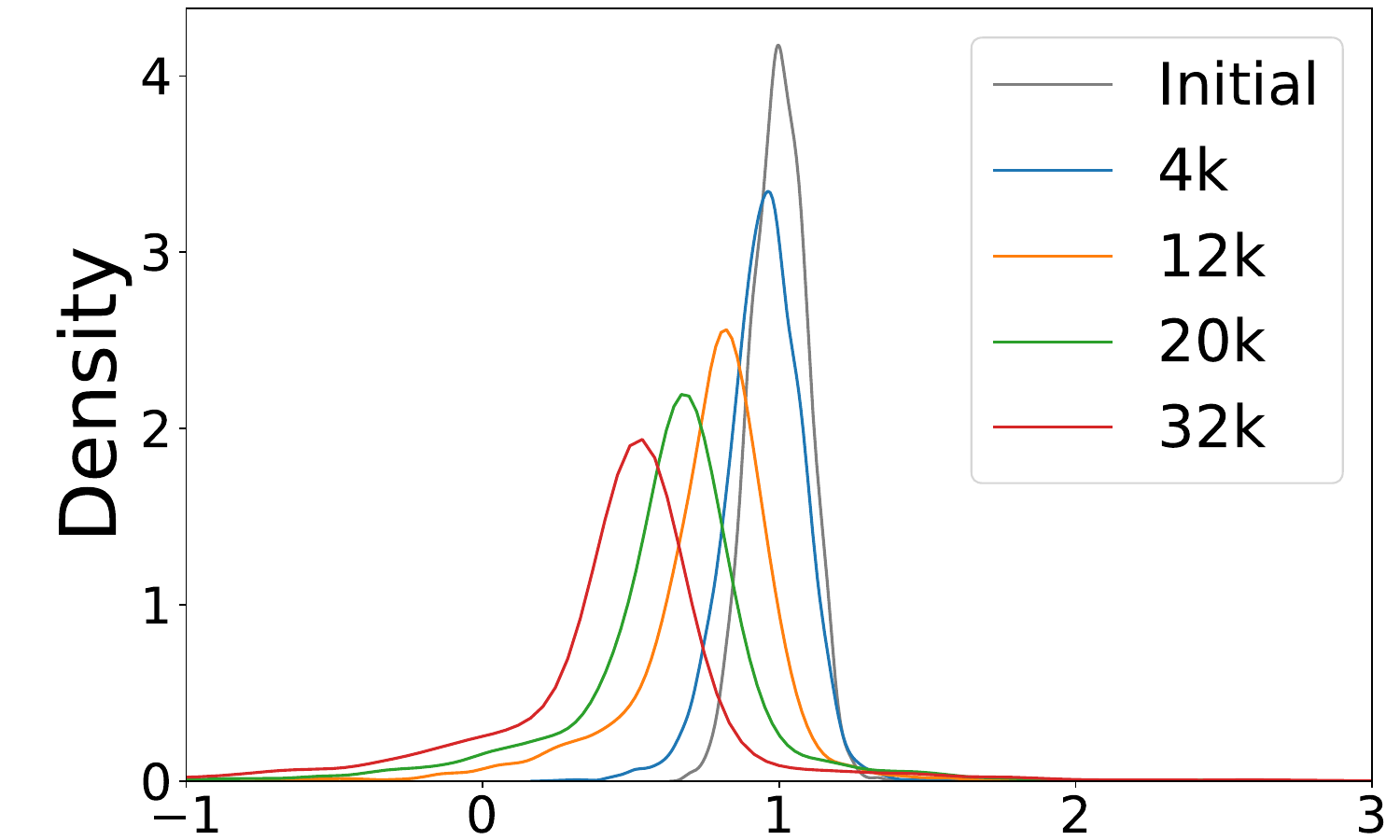}
      \caption{Block 23 in the large model  }
    \end{subfigure}%
  \begin{subfigure}{.32\columnwidth}
      \centering
      \includegraphics[width=\columnwidth,]{ 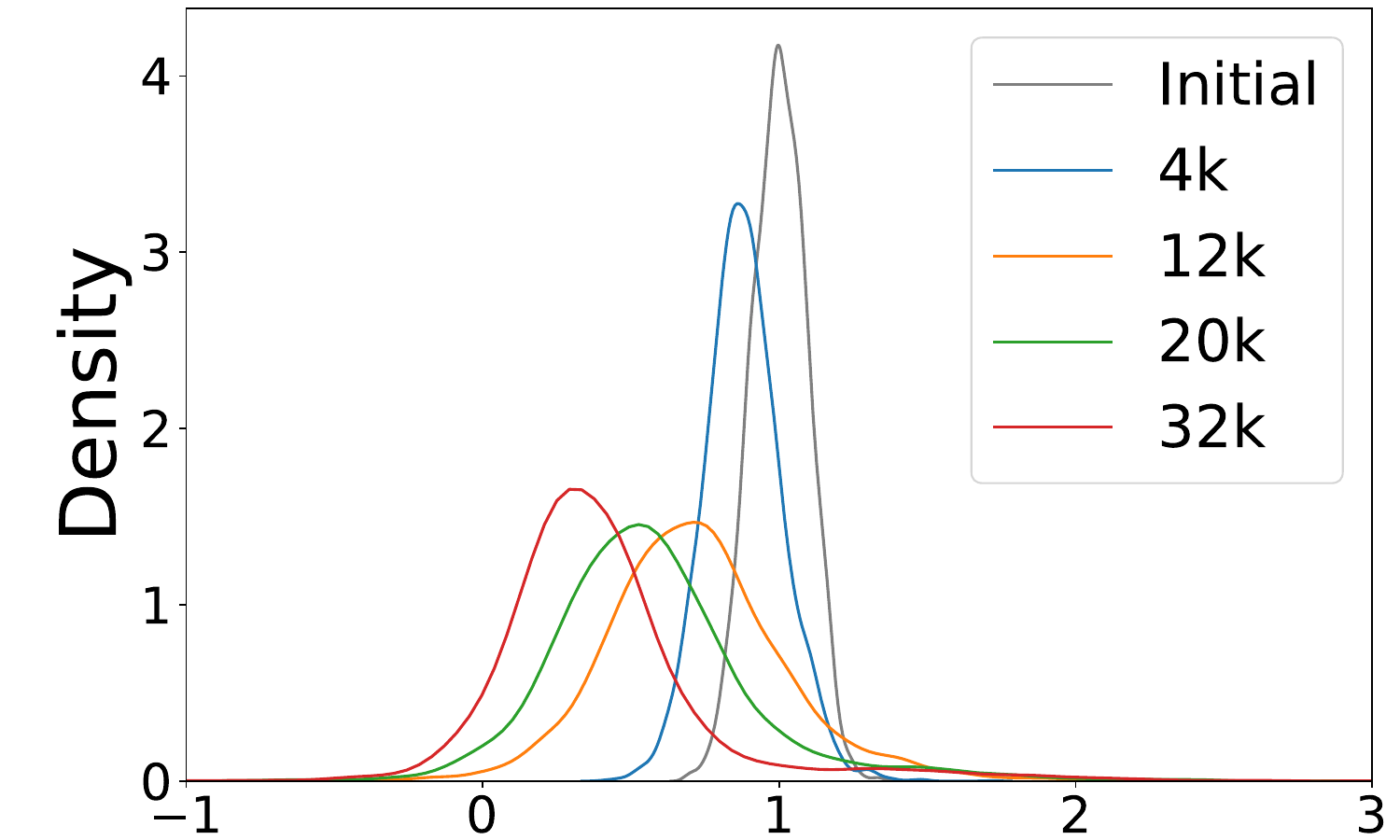}
      \caption{Block 11 in the large model  }
    \end{subfigure}%
  \caption{   
    Distribution of the row means in the MLP layer weight matrix $\bR^{N\times 4N}$ for the small model $N=256$ and large model $N=1024$ in different training step $T\in\{4k,12k,20k,32k\}$ in different blocks with the weight taking i.i.d initialisation of $\cN(1,3)$.
} 
  \label{fig:5: LLM distribution updates}
\end{figure}
\subsubsection{For Zero-Shot Weight Transfer Verification for GPT3} Figure \ref{fig:5: LLM distribution updates} (a)-(f) show the distribution update trajectories.  In Blocks 5 and 23, the distribution moves from the positive side with a mean of 1.0 to the negative side,  the distribution becomes sparser as the step increases. Notice that the row mean admits an initial distribution of $\cN(1,3/\sqrt{N})$. 

Notice that not all blocks admit such a clear distribution update, some blocks admit more complicated distribution behaviour, block 0 in the small model has a more complicated update where the peak moves to the negative side in early stage but moves to the positive side later. Block 11 in the large model has a greater distribution change from $T=4k$ to $T=12k$ where the peak drops faster than others.  
Later, we shall discuss why there is a gap between the distribution of the large model with the small model. 

\begin{figure}[h!bt]
    \centering
    \begin{subfigure}{.8\columnwidth}
      \centering
      \includegraphics[width=.6\columnwidth,]{ 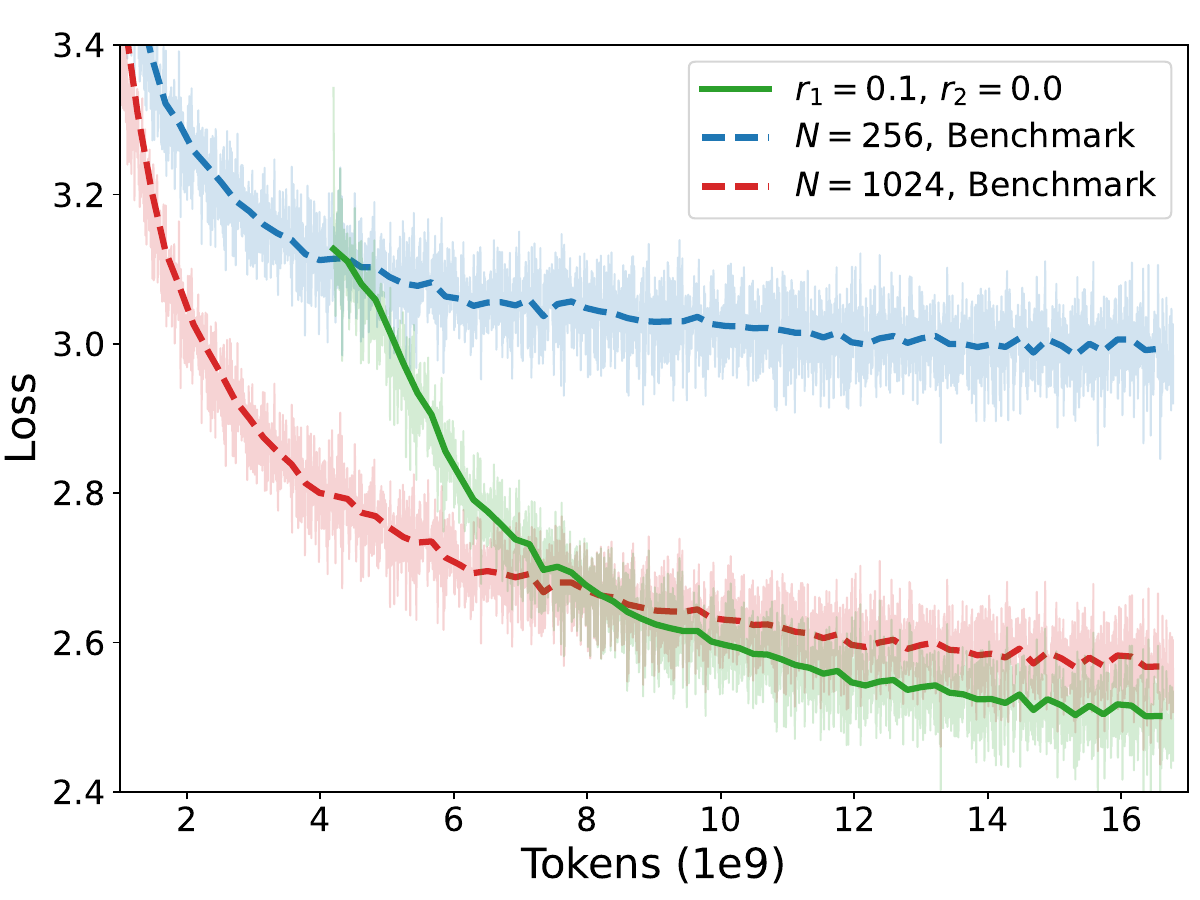}
    \end{subfigure}%
    \caption{   
      Weight transfer for the MF setup GPT-3, the initial distribution is taken $\cN(1,3)$, where  we take $N=256$ for the small model and $N=1024$ for the large model. 
 } 
    \label{fig:6: GPT-3 WT }
\end{figure}

In Figure \ref{fig:6: GPT-3 WT }, we apply the weight transfer to the MFNN from the small model with $N=256$ to the large model with $N=1024$. We sample the weights at $T=4B$ and use the naive weight transfer as Example \ref{example: wt} under Algorithm \ref{alg: RC wt} (same as the model growth method in Net2Net) with random ration $r_1 = 0.1$ as the MLP example, the large model after weight transfer works as expected and even perform better than the benchmark large model.  
 
Combining the result in Figure \ref{fig:5: LLM distribution updates} and Figure \ref{fig:6: GPT-3 WT }, we would like to discuss 2 questions: 1.Why do the weight distributions of the large and small models not appear to converge? 2. What explains the superior performance of the weight transfer model compared to the benchmark large model?

For question 1: we apply the Adam optimiser during training, and unlike the traditional propagation of chaos (PoC) type convergence result like \cite[Theorem 3.14]{adams2020large} for the mean-field process where the timestep size is fixed, the training timestep size is connected to the training loss. In Figure \ref{fig:6: GPT-3 WT }, we can see the loss value in the small model is always higher than the loss in the large model, which leads to the actual update stepsize of the small model is larger than the stepsize of the large model. The distribution update in the large model is slower in Figure \ref{fig:5: LLM distribution updates}, which leads to a distribution gap. 

For question 2: if we further assume there exists some limit distribution $\mu^*$ as $T\rightarrow \infty$, this means the small model admits a faster convergence to $\mu^*$ than the large model but small $N$ connected to the PoC error let the small model have a higher loss even it is closer to the $\mu^*$ in the time sense.  This means the weight transfer from the small model has some benefit on time $T$, it needs to complement the PoC error by increasing the $N$  and can reach a better loss result at $T=32k$.

\section{Conclusion}

In this work, we show the mean field point of view to MFNN is adequate by using the RC ansatz and experiments to theoretically and empirically support this view. From the mean-field standpoint, we show the weight transfer methods are equivalent to sampling methods from the corresponding empirical distribution, verified through experiments on MLPs and LLMs.
Furthermore, we illustrate the applicability of the RC ansatz to LLMs trained for sufficiently long, even without using the MFP, which provide theoretical support for the corresponding Zero-short weight transfer.

From a mean field perspective, the NN training process can be viewed as searching for an optimal measure in the measure space. However, this approach presents several challenges. Firstly, there is a limited guarantee of the existence and uniqueness of the optimal measure. In general settings without regularisation, certain measures may yield the same result. Secondly, how to reach the optimal measure remains unclear. We believe analysing the measure movement trajectories could provide useful insights. These unresolved questions are left for future research.




\newcommand{\etalchar}[1]{$^{#1}$}

\newpage 
\onecolumn
\section{Appendix}
\label{sec: appendix}

\subsection{ Notation and Spaces }
Without state specifically, in this paper, we consider the mean field neural network using setup as in \cite{mf-2022-path2,nguyen2023rigorous}, where different types of neural network is introduced in Example \ref{example: differnt setups}, we refer to \cite{tp4} for a more detailed comparison.

We follow the notation and framework set in  \cite{2024weak}. 
For a vector in $\mathbb{R}^d$, $d \geq 1$, we will write $\boldsymbol{x} := (x_1,\ldots, x_d) \in \mathbb{R}^d$. The inner product of two vectors $\boldsymbol{a}, \boldsymbol{b} \in \mathbb{R}^d$ is denoted by $\left \langle \boldsymbol{a}, \boldsymbol{b} \right \rangle$ and for standard Euclidean norm we will use the notation $|\cdot|$. Throughout this article $\mathcal{O}(\cdot)$ refers to the standard Landau (big `O') 
 notation. 
For a twice continuously differentiable function $f:\mathbb{R}^{d} \to \mathbb{R}$, we denote by $\partial_{x_i}f: \mathbb{R}^{d} \to \mathbb{R}$ the partial derivative with respect to the $i$-th component. 

 Let our probability space be a completion of $(\Omega, \bF, \mathcal{F},\bP)$ on which all the random variables are defined. The probability space is
equipped with a filtration $\bF=\lbrace \mathcal{F}_t \rbrace_{t\geq 0}$  contains all $\bP$-negligible sets. We denote by $\bE[\cdot]=\bE^\bP[\cdot]$, the expectation with respect to $\bP$.  For any $ p\geq 2$, we define $L^p(\Omega, \bR^d)$ as the space of $\bR^d$-valued measurable random variables such that $\bE[\,|X|^p  ]^\frac1p <\infty$. Let $\mathcal{P}_p(\bR^d)$ denote the space of probability measures $\mu$ on $\bR^d$ such that $\int_{\bR^d} |x|^p \mu(\mathrm{d}x) <\infty$.   
Given a measure $\mu \in \cP(\bR^d)$ and a function $f:\bR^d \rightarrow \bR$, we denote the notation $\langle f, \mu \rangle$ such that  

 \begin{align*}
     \langle f, \mu \rangle := \int_{\bR^d} f(x) \mu (\dd x ). 
 \end{align*}

\subsection{ More about weight transfer  }
\label{sec: more on wt}

We provide the detailed weight transfer framework in Algorithm \ref{alg: RC wt detialed} to enrich the Algorithm \ref{alg: RC wt} with some auxiliary functions.  To make the framework clear, we provide the following example applying weight transfer to width $\tilde{N} \in \bN$:

\begin{algorithm*}[htb!]
	\SetAlgoLined
	\KwIn{ 
        The weights set $ \btheta^N :=\big \{ \{W^{(\ell)}\}_{\ell =1}^{L_w} \cup \{B^{(\ell)}\}_{\ell =1}^{L_b},U,V,B^{(U)},B^{(V)} \big \}$ in the NN with corresponding  $\gamma$ notation and $\Gamma$ set, functions $ F^{\cR }, F^{\cC }$ to generate the row and column information, functions $ G^{\cR }, G^{\cC }$ to generate the new matrix based on row and column indices. 
        }
        \  
        \
        \For{$i \gets 1$ to $|\Gamma|$}{
            $\btheta_i \gets \{ \emptyset \}$                       \;
            \For{$\bR \ni \eta$ in $\Gamma_i$}{
                \For{$W$ in $\btheta^N $}{
                    \If{
                    $\eta$  has shown up at the lower index of $W$ in the NN computation
                    }{ 
                        \eIf{$\eta$ at the row index  }{$\btheta_i  =  \btheta_i\cup F^{\cR }(W)$ \;}{
                        $\btheta_i  =  \btheta_i\cup F^{\cC }(W)$ \;
                        }
                      
                    }
            }      
            
            }
            Compute the new index set $\cI_i \in \bR^{N_i}$ for the weights used in $\Gamma_i$ based on $\btheta_i$ and the target width $N_i$ \;
            \For{$\bR \ni \eta$ in $\Gamma_i$}{
                \For{$W$ in $\btheta^N $}{
                    \If{
                    $\eta$  has shown up at the lower index of $W$ in the NN computation
                    }{ 
                        \eIf{$\eta$ at the row index  }
                        {$W \gets  G^{\cR }(W,\btheta_i)$ \;}
                        {
                        $W \gets  G^{\cC }(W,\btheta_i)$ \;
                        }
                    }
            }      
            }
	}
        \KwOut{The transferred weights  set $ \btheta^N$.    }   
	\caption{ Weight transfer framework }
	\label{alg: RC wt detialed}
\end{algorithm*}

\begin{example}[Weight transfer for Example \ref{example: gamma set} ]  
    \label{example: wt-detailed} 
    For the target width $\tilde{N} \in \bN$, the NN after weight transfer admit the following 
    \begin{align}
     \nonumber 
        f(x)  =& 
           \frac{1}{\tilde{N}}\sum_{ {\gamma_3}=1}^{N}    V_{{\gamma_3}} \psi  
           \bigg(   \frac{1}{\tilde{N}}\sum_{ {\gamma_2}=1}^{N}  W^{(2)} _{{\gamma_3},{\gamma_2}}
            \psi  \Big(   \frac{1}{\tilde{N}} 
         \sum_{ {\gamma_1}=1}^{N}    W^{(1)} _{{\gamma_2},{\gamma_1}} \psi
        \big(   U_{ {\gamma_1}} x  +  B^{(U)} _{ {\gamma_1}} \big) + B^{(1)} _{ {\gamma_2}}
        \Big) + B^{(2)} _{ {\gamma_2}}
        \\
        \label{eq: wt example: mlp nn function - 4-layer 1d }
        & \qquad 
        + \frac{1}{\tilde{N}}
         \sum_{ {\gamma_1}=1}^{N}    W^{(1)} _{{\gamma_3},{\gamma_1}} \psi
        \big(   U_{ {\gamma_1}} x  +  B^{(U)} _{ {\gamma_1}} \big) + B^{(1)} _{ {\gamma_3}}
        \bigg) 
        +  B^{(V)} _{ {\gamma_4}},\qquad \gamma_4 = 1, 
\end{align}
where the $\Gamma=\big\{   \{\gamma_1\},  \{\gamma_2,\gamma_3\}, \{\gamma_4\}    \big\}$, $|\Gamma|=3$,  so that 
\begin{enumerate}
    \item the row of $U, B^{(U)}$ and the column of  $W^{(1)}$ shall generate with the index set $\cI_1 \in \bR^{ \tilde{N} }$ from Algorithm \ref{alg: RC wt detialed};
    \item the row of $V, W^{(2)},W^{(1)},B^{(1)},B^{(2)}$ and the column of  $W^{(2)}$ shall generate with the index set $\cI_2 \in \bR^{ \tilde{N} }$ from Algorithm \ref{alg: RC wt detialed};
    \item the row of $ B^{(V)}$ shall generate with the index set $\cI_3 = \{1\}$ from Algorithm \ref{alg: RC wt detialed}.  
\end{enumerate}

\end{example}

\subsubsection{Discussion on the  RC ansatz, the weight transfer, and the real world case}
We would like to discuss    the RC ansarz \ref{ansatz: rc ans} for MFNN  and different model growth methods in \cite[Section 3.1]{fujie2024} under the framework in Algorithm \ref{alg: RC wt detialed}.  

The RC ansatz introduced in this paper supports the mean field point of view of the MFNN, where the weights after training admit some distribution, we expect to have some limit measures as the width $N\rightarrow \infty$,   though they may not be unique.  Similar to the standard Monte Carlo method, where a larger sample size is expected to bring a better result, in the MFNN, we also expect that a larger width of weights (thus more samples) leads to better performance.

From the mean-field point of view, training a MFNN is to search for a "nice measure" to satisfy our requirement,  thus the weight transfer may play a crucial role to the MFNN. We can find the "nice measure" by starting from different guessed measures on small models and applying the weight transfer to a large model to enrich the samples, where the propagation of chaos results shall be included.  
However, there are many miseries behind this methodology, just to introduce some: 1. for a given loss value,   the existence and uniqueness of the corresponding measure are not guaranteed by existing theoretical results; 2. the existence of a limit measure for the MFNN to reach a given loss value is unknown, we refer to \cite{e2020banach} for a more detailed discussion.  

Notice that the MFNN may be far from what is being used in practice, where most people used SP and $\mu$P when training the NN. The neural-tangent kernel (NTK) theory \cite{2018ntkorigin} provides another aspect of the training dynamic, where the choices of the scalars for SP and $\mu$P  in Example \ref{example: differnt setups} are closer to the NTK. 

In the experiment section, we show that it is good to consider the mean field point of view for a NN even if the initial setting is SP or $\mu$P.  In the LMM example, there exists row and column correlation which is not supported by the NTK theory due to the lazy training, more detailed results, and discussion in Figure \ref{fig:4: LLM correlation}. Thus,  we suggest to consider the RC ansatz.

Now, for different model growth methods introduced in  \cite[Section 3.1]{fujie2024},  from the mean filed point of view with the RC ansatz, each method corresponds to sampling strategies $ G^{\cR }, G^{\cC }$ in Algorithm \ref{alg: RC wt}, for example, adding randomness to the weights is a standard technique to produce more samples from a given sample sets, adding zeros corresponding to measure change by allocate measure to the Dirac 0 measure, etc.    The model pruning methods work similarly to some sampling methods introduced at the end of the Appendix, where the optimized sampling method with different test functions can be taken to be the output of the corresponding NN layer thus corresponding to different pruning strategies.  

Many sampling methods from the optimal transport literature can help to find the nice measure based on the measure update trajectory shown in Figure \ref{fig:5: LLM distribution updates}, and one can also consider the corresponding Fokker-Planck equations \cite{2018-mf-1,2019-mf-lln} to develop measure searching proper strategy.

\subsection{   the MFNN toy models and the RC-ansatz }
\label{section: mfnn and new ans}

In this section, starting from the calculations of the simplified 2-layer NN under proper assumptions, we aim to show the main ideas and steps of the analysis from the mean-field point of view. We then provide a more detailed discussion of the simplified 5-layer NN case, where we show the difficulty when building a proper measure structure (which is not the case for the 2-layer case)   by introducing a wrong measure ansatz, we then introduce the proper one: the RC-ansatz, under which the measure dynamic is closed.  We also provide a receipt to build the corresponding measure structure for general NN architectures. Notice that this paper is not for mathematical rigorous arguments, we do not provide the well-posedness results for the corresponding MFNN functions, we try to convey the key idea instead. The rigorous mathematical proof is left for future work following the approaches in  \cite{mf-2022-path2,nguyen2023rigorous}.

\subsubsection{Analysis of the 2-layer NN}
\label{section: 2layers}

In this section, we follow the analysis step in  \cite{mf-2022-path2}, the detailed analysis receipt is provided in later Section. 
Consider the dataset pairs $\{\bx,\by\} \sim \pi$ for some distribution $\pi \in \cP(\bR^{D_x+D_y})$ with $D_x=D_y=1$,  recall the notations and setup \eqref{eq: 2-layer NN t} in Example \ref{example: 2-layer}, the $N$-width 2-layer NN admits the following  
\begin{align}
        \label{eq: 2-layer NN t x}
        f(x,\btheta_t^N) 
        =\frac{1}{N} \sum_{i=1}^N v_{t,i} \psi \big(  {u}_{t,i} x \big) ,
        \quad 
        \mu_t^N := \frac{1}{N} \sum_{i=1}^N \delta_{v_{t,i}, u_{t,i}  }
\end{align}

If we further consider the square loss function $L: \bR^2 \rightarrow \bR $ with $L(x,y)=(y-x)^2/2$, the updates of the weights for  given $x,y$ are

\begin{align}
    \label{eq: dv in 2l}
    \Delta  {v}_{t,i} 
    \Big|_{x,y} 
    &= -      \eta N \frac{ \partial \cL }{\partial v_i} \Big|_{x,y} 
    = \eta\Big(  y - \frac{1}{N} \sum_{i=1}^N v_{t,i} \psi \big( {u}_{t,i} x   \big) \Big) \psi(u_{t,i} x)
     , 
    \\
    \label{eq: du in 2l}
    \Delta  {u}_{t,i}
    \Big|_{x,y} 
    &= -      \eta N \frac{ \partial \cL }{\partial u_i} \Big|_{x,y} 
    = \eta\Big(  y - \frac{1}{N} \sum_{i=1}^N v_{t,i} \psi \big( {u}_{t,i} x   \big) \Big) v_{t,i} \psi'(u_{t,i} x)   x
    ,
\end{align}
where the scalar on the learning rate $\eta\in \bR^+$  with $\{v, u\}$ are $N$.  
Taking some proper test function $\varphi: \bR^2\rightarrow \bR$, for the pair of random variables $(u_t,v_t) \sim \mu_t $ in the limitation of \eqref{eq: 2-layer NN t x} with $N\rightarrow \infty$, where we apply the stochastic gradient descent updates via \eqref{eq: dv in 2l}-\eqref{eq: du in 2l} as in  \cite{mf-2022-path2}, we have  

\begin{align}
    \notag 
    \bE[ &\varphi(u_t,v_t)  ]
    = \int_{\bR^2}  \varphi(u,v) \mu_t (\dd u, \dd v)
    = 
    \bE[ \varphi(u_0,v_0)  ]
    + \int_{0}^t
        \frac{ \dd     \varphi(u_s,v_s)   }{ \dd s } \dd s 
    \\\notag 
    &
    = \bE[ \varphi(u_0,v_0)  ]
    + \int_{0}^t \Big( 
    \frac{\partial \varphi}{ \partial v} \frac{\dd v }{ \dd s} 
    + 
    \frac{\partial \varphi}{ \partial u} \frac{\dd u }{ \dd s} 
    \Big)\dd s
     \\
    \label{eq: dynamic 2 l}
    &
    = \bE[ \varphi(u_0,v_0)  ]
    +
      \int_{0}^t\Big( 
    \int_{ X \times Y}  \eta \big\langle  y-   v \psi(  u x ), \mu_s  \big\rangle
    \Big\langle   
    \frac{\partial \varphi}{ \partial v}  \psi(ux)  
        +
    \frac{\partial \varphi}{ \partial u} ( v \psi'(ux)x  )
    , \mu_s   
    \Big\rangle
    ~
    \pi(  \dd x , \dd y )
    \Big)\dd s,  
\end{align}
where the second measure integration with $\mu_s$ in \eqref{eq: dynamic 2 l} correspond to the gradient descent in \eqref{eq: dv in 2l}-\eqref{eq: du in 2l}.   We can then show the corresponding Fokker-Planck equation with respect to (w.r.t)  gradient flow in $\cP(\bR^2)$  
endowed with the Wasserstein metric:

\begin{align}
    \notag 
    \frac{\partial \mu_t }{ \partial t}
   & =  -\eta  \nabla      \bigg( 
    \int_{ X \times Y}  
    \big\langle  y-   v \psi(  u x ), \mu_t  \big\rangle  
     \cdot \mu_t      
     \nabla   \big( v \psi(ux) \big) 
    \pi(  \dd x , \dd y ) ~\bigg) 
    \\
    & = 
     -\eta       
    \int_{ X \times Y}  
    \big\langle  y-   v \psi(  u x ), \mu_t  \big\rangle  
    \nabla   \Big(
    \mu_t   
     ~ \nabla   \big( v \psi(ux) \big) 
       \Big)
    ~
    \pi(  \dd x , \dd y ) ,  
\end{align}
where  $\nabla   f =  \frac{\partial f }{ \partial u} + \frac{\partial f }{ \partial v}$ denotes the divergence operator.

\subsubsection{ Sampling   from the 2-layer NN}
\label{sec: sample from 2 l }
Based on the theoretical result that the empirical measure $\mu_t^N$ shall converges to some  limit measure $\mu_t$ weakly \cite{2018-mf-1,2019-mf-lln}, we can then reproduce a different width size $\hat{N}$-MFNN $f^{rep}$ with the new sample $\btheta^{rep}\in \bR^{ 2\hat{N}  }$ such that 

\begin{align}
        \label{eq: 2-layer NN t x22}
        f^{rep}(x,\btheta_t^{rep})   =\frac{1}{\hat{N} } \sum_{i=1}^{\hat{N}}v^{rep}_{t,i} \psi \big( {u}^{rep}_{t,i} x \big) .
\end{align}
We first train an $N$-size NN to get $N$-different samples of the weights via the proper couples $ \{ {u},v\}_{i=1}^N $, note that there are $2N$ different values for the $\bR^2$- distribution, however, the couple via the index $i$ in \eqref{eq: 2-layer NN t x22} leads to $N$-samples of the joint distribution instead of $N^2$-samples i.e  

\begin{align*}
\text{let} \quad &
    \mu_t^N := \frac{1}{N} \sum_{i=1}^N \delta_{v_{t,i}, u_{t,i}  },\quad 
    \nu_t^N := \frac{1}{N^2} \sum_{i,j=1}^N \delta_{v_{t,i}, u_{t,j}  },
    ~
    \text{but} ~ 
    \lim_{N\rightarrow \infty} \mu_t^N \neq \lim_{N\rightarrow \infty} \nu_t^N. 
\end{align*}
Here, the different empirical measures   $\mu_t^N$ and $\nu_t^N$ correspond to two different points of view for the MFNN dynamic:  the dependent dynamic of  $\mu_t^N$, and the independent type using $\nu_t^N$. The proper weight transfer is developed on  $\mu_t^N$ instead of $\nu_t^N$, which is supported by the theories in \cite{2018-mf-1,2019-mf-lln}.  In the later sections, we provide more insights into the importance of the dependence across different weights, and how to properly deal with them by developing proper measure structure.

Based on a  trained MFNN, we can develop the empirical measure $\mu_t^N$ with  $N$-samples of $\lbtheta_{t} \in \bR^{2}$, and we can then apply different strategies to generate new samples. We provide the following moment matching method, for a target sample size $\hat{N}$, we can choose $\btheta_t^{rep} \in \bR^{2\hat{N}} $ by minimizing the following 

\begin{align*}
    \cL^{dis}_p( \btheta_t, \btheta_t^{rep}   )&:= \sum_{q=1}^p\Big(\Big|  
    \frac{1}{N} \sum_{i=1}^N |\lbtheta_{t,i}|^q  - \frac{1}{\hat{N}}\sum_{j=1}^{\hat{N}}     |\lbtheta^{rep}_{t,j}|^q   \Big|   \Big)
    \\
    &=\sum_{q=1}^p \Big| ~
    \frac{1}{N\hat{N}} \sum_{i=1}^N \sum_{j=1}^{\hat{N}} \big(  |\lbtheta_{t,i}|^q-|\lbtheta^{rep}_{t,j}|^q      \big)\Big|
    \\
    &\leq  \sum_{q=1}^p
    \frac{1}{N\hat{N}} \sum_{i=1}^N \sum_{j=1}^{\hat{N}} \Big(  \big| ~ |\lbtheta_{t,i}|^q-|\lbtheta^{rep}_{t,j}|^q   \big|   \Big),
\end{align*}
for some positive integer $p$, this method is just matching different $q$-moment for $1\leq q \leq p$, more sampling methods are provided in the last section.

\subsubsection{ Analysis of a  5-layer MFNN}
\label{sec: mlp ffn case}
In this section, we consider the multi-layer fully-connected MFNN cases with the dataset pairs $\{\bx,\by\} \sim \pi$ for some distribution $\pi \in \cP_2(\bR^{D_x+D_y})$ with $D_x=D_y=1$.

Notes that the settings in this section can easily adapt to more general cases, e.g. partially-connected cases and the cases with $D_x,D_y>1$,  
we shall walk through a simplified 5-layer NN to show some key observations and the ideas of weight transfer. 
The idea of analysing a 5-layer NN instead of other amounts has been discussed in \cite{nguyen2023rigorous}, the first and the last layers are special layers connecting the input $x$ and the output $f(x)$ where the corresponding weights have only one dimension connected with the changeable width $N$ of the NN, the second and the second last layers (for $L\geq 3$) are directly connected with the special layers, and there exists at least one layer for $L\geq 5$ which connected with roughly $N\times N$ shape like matrix weights which is important for the analysis.

We consider the 5-layer NN without bias of the following form

\begin{align}
    \label{eq: mlp nn function - 5-layer 1d}
        f(\bx,\btheta)  =
        \frac{1}{N_4} \sum_{\gamma_4=1}^{N_4}v_{\gamma_4}
        \psi_4
        \Bigg(    \frac{1}{N_3} \sum_{{\gamma_3}=1}^{N_3}    w_{{\gamma_4},{\gamma_3}}^{(3)}  \psi_3
        \bigg( \frac{1}{N_2} \sum_{ {\gamma_2}=1}^{N_2}    w^{(2)} _{{\gamma_3},{\gamma_2}} \psi_2  \Big(    
        \frac{1}{N_1} \sum_{ {\gamma_1}=1}^{N_1}    w^{(1)} _{{\gamma_2},{\gamma_1}} \psi_1
        \big(   {u}_{\gamma_1} x  \big)
        \Big) 
        \bigg )   ~ \Bigg),
\end{align}
where $\bR^{N_1+N_1N_2+N_2N_3+N_3N_4+N_4}\ni \btheta:=(\btheta_{1,1,1,1},\dots,\btheta_{N_1,N_2,N_3,N_4})$, with $\bR^{5}\ni \btheta_{\gamma_1,\gamma_2,\gamma_3,\gamma_4}=( {u}_{\gamma_1},w^{(1)} _{{\gamma_2},{\gamma_1}},$ $w^{(2)} _{{\gamma_3},{\gamma_2}},w_{{\gamma_4},{\gamma_3}}^{(3)} ,v_{\gamma_4})$ and $\psi_1,\psi_2,\psi_3,\psi_4:\bR\rightarrow\bR$ are activation functions, $N_1,N_2,N_3,N_4$ are the width choices for different layers, and are proportional to  the general width $N$, one can set  $N_1=N_2=N_3=N_4=N$ for simplicity.

\medskip 
\medskip 
\medskip 

\textbf{ We follow the receipt to analysis the NN weights via the approach in \cite{mf-2022-path2}: }

\begin{enumerate}
    \item Write down the NN function $f$ for finite width $N$ form,  e.g \eqref{eq: mlp nn function - 5-layer 1d}. Here, we only consider the forward dynamic, the weights can be independent initially, while after training, the weights can be correlated, thus we can not reach a conclusion just by the forward form \eqref{eq: mlp nn function - 5-layer 1d}. 
    \item Write down the updated dynamic (mostly the first-order gradient descent based method) of the weights given data pair $\{x,y\}$, keeping the $\gamma$ index for better checking,  which helps to check the dependency over different weights within the same layer and across other layers. At this stage, we try to build the full empirical measure dynamic (an ansatz), and notice that only the training step changes the dependence (i.e. independent initialization can be correlated after one step of training).   
    \item Once we build our ansatz for the empirical measure of the weights, we further write down the limit equations. We then check if the behaviour of the NN function $f$ admits a closed dynamic (i.e. the evolution of the guessed measure only depends on the guessed measure, check \cite[Section 4.3]{mf-2022-path2} for the challenges).
\end{enumerate}

By following the 3 steps above, we propose the RC ansatz which has a similar measure structure to  \cite{nguyen2023rigorous}, under which the measure dynamic is closed. We provide more discussions as we go through the example below and show how to extend it to more general NN structures. We refer to the experiment section for numerical validation and some interesting findings.

\subsubsection{  Updates of the weights in the 5-layer MFNN  }
\label{section: mlp 5-layer}
Following the analysis receipt, 
we proceed to show the following first-order partial derivatives for \eqref{eq: mlp nn function - 5-layer 1d} which is important to the later analysis. For any $x\in \bR$ and $\btheta^N \in \bR^{N_1+N_1N_2+N_2N_3+N_3N_4+N_4}$, we have

\begin{align}
    \label{eq: 5layer partial v}
    \frac{ \partial f(\bx,\btheta^N)  }{ \partial   {v}_{{\color{blue} \mathbf{i} }} } 
    &=  \frac{1}{N_4}
    \psi_4
        \Bigg(    \frac{1}{N_3} \sum_{{\gamma_3}=1}^{N_3}    w_{{\color{blue} \mathbf{i} },{\gamma_3}}^{(3)}  \psi_3
        \bigg( \frac{1}{N_2} \sum_{ {\gamma_2}=1}^{N_2}    w^{(2)} _{{\gamma_3},{\gamma_2}} \psi_2  \Big(    
        \frac{1}{N_1} \sum_{ {\gamma_1}=1}^{N_1}    w^{(1)} _{{\gamma_2},{\gamma_1}} \psi_1
        \big(   {u}_{\gamma_1} x  \big)
        \Big) 
        \bigg )   ~ \Bigg) 
    , 
     \\
    \notag 
     \frac{ \partial f(\bx,\btheta^N)  }{ \partial   {w}_{ {\color{blue} \mathbf{i} },{\color{red} \mathbf{j} }}^{(3)} } 
    &= \frac{1}{N_4}
          {v}_{ {\color{blue} \mathbf{i} }}\psi_4'
        \Bigg(    \frac{1}{N_3} \sum_{{\gamma_3}=1}^{N_3}    w_{{\color{blue} \mathbf{i} },{\gamma_3}}^{(3)}  \psi_3
        \bigg( \frac{1}{N_2} \sum_{ {\gamma_2}=1}^{N_2}    w^{(2)} _{{\gamma_3},{\gamma_2}} \psi_2  \Big(    
        \frac{1}{N_1} \sum_{ {\gamma_1}=1}^{N_1}    w^{(1)} _{{\gamma_2},{\gamma_1}} \psi_1
        \big(   {u}_{\gamma_1} x  \big)
        \Big) 
        \bigg )   ~ \Bigg) 
    \\\notag 
    &\quad  \cdot  \frac{1}{N_3}
        \psi_3
        \bigg( \frac{1}{N_2} \sum_{ {\gamma_2}=1}^{N_2}    w^{(2)} _{{{\color{red} \mathbf{j} }},{\gamma_2}} \psi_2  \Big(    
        \frac{1}{N_1} \sum_{ {\gamma_1}=1}^{N_1}    w^{(1)} _{{\gamma_2},{\gamma_1}} \psi_1
        \big(   {u}_{\gamma_1} x  \big)
        \Big) 
        \bigg )
     ,
     \\
    \notag 
    \frac{ \partial f(\bx,\btheta^N)  }{ \partial   {w}_{ {\color{blue} \mathbf{i} },{\color{red} \mathbf{j} }}^{(2)} }
    &= 
     \frac{1}{N_4} \sum_{\gamma_4=1}^{N_4} \Bigg[   
     {v}_{ {\gamma_4}}\psi_4'
        \Bigg(    \frac{1}{N_3} \sum_{{\gamma_3}=1}^{N_3}    w_{{\gamma_4},{\gamma_3}}^{(3)}  \psi_3
        \bigg( \frac{1}{N_2} \sum_{ {\gamma_2}=1}^{N_2}    w^{(2)} _{{\gamma_3},{\gamma_2}} \psi_2  \Big(    
        \frac{1}{N_1} \sum_{ {\gamma_1}=1}^{N_1}    w^{(1)} _{{\gamma_2},{\gamma_1}} \psi_1
        \big(   {u}_{\gamma_1} x  \big)
        \Big) 
        \bigg )   ~ \Bigg) 
    \\
    \notag 
    &\quad \cdot 
    \frac{1}{N_3}
    w_{{\gamma_4},{\color{blue} \mathbf{i} }}^{(3)}
    \Bigg]
        \psi_3' 
        \bigg( \frac{1}{N_2} \sum_{ {\gamma_2}=1}^{N_2}    w^{(2)} _{{\color{blue} \mathbf{i} },{\gamma_2}} \psi_2  \Big(    
        \frac{1}{N_1} \sum_{ {\gamma_1}=1}^{N_1}    w^{(1)} _{{\gamma_2},{\gamma_1}} \psi_1
        \big(   {u}_{\gamma_1} x  \big)
        \Big) 
        \bigg )
        \cdot \frac{1}{N_2}
        \psi_2  \Big(    
        \frac{1}{N_1} \sum_{ {\gamma_1}=1}^{N_1}    w^{(1)} _{{{\color{red} \mathbf{j} }},{\gamma_1}} \psi_1
        \big(   {u}_{\gamma_1} x  \big)
        \Big) 
        ,
    \\
    \notag 
    \frac{ \partial f(\bx,\btheta^N)  }{ \partial   {w}_{ {\color{blue} \mathbf{i} },{\color{red} \mathbf{j} }}^{(1)} } 
    &=       
     \frac{1}{N_4} \sum_{\gamma_4=1}^{N_4} \Bigg[   
     {v}_{ {\gamma_4}}\psi_4'
        \Bigg(    \frac{1}{N_3} \sum_{{\gamma_3}=1}^{N_3}    w_{{\gamma_4},{\gamma_3}}^{(3)}  \psi_3
        \bigg( \frac{1}{N_2} \sum_{ {\gamma_2}=1}^{N_2}    w^{(2)} _{{\gamma_3},{\gamma_2}} \psi_2  \Big(    
        \frac{1}{N_1} \sum_{ {\gamma_1}=1}^{N_1}    w^{(1)} _{{\gamma_2},{\gamma_1}} \psi_1
        \big(   {u}_{\gamma_1} x  \big)
        \Big) 
        \bigg )   ~ \Bigg) 
    \\
    \notag 
    &\quad \quad \cdot
     \frac{1}{N_3} \sum_{\gamma_3=1}^{N_3}
     \bigg[   
    w_{{\gamma_4},{\gamma_3}}^{(3)}
        \psi_3' 
        \bigg( \frac{1}{N_2} \sum_{ {\gamma_2}=1}^{N_2}    w^{(2)} _{{\gamma_3},{\gamma_2}} \psi_2  \Big(    
        \frac{1}{N_1} \sum_{ {\gamma_1}=1}^{N_1}    w^{(1)} _{{\gamma_2},{\gamma_1}} \psi_1
        \big(   {u}_{\gamma_1} x  \big)
        \Big) 
        \bigg )
    \\
    \notag 
    &\quad \qquad 
    \cdot \frac{1}{N_2}
    {w}_{ \gamma_3,{\color{blue} \mathbf{i} }}^{(2)}
    \bigg]\Bigg]
    \psi_2' \Big(    
    \frac{1}{N_1} \sum_{ {\gamma_1}=1}^{N_1}    w^{(1)} _{{\color{blue} \mathbf{i} },{\gamma_1}} \psi_1
    \big(   {u}_{\gamma_1} x  \big)
    \Big) 
    \cdot \frac{1}{N_1}
    \psi_1
    \big(   {u}_{{\color{red} \mathbf{j} }} x  \big) 
    ,  
     \\
    \notag 
    \frac{ \partial f(\bx,\btheta^N)  }{ \partial  {u}_{ {\color{red} \mathbf{j} }}^{(1)}   }  
    &=   
     \frac{1}{N_4} \sum_{\gamma_4=1}^{N_4} \Bigg[   
     {v}_{ {\gamma_4}}\psi_4'
        \Bigg(    \frac{1}{N_3} \sum_{{\gamma_3}=1}^{N_3}    w_{{\gamma_4},{\gamma_3}}^{(3)}  \psi_3
        \bigg( \frac{1}{N_2} \sum_{ {\gamma_2}=1}^{N_2}    w^{(2)} _{{\gamma_3},{\gamma_2}} \psi_2  \Big(    
        \frac{1}{N_1} \sum_{ {\gamma_1}=1}^{N_1}    w^{(1)} _{{\gamma_2},{\gamma_1}} \psi_1
        \big(   {u}_{\gamma_1} x  \big)
        \Big) 
        \bigg )   ~ \Bigg) 
    \\
    \notag 
    &\quad \quad \cdot
     \frac{1}{N_3} \sum_{\gamma_3=1}^{N_3}
     \bigg[   
    w_{{\gamma_4},{\gamma_3}}^{(3)}
        \psi_3' 
        \bigg( \frac{1}{N_2} \sum_{ {\gamma_2}=1}^{N_2}    w^{(2)} _{{\gamma_3},{\gamma_2}} \psi_2  \Big(    
        \frac{1}{N_1} \sum_{ {\gamma_1}=1}^{N_1}    w^{(1)} _{{\gamma_2},{\gamma_1}} \psi_1
        \big(   {u}_{\gamma_1} x  \big)
        \Big) 
        \bigg )
    \\
    \notag 
    &\quad \qquad 
    \cdot
    \frac{1}{N_2} \sum_{\gamma_2=1}^{N_2}  \Big[   
    {w}_{ \gamma_3,\gamma_2}^{(2)}
    \psi_2' \Big(    
    \frac{1}{N_1} \sum_{ {\gamma_1}=1}^{N_1}    w^{(1)} _{{\gamma_2},{\gamma_1}} \psi_1
    \big(   {u}_{\gamma_1} x  \big)
    \Big) 
    \cdot \frac{1}{N_1} 
      w^{(1)} _{{\gamma_2},{{\color{red} \mathbf{j} }}} \psi_1'
    \big(   {u}_{{\color{red} \mathbf{j} }} x  \big)  x 
    \Big]  ~\bigg]~\Bigg],
\end{align}
where we highlight the lower indices $\mathbf{i},\mathbf{j}$ for better tracking the dependency across different layers. Notice that these partial derivatives are important components in the update processes of the weights. If we  further consider the square loss function $L: \bR^2 \rightarrow \bR $ with $L(x,y)=(y-x)^2/2$ and taking $\eta\in \bR$ independent of $N$ as the learning rate with proper scalar described later, for a given data pair $x,y$, we have

\begin{align}
    \label{eq: delta v eg eq}
    \Delta {v}_{  {\color{blue} \mathbf{i} }}  \Big|_{x,y} 
    &=  
    \eta \Big(  y - f(x,\btheta^N) \Big) 
    \psi_4
        \Bigg(    \frac{1}{N_3} \sum_{{\gamma_3}=1}^{N_3}    w_{{\color{blue} \mathbf{i} },{\gamma_3}}^{(3)}  \psi_3
        \bigg( \frac{1}{N_2} \sum_{ {\gamma_2}=1}^{N_2}    w^{(2)} _{{\gamma_3},{\gamma_2}} \psi_2  \Big(    
        \frac{1}{N_1} \sum_{ {\gamma_1}=1}^{N_1}    w^{(1)} _{{\gamma_2},{\gamma_1}} \psi_1
        \big(   {u}_{\gamma_1} x  \big)
        \Big) 
        \bigg )   ~ \Bigg) 
    , 
     \\
    \notag 
    \Delta {w}_{  {\color{blue} \mathbf{i} },{\color{red} \mathbf{j} }}^{(3)}   \Big|_{x,y} 
    &= 
     \eta \Big(  y - f(x,\btheta^N) \Big)      {v}_{  {\color{blue} \mathbf{i} }}\psi_4'
        \Bigg(    \frac{1}{N_3} \sum_{{\gamma_3}=1}^{N_3}    w_{{\color{blue} \mathbf{i} },{\gamma_3}}^{(3)}  \psi_3
        \bigg( \frac{1}{N_2} \sum_{ {\gamma_2}=1}^{N_2}    w^{(2)} _{{\gamma_3},{\gamma_2}} \psi_2  \Big(    
        \frac{1}{N_1} \sum_{ {\gamma_1}=1}^{N_1}    w^{(1)} _{{\gamma_2},{\gamma_1}} \psi_1
        \big(   {u}_{\gamma_1} x  \big)
        \Big) 
        \bigg )   ~ \Bigg) 
    \\
    \label{eq: delta w3 eg eq}
    &\quad \quad \cdot 
        \psi_3
        \bigg( \frac{1}{N_2} \sum_{ {\gamma_2}=1}^{N_2}    w^{(2)} _{{{\color{red} \mathbf{j} }},{\gamma_2}} \psi_2  \Big(    
        \frac{1}{N_1} \sum_{ {\gamma_1}=1}^{N_1}    w^{(1)} _{{\gamma_2},{\gamma_1}} \psi_1
        \big(   {u}_{\gamma_1} x  \big)
        \Big) 
        \bigg )
     ,
     \\
    \notag 
    \Delta {w}_{  {\color{blue} \mathbf{i} },{\color{red} \mathbf{j} }}^{(2)}   \Big|_{x,y} 
    &= 
     \eta \Big(  y - f(x,\btheta^N) \Big)     \cdot
     \frac{1}{N_4} \sum_{\gamma_4=1}^{N_4} \Bigg[   
     {v}_{  {\gamma_4}}\psi_4'
        \Bigg(    \frac{1}{N_3} \sum_{{\gamma_3}=1}^{N_3}    w_{{\gamma_4},{\gamma_3}}^{(3)}  \psi_3
        \bigg( \frac{1}{N_2} \sum_{ {\gamma_2}=1}^{N_2}  w^{(2)} _{{\gamma_3},{\gamma_2}}   
    \\
    \notag 
    &\quad \quad\cdot 
      \psi_2  \Big(    
        \frac{1}{N_1} \sum_{ {\gamma_1}=1}^{N_1}    w^{(1)} _{{\gamma_2},{\gamma_1}} \psi_1
        \big(   {u}_{\gamma_1} x  \big)
        \Big) 
        \bigg )   ~ \Bigg) 
    w_{{\gamma_4},{\color{blue} \mathbf{i} }}^{(3)}
    \Bigg]
        \psi_3' 
        \bigg( \frac{1}{N_2} \sum_{ {\gamma_2}=1}^{N_2}    w^{(2)} _{{\color{blue} \mathbf{i} },{\gamma_2}} 
    \\
    \label{eq: delta w2 eg eq}
    &\quad \quad\cdot 
    \psi_2  \Big(    
        \frac{1}{N_1} \sum_{ {\gamma_1}=1}^{N_1}    w^{(1)} _{{\gamma_2},{\gamma_1}} \psi_1
        \big(   {u}_{\gamma_1} x  \big)
        \Big) 
        \bigg )
        \cdot 
        \psi_2  \Big(    
        \frac{1}{N_1} \sum_{ {\gamma_1}=1}^{N_1}    w^{(1)} _{{{\color{red} \mathbf{j} }},{\gamma_1}} \psi_1
        \big(   {u}_{\gamma_1} x  \big)
        \Big) 
        ,
    \\
    \notag 
    \Delta {w}_{  {\color{blue} \mathbf{i} },{\color{red} \mathbf{j} }}^{(1)}   \Big|_{x,y} 
    &= 
     \eta \Big(  y - f(x,\btheta^N) \Big)     \cdot
     \frac{1}{N_4} \sum_{\gamma_4=1}^{N_4} \Bigg[   
     {v}_{  {\gamma_4}}\psi_4'
        \Bigg(    \frac{1}{N_3} \sum_{{\gamma_3}=1}^{N_3}    w_{{\gamma_4},{\gamma_3}}^{(3)}  \psi_3
        \bigg( \frac{1}{N_2} \sum_{ {\gamma_2}=1}^{N_2}    w^{(2)} _{{\gamma_3},{\gamma_2}} 
    \\
    \notag 
    &\quad \quad \cdot
    \psi_2  \Big(    
        \frac{1}{N_1} \sum_{ {\gamma_1}=1}^{N_1}    w^{(1)} _{{\gamma_2},{\gamma_1}} \psi_1
        \big(   {u}_{\gamma_1} x  \big)
        \Big) 
        \bigg )   ~ \Bigg) \cdot
     \frac{1}{N_3} \sum_{\gamma_3=1}^{N_3}
     \bigg[   
    w_{{\gamma_4},{\gamma_3}}^{(3)}
        \psi_3' 
        \bigg( \frac{1}{N_2} \sum_{ {\gamma_2}=1}^{N_2}    w^{(2)} _{{\gamma_3},{\gamma_2}} 
    \\
    \label{eq: delta w1 eg eq}
    &\quad \quad 
    \cdot 
    \psi_2  \Big(    
        \frac{1}{N_1} \sum_{ {\gamma_1}=1}^{N_1}    w^{(1)} _{{\gamma_2},{\gamma_1}} \psi_1
        \big(   {u}_{\gamma_1} x  \big)
        \Big) 
        \bigg )
    {w}_{  \gamma_3,{\color{blue} \mathbf{i} }}^{(2)}
    \bigg]\Bigg]
    \psi_2' \Big(    
    \frac{1}{N_1} \sum_{ {\gamma_1}=1}^{N_1}    w^{(1)} _{{\color{blue} \mathbf{i} },{\gamma_1}} \psi_1
    \big(   {u}_{\gamma_1} x  \big)
    \Big) 
    \cdot
    \psi_1
    \big(   {u}_{{\color{red} \mathbf{j} }} x  \big) 
    ,  
     \\
        \notag 
    \Delta {u}_{  {\color{red} \mathbf{j} }}^{(1)}   \Big|_{x,y} 
    &= 
     \eta \Big(  y - f(x,\btheta^N) \Big)   \cdot  
     \frac{1}{N_4} \sum_{\gamma_4=1}^{N_4} \Bigg[   
     {v}_{  {\gamma_4}}\psi_4'
        \Bigg(    \frac{1}{N_3} \sum_{{\gamma_3}=1}^{N_3}    w_{{\gamma_4},{\gamma_3}}^{(3)}  \psi_3
        \bigg( \frac{1}{N_2} \sum_{ {\gamma_2}=1}^{N_2}    w^{(2)} _{{\gamma_3},{\gamma_2}}  
    \\
    \notag 
    & \cdot
    \psi_2  \Big(    
        \frac{1}{N_1} \sum_{ {\gamma_1}=1}^{N_1}    w^{(1)} _{{\gamma_2},{\gamma_1}} \psi_1
        \big(   {u}_{\gamma_1} x  \big)
        \Big) 
        \bigg )   ~ \Bigg) \cdot
     \frac{1}{N_3} \sum_{\gamma_3=1}^{N_3}
     \bigg[   
    w_{{\gamma_4},{\gamma_3}}^{(3)}
        \psi_3' 
        \bigg( \frac{1}{N_2} \sum_{ {\gamma_2}=1}^{N_2}    w^{(2)} _{{\gamma_3},{\gamma_2}} \psi_2  \Big(    
        \frac{1}{N_1} \sum_{ {\gamma_1}=1}^{N_1}    w^{(1)} _{{\gamma_2},{\gamma_1}} 
    \\
    \label{eq: delta  eg eq}
    &
    \cdot
    \psi_1
        \big(   {u}_{\gamma_1} x  \big)
        \Big) 
        \bigg )\cdot
    \frac{1}{N_2} \sum_{\gamma_2=1}^{N_2}  \Big[   
    {w}_{  \gamma_3,\gamma_2}^{(2)}
    \psi_2' \Big(    
    \frac{1}{N_1} \sum_{ {\gamma_1}=1}^{N_1}    w^{(1)} _{{\gamma_2},{\gamma_1}} \psi_1
    \big(   {u}_{\gamma_1} x  \big)
    \Big) 
    \cdot
      w^{(1)} _{{\gamma_2},{{\color{red} \mathbf{j} }}} \psi_1'
    \big(   {u}_{{\color{red} \mathbf{j} }} x  \big)  x 
    \Big]  ~\bigg]~\Bigg],
\end{align}
where the scalar on the learning rate $\eta$  with $\{v,w^{(3)},w^{(2)},w^{(1)},u\}$ are $\{N_4,N_4N_3, N_3N_2,N_2N_1,N_1\}$. The key idea of the scalar here is to guarantee that for any single weight $\theta\in \bR$ from $ \theta \in \btheta^N$ such that $\theta \sim \cO(1)$, the one step updates $\{\Delta v,\Delta w^{(3)},\Delta w^{(2)}$,  $\Delta w^{(1)},\Delta u\}$ are at most $\cO(1)$.  Notice that the update order on $N$ for $\theta \in \btheta^N$ might be smaller than $\cO(1)$,  \cite[Section E]{tp4} has addressed this situation, and it is suggested to take none-zero mean type initialization  as in \cite[Section 4.2.1]{nguyen2019mfexperiment}, we  suggest a new initialization method called "RC-initialization" in Section \ref{sec: rc ini}, we leave the discussion to a  later section after the analysis of the dynamic for \eqref{eq: mlp nn function - 5-layer 1d}. 

Take a careful look, for any properly chosen $i,j$ with  $\ell \notin \{i,j\}$, we use below the dot notation in the lower index to denote any choices of indices at the position, we have the following observations

\begin{align*}
    \frac{ \partial \Delta {v}_ { i}}{\partial v_ { \cdot}  }& \sim \cO( \frac{1}{N_4}  ),
    \quad 
     \frac{ \partial \Delta {v}_ { i}}{\partial w_ { {i},\cdot}^{(3)} } \sim \cO( \frac{1}{N_3}  ),
     \quad 
     \frac{ \partial \Delta {v}_ { i}}{\partial w_ {  {\ell},\cdot }^{(3)} } \sim \cO( \frac{1}{N_3N_4}  ), 
     \\\\
     \frac{ \partial \Delta  w_ { {i},{j}}^{(3)}}{\partial v_ { {i}} } &\sim \cO( 1 ),
      \quad 
      \frac{ \partial \Delta  w_ { {i},{j}}^{(3)}}{\partial v_ { {\ell}} } \sim \cO( \frac{1}{N_4} ),
      \quad
    \frac{ \partial \Delta  w_ { {i},{j}}^{(3)}}{\partial w_ { {i},\cdot}^{(3)} } \sim \cO( \frac{1}{N_3}  ),
     \quad 
     \frac{ \partial \Delta  w_ { {i},{j}}^{(3)}}{\partial w_ { {\ell},\cdot}^{(3)} } \sim \cO( \frac{1}{N_3N_4}  ),
     \\
     \frac{ \partial \Delta  w_ { {i},{j}}^{(3)}}{\partial w_ { {j},\cdot}^{(2)} } 
     &\sim \cO( \frac{1}{N_2}  ),
     \quad 
     \frac{ \partial \Delta  w_ { {i},{j}}^{(3)}}{\partial w_ { {\ell},\cdot}^{(2)} } 
     \sim \cO( \frac{1}{N_2N_3}  ),
     \\\\
     \frac{ \partial \Delta  w_ { {i},{j}}^{(2)}}{\partial  w_ { \cdot,{i}}^{(3)} } &\sim \cO( \frac{1}{N_4} ),
      \quad 
       \frac{ \partial \Delta  w_ { {i},{j}}^{(2)}}{\partial  w_ { \cdot,{\ell}}^{(3)}} \sim \cO( \frac{1}{N_3N_4} ),
      \quad
    \frac{ \partial \Delta  w_ { {i},{j}}^{(2)}}{\partial w_ { {i},\cdot}^{(2)} } \sim \cO( \frac{1}{N_2}  ),
     \quad 
     \frac{ \partial \Delta  w_ { {i},{j}}^{(2)}}{\partial w_ { {\ell},\cdot}^{(2)} } \sim \cO( \frac{1}{N_2N_3}  ),
     \\
     \frac{ \partial \Delta  w_ { {i},{j}}^{(2)}}{\partial w_ { {j},\cdot}^{(1)} } 
     &\sim \cO( \frac{1}{N_1}  ),
     \quad 
     \frac{ \partial \Delta  w_ { {i},{j}}^{(2)}}{\partial w_ { {\ell},\cdot}^{(1)} } 
     \sim \cO( \frac{1}{N_1N_2}  ),
      \\\\
     \frac{ \partial \Delta  w_ { {i},{j}}^{(1)}}{\partial  w_ { \cdot,{i}}^{(2)} } &\sim \cO( \frac{1}{N_3} ),
      \quad 
       \frac{ \partial \Delta  w_ { {i},{j}}^{(1)}}{\partial  w_ { \cdot,{\ell}}^{(2)}} \sim \cO( \frac{1}{N_2N_3} ),
      \quad
    \frac{ \partial \Delta  w_ { {i},{j}}^{(1)}}{\partial w_ { {i},\cdot}^{(1)} } \sim \cO( \frac{1}{N_1}  ),
     \quad 
     \frac{ \partial \Delta  w_ { {i},{j}}^{(1)}}{\partial w_ { {\ell},\cdot}^{(1)} } \sim \cO( \frac{1}{N_1N_2}  ),
     \\
     \frac{ \partial \Delta  w_ { {i},{j}}^{(1)}}{\partial u_ { {j}} } 
     &\sim \cO( 1  ),
     \quad 
     \frac{ \partial \Delta  w_ { {i},{j}}^{(1)}}{\partial u_ { {\ell}} } 
     \sim \cO( \frac{1}{N_1}  ),
     \\\\
      \frac{ \partial \Delta {u}_ { j}}{\partial u_ { j}  }& \sim \cO( 1  ),
      \quad 
       \frac{ \partial \Delta {u}_ { j}}{\partial u_ { \ell}  }  \sim \cO( \frac{1}{N_1}  ),
    \quad 
     \frac{ \partial \Delta {u}_ { j}}{\partial w_ { \cdot,{j}}^{(1)} } \sim \cO( \frac{1}{N_2}  ),
     \quad 
     \frac{ \partial \Delta {u}_ { j}}{\partial w_ { \cdot,{\ell}}^{(1)} } \sim \cO( \frac{1}{N_1N_2}  ),
\end{align*}

we can see the dependencies across weights are different, for example, $\Delta {v}_ { i}$ depends on one order higher magnitude on $w^{(3)}$ with row index $i$ than those of other row indexes, these different dependencies lead to the main difficulty to develop proper measure structure. Before showing the proper RC ansatz for the measure structure, we first introduced the following wrong ansatz to show what can be wrong.

\subsubsection{A wrong ansatz for the 5-layer MFNN}
\label{sec: wrong ansatz}

We first write down the limit equation of \eqref{eq: mlp nn function - 5-layer 1d} based on some measure of $( v,  w^{(3)},   w^{(2)}, w^{(1)} , u)_t \sim \mu_t \in \cP(\bR^5)$, notice that we introduce the time component $t$ here for  the measure evolution 

\begin{align}
    \notag 
    f(x,\mu_t)= &\int_{\bR} v \psi_4 \Bigg(  
    \int_{\bR} w^{(3)}
      \psi_3 \bigg( 
    \int_{\bR}   w^{(2)}  
    \psi_2 \Big(  \int_{\bR^2}
    w^{(1)}\psi_1( u x )
     \mu_t( v,  w^{(3)},   w^{(2)},\dd w^{(1)} ,\dd u  )
     \Big) 
    \\
    \label{eq: 5 layers int form correct}
    &
    \qquad 
    \mu_t(   v,    w^{(3)} ,\dd w^{(2)}, \cdot ,\cdot  ) \bigg)~
    \mu_t(   v,  \dd w^{(3)} ,\cdot , \cdot ,\cdot  )
    \Bigg) \mu_t( \dd v, \cdot ,\cdot , \cdot ,\cdot  ) 
\end{align}
where $    \cdot  $ notation denotes the corresponding marginal distribution. In deed, there are many different forms of the measure dependency structure here, for example, all of $( v,  w^{(3)},   w^{(2)}, w^{(1)} , u)$ are independent, or they are dependent in a sequential sense.
We choose to show the following structure
\begin{align}
    \notag 
    \mu_t( &\dd v, \dd w^{(3)}, \dd  w^{(2)}, \dd  w^{(1)} , \dd u) 
    \\
    \label{eq: 5-layer  wrong dis setup}
    & =
    \mu_t( \dd v,  \cdot ,\cdot , \cdot ,\cdot) 
    \mu_t( v,  \dd w^{(3)} ,\cdot , \cdot ,\cdot) 
    \mu_t( \cdot,  w^{(3)}, \dd w^{(2)} , \cdot ,\cdot) 
    \mu_t( \cdot,  \cdot,w^{(2)} , \dd w^{(1)}   ,\cdot) 
    \mu_t(    \cdot ,\cdot , \cdot ,w^{(1)}, \dd u), 
\end{align}
where the $\cdot$ notation denote the marginals, notice that the measure of the weights are jointly path dependent, we provide the following  notations for \eqref{eq: 5-layer  wrong dis setup}

\begin{align}
    \notag 
   \mu_t^v(v)&:=   \int_{\bR^4} \mu_t(  v,  \dd w^{(3)}, \dd  w^{(2)}, \dd  w^{(1)} , \dd u),
   \\
       \notag 
   \mu_t^{w^{(3)},v}(w^{(3)}| v )&:=  
   \frac{    
   \int_{\bR^3}  \mu_t(     v  , w^{(3)},  \dd w^{(2)}, \dd  w^{(1)} , \dd u )  
   }{      \int_{\bR^4}  \mu_t(     v  , \dd w^{(3)},  \dd  w^{(2)}, \dd  w^{(1)} , \dd u )     }
   ,
   \\
       \notag 
   \mu_t^{w^{(2)},w^{(3)}}(w^{(2)}|w^{(3)}  )&:=  
   \frac{    
   \int_{\bR^3}  \mu_t(    \dd v  , w^{(3)},   w^{(2)},  \dd w^{(1)} , \dd u )  
   }{      \int_{\bR^4}  \mu_t(    \dd v  , w^{(3)},  \dd  w^{(2)}, \dd  w^{(1)} , \dd u )     }
   ,
    \\
        \notag 
   \mu_t^{w^{(1)},w^{(2)}}(w^{(1)}|w^{(2)}   )&:=  
   \frac{    
   \int_{\bR^3}  \mu_t(    \dd v  , \dd w^{(3)},   w^{(2)},  w^{(1)} , \dd  u )  
   }{      \int_{\bR^4}  \mu_t(    \dd v  , \dd w^{(3)},   w^{(2)}, \dd  w^{(1)} , \dd  u )     }
   ,
    \\
    \label{eq: seperate meas}
   \mu_t^{u,w^{(1)}}(u |w^{(1)}  )&:=  
   \frac{    
   \int_{\bR^3}  \mu_t(    \dd v  , \dd w^{(3)},   \dd w^{(2)},  w^{(1)} ,   u )  
   }{      \int_{\bR^4} \mu_t(    \dd v  , \dd w^{(3)},   \dd w^{(2)},  w^{(1)} , \dd  u )     }
   .
\end{align}

so that $  \mu_t(   v,   w^{(3)},    w^{(2)},    w^{(1)} ,  u)= \mu_t^v(v) \times \mu_t^{w^{(3)},v}(w^{(3)}| v ) \times \mu_t^{w^{(2)},w^{(3)}}(w^{(2)}|w^{(3)}  ) \times \mu_t^{w^{(1)},w^{(2)}}(w^{(1)}|w^{(2)}   ) \times \mu_t^{u,w^{(1)}}(u |w^{(1)}  )$. 
We also introduce the following notation, given \eqref{eq: mlp nn function - 5-layer 1d} and \eqref{eq: 5-layer  wrong dis setup}, for any $\mu_t \in \cP(\bR^5)$, $t\geq 0$, using the setup in  \eqref{eq: seperate meas},   we let 

\begin{align}
    \label{eq: 5-layer NN formula xx}
    &f * \mu (x):= \int_{\bR} v \sigma_4 \Bigg(  \int_{\bR^2} w^{(3)}       
    \sigma_3 \bigg(   
    \int_{\bR} w^{(2)} 
    \cdot 
    \sigma_2 \Big(   
    \int_{\bR}\int_{\bR} w^{(1)}
    \sigma_1(  u x   )
    ~ \mu_t^{u,w^{(1)}}(\dd u |w^{(1)}  )
    \\
    &  
        \notag 
    \qquad \qquad\qquad \qquad \qquad\qquad \cdot \mu_t^{w^{(1)},w^{(2)}}(\dd w^{(1)}|w^{(2)}   )
    \Big)
    \cdot 
    ~ \mu_t^{w^{(2)},w^{(3)}}(\dd w^{(2)}|w^{(3)}  )
    \bigg)
    ~ \mu_t^{w^{(3)},v}(\dd w^{(3)}| v )
    \Bigg)
    ~ \mu_t^v(\dd v) , 
\end{align}

We will show later that this setup is not proper, but let's progress to write down the test function dynamic following the receipt in Section \ref{sec: mlp ffn case}.  So that for the stochastic sample pair $\{v,w^{(3)},w^{(2)},w^{(1)},u\}_t\sim \mu_t \in \cP(\bR^5)$ and for some proper test function $\varphi:\bR\rightarrow \bR$,  we have 

\begin{align} 
    \notag 
    &\bE[ \varphi(v_t)  ]
    =  
    \bE[ \varphi(v_0)  ]+
      \int_{0}^t\Bigg( 
    \int_{ X \times Y} \eta        \Big( y - f *\mu_s(x) \Big)   
    \\
    \notag  
    &  \cdot 
    \bigg\langle    
    \partial_x \varphi({\color{blue}v})
    \cdot  \psi_4\bigg(  
    \int_{\bR} \hat w^{(3)}
      \psi_3 \bigg( 
    \int_{\bR}  \hat w^{(2)}  
    \psi_2 \Big(  \int_{\bR}\int_{\bR}
    \hat w^{(1)}\psi_1( \hat u x ) \ 
    \mu_t^{  u,{w}^{(1)}}(\dd \hat u | \hat w^{(1)} )
     \mu_t^{  {w}^{(1)},w^{(2)}}(\dd \hat w^{(1)}| \hat w^{(2)} )
     \Big) 
    \\ 
    \label{eq: v form 5l- eq 3}
    &
    \qquad  \cdot 
     \mu_t^{  {w}^{(2)},w^{(3)}}(\dd \hat w^{(2)}| \hat w^{(3)}) \bigg)~
    \mu_t^{w^{(3)},v}(\dd \hat w^{(3)}|{\color{blue}  v })
    \bigg)  
    , ~ \mu_s^v (  {\color{blue} v  })  \bigg\rangle
    ~
    \pi(  \dd x , \dd y )
    \Bigg)\dd s,
\end{align}

where we have a contradiction to  \eqref{eq: 5layer partial v}:  
\begin{align}
    \notag 
    &\lim_{ N_1,N_2,N_3 \rightarrow \infty } 
    \psi_3
        \bigg( \frac{1}{N_2} \sum_{ {\gamma_2}=1}^{N_2}    w^{(2)} _{{\gamma_3},{\gamma_2}} \psi_2  \Big(    
        \frac{1}{N_1} \sum_{ {\gamma_1}=1}^{N_1}    w^{(1)} _{{\gamma_2},{\gamma_1}} \psi_1
        \big(   {u}_{\gamma_1} x  \big)
        \Big) 
        \bigg )
        \\
        \notag 
        &~\neq 
         \psi_3 \bigg( 
    \int_{\bR}  \hat w^{(2)}  
    \psi_2 \Big(  \int_{\bR}\int_{\bR}
    \hat w^{(1)}\psi_1( \hat u x ) \ 
    \mu_t^{  u,{w}^{(1)}}(\dd \hat u | \hat w^{(1)} )
     \mu_t^{  {w}^{(1)},w^{(2)}}(\dd \hat w^{(1)}| \hat w^{(2)} )
     \Big) 
     \mu_t^{  {w}^{(2)},w^{(3)}}(\dd \hat w^{(2)}| \hat w^{(3)}) \bigg). 
\end{align}
The $  w^{(2)} _{{\gamma_3},\cdot }$ shall dependent on the whole column $\{  w^{(3)} _{\ell,{{\gamma_3}}}  \}_{\ell =1}^{N_4}$ instead of any specific element $ w^{(3)}_{\ell,{\gamma_3}}\in \bR$, this contradiction shows not only in \eqref{eq: v form 5l- eq 3} but also in other lines (e.g \eqref{eq: 5-layer NN formula xx}) so that the measure structure in \eqref{eq: 5-layer  wrong dis setup} cannot properly describe the update dynamic of the 5-layer NN \eqref{eq: mlp nn function - 5-layer 1d} which don't admit a closed dynamic. Indeed, the difficulty of building a proper measure structure has been discussed in  \cite[Section 4.3]{mf-2022-path2}  and \cite{nguyen2023rigorous}, and other measure approaches such as “nested measures” also leads to similar issues. Without a proper measure structure, we cannot guarantee the mean filed point of view for the MFNN is proper, nor the sampling ideas, thus we develop the RC ansatz.

\subsubsection{The RC ansatz for the 5-layer MFNN}
\label{section: anstazr for 5 layers}

The problem described in Section \ref{sec: wrong ansatz} motivates us to build a new setup of measure structure, instead of viewing the middle layers components $\{w^{(\ell)}\}_{i,j}$ as a single measure, we split it into a component of 2 RVs such that for any $i\in\{i,\dots,N_{\ell +1}\},\ j\in\{i,\dots,N_\ell\}$ with $\ell\in\{1,2,3\}$,   we have 
\begin{align}
    \label{eq: emb function}
    w^{(\ell)}_{i,j} = \phi^{(\ell)}( \mathcal{R}_{i}^{\ell} , \mathcal{C}^{\ell}_{j}   ),
\end{align}
for some   function $ \phi^{(\ell)}:\bR^2\to \bR$ with random variables $ \cR_{i}^{\ell},\cC_{j}^{\ell}$, so that the 5-layer MFNN in \eqref{eq: mlp nn function - 5-layer 1d} can be viewed as sampling from $\{u,\cR^{(1)},\cC^{(1)},\cR^{(2)},\cC^{(2)},\cR^{(3)},\cC^{(3)},v   \} \in \cP(\bR^8) $ instead of    
$\{u,w^{(1)},w^{(2)},w^{(3)},v   \} \in \cP(\bR^5) $.

\begin{example}
    One possible form of  $\phi^{(\ell)}$ is  $\phi^{(\ell)}(x,y)=xy$, then $w^\ell$ can be viewed as a tensor product of $\cR^\ell$ and $\cC^\ell$, i.e., $w^\ell = \cR^\ell \otimes \cC^\ell$, other possible form is  $\phi^{(\ell)}(x,y)=x+y$. 

    Potentially, this operation can be understood as a executing a copula on the CDF's corresponding to the marginals.
\end{example}

We don't specify the exact form of $\phi^{(\ell)}$ here, since for most NN training, we update the weights $\{w\}^{(\ell)}_{i,j}$ instead of $\{ \cR,\cC \}^{\ell}_{i,j}$, and there are multiple choices for $\{ \cR,\cC \}^{\ell}_{i,j}$ and $\phi^{(\ell)}$,  where different  $ \phi^{(\ell)}$ leads to a different measure of $\{ \cR,\cC \}^{\ell}_{i,j}$. This non-uniqueness of the pairs $  \{ \phi, \mathcal{R} , \mathcal{C} \}^{\ell} $ is not a main research point of this paper and we shall leave it for future work. We also show later that there is no need to recover the measure of $\{ \cR,\cC \}^{\ell}_{i,j}$ when performing weight initialization and transfer.

This idea in \eqref{eq: emb function} is similar   to the "neuronal embedding" introduced in \cite{nguyen2023rigorous}, where they use 4 RVs instead of 8 RVs for  \eqref{eq: mlp nn function - 5-layer 1d}.  
The RC ansatz includes more R.V.s to provide higher capacity, for example, the skip connection can be dealt with more flexibly, see Section \ref{sec: discuss on other measure structure} for more details.   We show in the following that the MFNN under the RC ansatz admits a closed dynamic, and we expect the empirical measure to converge to some limit measure.

We introduce the following notation for $f * \mu$, given \eqref{eq: mlp nn function - 5-layer 1d} and \eqref{eq: emb function}, for any $\mu \in \cP(\bR^8)$,   we have 

\begin{align}
    \notag 
    f * \mu (x):=& \int_{\bR^2} v \sigma_4 \Bigg(  \int_{\bR^2} \phi^{(3)}(  \cR^{(3)} ,\cC^{(3)})       
    \sigma_3 \bigg(   
    \int_{\bR^2} \phi^{(2)}(  \cR^{(2)} ,\cC^{(2)})  
    \\
        \notag 
    &\quad 
    \sigma_2 \Big(   
    \int_{\bR^2} \phi^{(1)}(  \cR^{(1)} ,\cC^{(1)})
    \sigma_1(  u x   )
    ~ \mu( \dd u, \dd \cR^{(1)}, \cdot   )
    \Big)
    \\
        \label{eq: 5-layer NN formula}
    &\qquad 
    ~ \mu( \dd \cC^{(1)}, \dd \cR^{(2)}, \cdot   )
    \bigg)
    ~ \mu( \dd \cC^{(2)}, \dd \cR^{(3)}, \cdot   )
    \Bigg)
    ~ \mu( \dd \cC^{(3)}, \dd v, \cdot   ), 
\end{align}
where  the dot notation denotes the corresponding marginal distribution.     For the pairs $\{u,\cR^{(1)},\cC^{(1)}$, $ \cR^{(2)},\cC^{(2)},\cR^{(3)}$, $\cC^{(3)},v   \}_t \sim \mu_t\in\cP(\bR^8)$, we split it via the $\Gamma$ set of \eqref{eq: mlp nn function - 5-layer 1d} such that $\mu_t = \mu_t^{v,\cR^{(3)}}  \times
   \mu_t^{\cC^{(3)},\cR^{(2)}} \times  \mu_t^{\cC^{(2)},\cR^{(1)}} \times \mu_t^{\cC^{(1)},u}   $, with $ \mu_t^{v,\cR^{(3)}} ,
   \mu_t^{\cC^{(3)},\cR^{(2)}} , \mu_t^{\cC^{(2)},\cR^{(1)}}, \mu_t^{\cC^{(1)},u} \in \cP(\bR^2)$. 
We also denote the corresponding marginal distribution as 
\begin{align} 
       \notag 
   \mu_t^{v,\cR^{(3)}}(v,\cdot) & := \int_{\bR} \mu_t^{v,\cR^{(3)}}(v,x) \dd x,
   \quad 
   \mu_t^{v,\cR^{(3)}}( \cdot,\cR^{(3)}) := \int_{\bR} \mu_t^{v,\cR^{(3)}}(x,\cR^{(3)}) \dd x,
   \\
       \notag 
   \mu_t^{\cC^{(3)},\cR^{(2)}}(\cC^{(3)},\cdot)  &:= \int_{\bR} \mu_t^{\cC^{(3)},\cR^{(2)}}(\cC^{(3)},x) \dd x,
   \quad 
   \mu_t^{\cC^{(3)},\cR^{(2)}}( \cdot,\cR^{(2)}) := \int_{\bR} \mu_t^{\cC^{(3)},\cR^{(2)}}(x,\cR^{(2)}) \dd x,
   \\
       \notag 
   \mu_t^{\cC^{(2)},\cR^{(1)}}(\cC^{(2)},\cdot)  &:= \int_{\bR} \mu_t^{\cC^{(2)},\cR^{(1)}}(\cC^{(2)},x) \dd x,
   \quad 
   \mu_t^{\cC^{(2)},\cR^{(1)}}( \cdot,\cR^{(1)}) := \int_{\bR} \mu_t^{\cC^{(2)},\cR^{(1)}}(x,\cR^{(1)}) \dd x,
    \\
    \label{eq: 4 marginals}
   \mu_t^{\cC^{(1)},u}(\cC^{(1)},\cdot)  &:= \int_{\bR} \mu_t^{\cC^{(1)},u}(\cC^{(1)},x) \dd x,
   \quad 
   \mu_t^{\cC^{(1)},u}( \cdot,\cR^{(2)}) := \int_{\bR} \mu_t^{\cC^{(1)},u}(x,u) \dd x.
\end{align}

The weights in \eqref{eq: mlp nn function - 5-layer 1d} admits the following     

\begin{align}
    \label{eq: RC set up}
    \big\{
    v_{\gamma_4}, ~
    w^{(3)}_{ \gamma_4,\gamma_3}  ,
     w^{(2)}_{ \gamma_3,\gamma_2}  ,
     w^{(1)}_{ \gamma_2,\gamma_1}  ,
    u_{\gamma_1}
    \big\}_\gamma
    =
     \big\{
    v_{\gamma_4}, 
    \phi^{(3)}(   \cR^{(3)}_{ \gamma_4} ,    \cC^{(3)}_{ \gamma_3}    ),
    \phi^{(2)}(   \cR^{(2)}_{ \gamma_3} ,    \cC^{(2)}_{ \gamma_2}    ),
    \phi^{(1)}(   \cR^{(1)}_{ \gamma_2} ,    \cC^{(1)}_{ \gamma_1}    ),
    u_{\gamma_1}
    \big\}_\gamma
    ,
\end{align}
where $\gamma:=\{\gamma_1,\dots,\gamma_4\}$ with $\gamma_1\in \{1,\dots ,N_1\}$, $\gamma_2\in \{1,\dots ,N_2\}$, $\gamma_3\in \{1,\dots ,N_3\}$ and $\gamma_4\in \{1,\dots ,N_4\}$. 

Now, following the receipt in Section \ref{sec: mlp ffn case}, for some proper test function  $\varphi:\bR\rightarrow \bR$,  we have 
 
{ \raggedright     
\textbf{ For $v_t$: } 
}

\begin{align*}
    \bE\big[ \varphi(v_t) \big] 
    =  &
    \bE\big[ \varphi(v_0)\big] 
    +
      \int_{0}^t\Bigg( 
    \int_{ X \times Y} \eta   \Big( y - f *\mu_s(x) \Big) 
    \\
    &~ \quad  \cdot 
    \bigg\langle  \partial_x \varphi({\color{blue}v})   \cdot  \psi_4\bigg(  
    \int_{\bR^2}   
    \phi^{(3)}( {\color{blue} \cR^{(3)}  }   ,   \hat\cC^{(3)}   )
      \psi_3 \bigg( 
    \int_{\bR^2}    
    \phi^{(2)}(   \hat \cR^{(2)}    ,   \hat\cC^{(2)}   )
    \\ 
    &
    \qquad ~
    \psi_2 \Big(  \int_{\bR^2}
    \phi^{(1)}(   \hat \cR^{(1)}    ,   \hat\cC^{(1)}   )
    \psi_1( \hat u x ) 
    ~
   \mu_s^{\cC^{(1)},u}(\dd \hat \cC^{(1)}, \dd \hat u)
     ~\Big) 
     \\ 
    &
    \qquad ~~ 
     \mu_s^{\cC^{(2)},\cR^{(1)}}(\dd \hat \cC^{(2)}, \dd \hat \cR^{(1)}) \bigg)~
    \mu_s^{\cC^{(3)},\cR^{(2)}}(\dd \hat \cC^{(3)}, \dd \hat \cR^{(2)})
    \bigg)  
    , ~  
    \mu_s^{v,\cR^{(3)}}({\color{blue} \dd v  },{\color{blue} \dd \cR^{(3)}  }  )
    \bigg\rangle
    ~
    \pi(  \dd x , \dd y )
    \Bigg)\dd s,
\end{align*}

Notice that the dependency between $\Delta {v}_{t,{\color{blue} \mathbf{i} }}$ and $w_{{\color{blue} \mathbf{i} },{\gamma_3}}^{(3)}$ in \eqref{eq: delta v eg eq} is dealt by constructing the joint distribution $\mu_s^{v,\cR^{(3)}}({ \dd v  },{  \cR^{(3)}  }  )$, so that the row-wise dependency is included, the dependency between other weights are demonstrated similarly below. 

\medskip

{ \raggedright     
\textbf{ For $ \phi^{(3)}(  \cR^{(3)}_t    ,   \cC^{(3)}_t    )$:  }
}

We introduced two versions (the split dynamics and the combined  dynamics) to show the dependence across different RVs. For the split type dynamics for $\cR^{(3)}_t$, using the dot  notation defined below with the corresponding marginal measure (the notation here is a bit ambiguous, but we did not apply it for the more general cases in this article),  we have 
\begin{align*}
    \bE\Big[ &\varphi\big(      \phi^{(3)}(  \cR^{(3)}_t    ,   \cdot    ) ~ \big) \Big] 
    := 
    \bE\Big[  \varphi\big(   \int_\bR   \phi^{(3)}(  \cR^{(3)}_t    ,   x    ) 
      \mu_t^{\cC^{(3)},\cR^{(2)}}(\dd x, \cdot )
    ~ \big) \Big] 
    \\
    =&
    \bE\Big[ \varphi\big(      \phi^{(3)}(  \cR^{(3)}_0    ,   \cdot    ) ~\big)   
    \Big]
    + \bE \bigg[  
    \int_0^t   \partial_x \varphi \big(      \phi^{(3)}(  {\color{blue}  \cR^{(3)}_s }   ,   \cdot    ) ~\big)    \dd  \phi^{(3)}(  {\color{blue}  \cR^{(3)}_s }   ,   \cdot    )  
    \bigg]
    \\
     =&
    \bE\Big[ \varphi\big(      \phi^{(3)}(  \cR^{(3)}_0    ,   \cdot    ) ~\big)  
    \Big]
     +
      \int_{0}^t\Bigg( 
    \int_{ X \times Y} \eta   \Big( y - f *\mu_s(x) \Big) 
    \\
      &
    ~  \cdot 
    \bigg\langle   
    \partial_x \varphi\big(  
    \int_\bR   \phi^{(3)}(  {\color{blue} \cR^{(3)} }   ,   x    ) 
      \mu_s^{\cC^{(3)},\cR^{(2)}}(\dd x, \cdot )
    \big)
    \cdot {\color{blue} v  } \cdot \psi_4' \bigg(  
    \int_{\bR^2}   
    \phi^{(3)}( {\color{blue} \cR^{(3)} }   ,   \hat\cC^{(3)}   )
      \psi_3 \bigg( 
    \int_{\bR^2}    
    \phi^{(2)}(   \hat \cR^{(2)}    ,   \hat\cC^{(2)}   )
    \\ 
    &
    \qquad ~
    \psi_2 \Big(  \int_{\bR^2}
    \phi^{(1)}(   \hat \cR^{(1)}    ,   \hat\cC^{(1)}   )
    \psi_1( \hat u x ) 
    ~
   \mu_s^{\cC^{(1)},u}(\dd \hat \cC^{(1)}, \dd \hat u)
     ~\Big) 
    \ 
     \mu_s^{\cC^{(2)},\cR^{(1)}}(\dd \hat \cC^{(2)}, \dd \hat \cR^{(1)}) \bigg)~
    \mu_s^{\cC^{(3)},\cR^{(2)}}(\dd \hat \cC^{(3)}, \dd \hat \cR^{(2)})
    \bigg)  
    \\ 
    &
    \quad \cdot \int_{\bR}
    \psi_3 \bigg( 
    \int_{\bR^2}    
    \phi^{(2)}(   \overline \cR^{(2)}    ,   \overline\cC^{(2)}   )
    \psi_2 \Big(  \int_{\bR^2}
    \phi^{(1)}(   \overline \cR^{(1)}    ,   \overline\cC^{(1)}   )
    \psi_1( \overline u x ) 
    ~
   \mu_s^{\cC^{(1)},u}(\dd \overline \cC^{(1)}, \dd \overline u)
     ~\Big)
     \mu_s^{\cC^{(2)},\cR^{(1)}}(\dd \overline \cC^{(2)}, \dd \overline \cR^{(1)}) \bigg)     
     \\ 
    &
    \qquad ~
    \mu_s^{\cC^{(3)},\cR^{(2)}}(\cdot , \dd \overline \cR^{(2)}) , ~  
    \mu_s^{v,\cR^{(3)}}({\color{blue} \dd v  },{\color{blue} \dd \cR^{(3)} } )
    \bigg\rangle
    ~
    \pi(  \dd x , \dd y )
    \Bigg)\dd s,
\end{align*}
where the dependency between $\Delta w_{{  {i} },{j}}^{(3)}$   and  $  {v}_{t,{ {i} }}$  in \eqref{eq: delta w3 eg eq} is dealt by constructing the joint distribution $\mu^{v,\cR^{(3)}}$.   Similarly, for the split type dynamics for $\cC^{(3)}_t$, we have

\begin{align*}
    \bE\Big[ &\varphi\big(      \phi^{(3)}(  \cdot     ,   \cC^{(3)}_t    ) ~ \big)
    \Big] 
    := 
    \bE\Big[  \varphi\big(   \int_\bR   \phi^{(3)}(      x ,   \cC^{(3)}_t    ) 
      \mu_t^{v,\cR^{(3)}}(  \cdot , \dd x  )
    ~ \big)  
    \Big]
    = 
    \bE\Big[ \varphi\big(      \phi^{(3)}(  \cdot    ,   \cC^{(3)}_0    ) \big) ~  
    \Big]
    \\
      &
    +
      \int_{0}^t\Bigg( 
    \int_{ X \times Y} \eta   \Big( y - f *\mu_s(x) \Big) 
    ~ \cdot 
    \bigg\langle 
    \partial_x \varphi\big(  
    \int_\bR   \phi^{(3)}( x, {\color{blue} \cC^{(3)} }        ) 
       \mu_s^{v,\cR^{(3)}}(  \cdot , \dd x  )
    \big)
    \cdot 
    \int_{\bR^2} \hat v  \cdot  \psi_4' \bigg(  
    \int_{\bR^2}   
    \phi^{(3)}(  \hat \cR^{(3)}    ,   \hat\cC^{(3)}   )
    \\ 
    &
    \quad ~ \cdot 
    \psi_3 \bigg( 
    \int_{\bR^2}    
    \phi^{(2)}(   \hat \cR^{(2)}    ,   \hat\cC^{(2)}   )
    \psi_2 \Big(  \int_{\bR^2}
    \phi^{(1)}(   \hat \cR^{(1)}    ,   \hat\cC^{(1)}   )
    \psi_1( \hat u x ) 
    ~
   \mu_s^{\cC^{(1)},u}(\dd \hat \cC^{(1)}, \dd \hat u)
     ~\Big) 
    \ 
     \mu_s^{\cC^{(2)},\cR^{(1)}}(\dd \hat \cC^{(2)}, \dd \hat \cR^{(1)}) \bigg)~
    \\ 
    &
    \quad \qquad \quad  
    \mu_s^{\cC^{(3)},\cR^{(2)}}(\dd \hat \cC^{(3)}, \dd \hat \cR^{(2)})
    \bigg)
    ~  
    \mu_s^{v,\cR^{(3)}}({\dd  \hat v  },{\dd  \hat \cR^{(3)} } )
    \\ 
    &
    \quad \cdot   
    \psi_3 \bigg( 
    \int_{\bR^2}    
    \phi^{(2)}(   {\color{blue} \cR^{(2)} }   ,   \overline\cC^{(2)}   )
    \psi_2 \Big(  \int_{\bR^2}
    \phi^{(1)}(   \overline \cR^{(1)}    ,   \overline\cC^{(1)}   )
    \psi_1( \overline u x ) 
    ~
   \mu_s^{\cC^{(1)},u}(\dd \overline \cC^{(1)}, \dd \overline u)
     ~\Big)
     \mu_s^{\cC^{(2)},\cR^{(1)}}(\dd \overline \cC^{(2)}, \dd \overline \cR^{(1)}) \bigg)~ 
     \\ 
    &
    \qquad ~ 
     ,  
    ~  \mu_s^{\cC^{(3)},\cR^{(2)}}({\color{blue}  \dd \cC^{(3)} },{\color{blue} \dd  \cR^{(2)} } )
    \bigg\rangle
    ~
    \pi(  \dd x , \dd y )
    \Bigg)\dd s,
\end{align*} 

where the dependency between $\Delta w_{{ i },{j}}^{(3)}$   and  $  w_{j,{\cdot}}^{(2)}$ in \eqref{eq: delta w3 eg eq}  is dealt by constructing the joint distribution $\mu^{\cC^{(3)},\cR^{(2)}} $.  
For the combined dynamics  for $\cR^{(3)}_t,\cC^{(3)}_t$, we have

\begin{align*}
    \bE\Big[ &\varphi\big(      \phi^{(3)}(  \cR^{(3)}_t    ,   \cC^{(3)}_t    ) ~ \big) 
    \Big] 
    = \bE\Big[ \varphi\big(      \phi^{(3)}(  \cR^{(3)}_0    ,   \cC^{(3)}_0    ) \big) 
    \Big]
     +
      \int_{0}^t\Bigg( 
    \int_{ X \times Y} \eta   \Big( y - f *\mu_s(x) \Big) 
    \\
      &
    \quad \cdot 
    \bigg\langle  \partial_x \varphi\big(  
        \phi^{(3)}( {\color{blue} \cR^{(3)} } , {\color{blue} \cC^{(3)} }        )  
    \big)
    \cdot {\color{blue} v  }\cdot  \psi_4' \bigg(  
    \int_{\bR^2}   
    \phi^{(3)}( {\color{blue} \cR^{(3)} }   ,   \hat\cC^{(3)}   )
      \psi_3 \bigg( 
    \int_{\bR^2}    
    \phi^{(2)}(   \hat \cR^{(2)}    ,   \hat\cC^{(2)}   )
    \\ 
    &
    \qquad ~
    \psi_2 \Big(  \int_{\bR^2}
    \phi^{(1)}(   \hat \cR^{(1)}    ,   \hat\cC^{(1)}   )
    \psi_1( \hat u x ) 
    ~
   \mu_s^{\cC^{(1)},u}(\dd \hat \cC^{(1)}, \dd \hat u)
     ~\Big) 
    \ 
     \mu_s^{\cC^{(2)},\cR^{(1)}}(\dd \hat \cC^{(2)}, \dd \hat \cR^{(1)}) \bigg)~
    \mu_s^{\cC^{(3)},\cR^{(2)}}(\dd \hat \cC^{(3)}, \dd \hat \cR^{(2)})
    \bigg)  
    \\ 
    &
    \quad \cdot  
    \psi_3 \bigg( 
    \int_{\bR^2}    
    \phi^{(2)}(   {\color{blue} \cR^{(2)} }   ,   \overline\cC^{(2)}   )
    \psi_2 \Big(  \int_{\bR^2}
    \phi^{(1)}(   \overline \cR^{(1)}    ,   \overline\cC^{(1)}   )
    \psi_1( \overline u x ) 
    ~
   \mu_s^{\cC^{(1)},u}(\dd \overline \cC^{(1)}, \dd \overline u)
     ~\Big)
     \mu_s^{\cC^{(2)},\cR^{(1)}}(\dd \overline \cC^{(2)}, \dd \overline \cR^{(1)}) \bigg) 
     \\ 
    &
    \qquad ~~ 
     , ~  
    \mu_s^{v,\cR^{(3)}}({\color{blue}  \dd  v  },{\color{blue}  \dd \cR^{(3)} } )
    ~  \mu_s^{\cC^{(3)},\cR^{(2)}}({\color{blue}   \dd \cC^{(3)} },{\color{blue} \dd  \cR^{(2)} } )
    \bigg\rangle
    ~
    \pi(  \dd x , \dd y )
    \Bigg)\dd s,
\end{align*} 
where the dependency between $\Delta w_{{ i },{j}}^{(3)}$   , $  {v}_{t,{ {i} }}$  and $  w_{j,{\cdot}}^{(2)}$  in \eqref{eq: delta w3 eg eq} is dealt by   $\mu^{v,\cR^{(3)}}$ and $ \mu^{\cC^{(3)},\cR^{(2)}}$.

\medskip

{ \raggedright     
\textbf{ For $ \phi^{(2)}(  \cR^{(2)}_t    ,   \cC^{(2)}_t    )$: }
}    

Similarly as above, for the split type dynamics for $\cR^{(2)}_t$, we have 

\begin{align*}
    \bE\Big[ &\varphi\big(      \phi^{(2)}(  \cR^{(2)}_t    ,   \cdot    ) ~ \big) 
    \Big]
    := 
    \bE\Big[  \varphi\big(   \int_\bR   \phi^{(2)}(  \cR^{(2)}_t    ,   x    ) 
      \mu_t^{\cC^{(2)},\cR^{(1)}}(\dd x, \cdot )
    \Big]
    = \bE\Big[ \varphi\big(      \phi^{(2)}(  \cR^{(2)}_0    ,   \cdot    )
    \Big]
    \\
      &
    +
      \int_{0}^t\Bigg( 
    \int_{ X \times Y} \eta   \Big( y - f *\mu_s(x) \Big) 
    ~  \cdot 
    \bigg\langle  \partial_x \varphi\big(  
    \int_\bR   \phi^{(2)}(  {\color{blue} \cR^{(2)} }   ,   x    ) 
      \mu_s^{\cC^{(2)},\cR^{(1)}}(\dd x, \cdot )
    \big) 
    \\ 
    &
    \qquad
    \cdot 
    \int_{\bR^2}
    {  \hat v  } \cdot \psi_4' \bigg(  
    \int_{\bR^2}   
    \phi^{(3)}( {\hat \cR^{(3)} }   ,   \hat\cC^{(3)}   )
      \psi_3 \bigg( 
    \int_{\bR^2}    
    \phi^{(2)}(   \hat \cR^{(2)}    ,   \hat\cC^{(2)}   )
    \\ 
    &
    \qquad ~
    \psi_2 \Big(  \int_{\bR^2}
    \phi^{(1)}(   \hat \cR^{(1)}    ,   \hat\cC^{(1)}   )
    \psi_1( \hat u x ) 
    ~
   \mu_s^{\cC^{(1)},u}(\dd \hat \cC^{(1)}, \dd \hat u)
     ~\Big) 
     \mu_s^{\cC^{(2)},\cR^{(1)}}(\dd \hat \cC^{(2)}, \dd \hat \cR^{(1)}) \bigg)
    \
    \\ 
    &
    \qquad \qquad \qquad ~
    ~
    \mu_s^{\cC^{(3)},\cR^{(2)}}(\dd \hat \cC^{(3)}, \dd \hat \cR^{(2)})    
    \bigg)
    \cdot 
    \phi^{(3)}( {\hat \cR^{(3)} }   ,    {\color{blue} \cC^{(3)} }   )
    ~ 
     \mu_s^{v,\cR^{(3)}}(\dd {\hat v  },{\dd \hat  \cR^{(3)} } )
    \\ 
    &
    \quad \cdot  
    \psi_3' \bigg( 
    \int_{\bR^2}    
    \phi^{(2)}(   {\color{blue} \cR^{(2)} }    ,   \overline\cC^{(2)}   )
    \psi_2 \Big(  \int_{\bR^2}
    \phi^{(1)}(   \overline \cR^{(1)}    ,   \overline\cC^{(1)}   )
    \psi_1( \overline u x ) 
    ~
   \mu_s^{\cC^{(1)},u}(\dd \overline \cC^{(1)}, \dd \overline u)
     ~\Big)
     \mu_s^{\cC^{(2)},\cR^{(1)}}(\dd \overline \cC^{(2)}, \dd \overline \cR^{(1)}) \bigg)~      
     \\ 
    &
    \qquad ~
    \cdot   \int_{\bR}~
    \psi_2
    \Big(  \int_{\bR^2}
    \phi^{(1)}(   \tilde \cR^{(1)}   ,   \tilde\cC^{(1)}   )
    \psi_1( \tilde u x ) 
    ~
   \mu_s^{\cC^{(1)},u}(\dd \tilde \cC^{(1)}, \dd \tilde u)
     ~\Big)
     \mu_s^{\cC^{(2)},\cR^{(1)}}(  \cdot,  \dd \tilde \cR^{(1)}  )
     \\ 
    &
    \qquad  \qquad \quad 
    , ~  
    \mu_s^{\cC^{(3)},\cR^{(2)}}( {\color{blue}  \dd \cC^{(3)} },{\color{blue}  \dd \cR^{(2)} }  )
    \bigg\rangle
    ~
    \pi(  \dd x , \dd y )
    \Bigg)\dd s,
\end{align*}
where the dependency between $\Delta w_{{ i },{j}}^{(2)}$   and  $  w_{ {\cdot},i}^{(3)}$ in in \eqref{eq: delta w2 eg eq}  is dealt by   the joint distribution $\mu^{\cC^{(3)},\cR^{(2)}} $. For the split type dynamics for $\cC^{(2)}_t$, we have 

\begin{align*}
    \bE\Big[ &\varphi\big(      \phi^{(2)}(  \cdot , \cC^{(2)}_t        ) ~ \big) 
    \Big] 
    := 
    \bE\Big[  \varphi\big(   \int_\bR   \phi^{(2)}(     x   , \cC^{(2)}_t   ) 
      \mu_t^{\cC^{(3)},\cR^{(2)}}( \cdot ,\dd x  )
    ~ \big) 
    \Big]
    = \bE\Big[ \varphi\big(      \phi^{(2)}( \cdot,  \cC^{(2)}_0            )\big)  ~  
    \Big]
    \\
      &
    +
      \int_{0}^t\Bigg( 
    \int_{ X \times Y} \eta   \Big( y - f *\mu_s(x) \Big) 
    ~  \cdot 
    \bigg\langle \partial_x \varphi\big(  
    \int_\bR   \phi^{(2)}(  x,{\color{blue} \cC^{(2)} }       ) 
      \mu_s^{\cC^{(3)},\cR^{(2)}}(\dd x, \cdot )
    ~ \big)
    \\ 
    &
    \qquad
    \cdot \int_{\bR^2}~
    \int_{\bR^2}
    {  \hat v  } \cdot \psi_4' \bigg(  
    \int_{\bR^2}   
    \phi^{(3)}( {\hat \cR^{(3)} }   ,   \hat\cC^{(3)}   )
      \psi_3 \bigg( 
    \int_{\bR^2}    
    \phi^{(2)}(   \hat \cR^{(2)}    ,   \hat\cC^{(2)}   )
    \\ 
    &
    \qquad ~
    \psi_2 \Big(  \int_{\bR^2}
    \phi^{(1)}(   \hat \cR^{(1)}    ,   \hat\cC^{(1)}   )
    \psi_1( \hat u x ) 
    ~
   \mu_s^{\cC^{(1)},u}(\dd \hat \cC^{(1)}, \dd \hat u)
     ~\Big) 
     \mu_s^{\cC^{(2)},\cR^{(1)}}(\dd \hat \cC^{(2)}, \dd \hat \cR^{(1)}) \bigg)
    \
    \\ 
    &
    \qquad \qquad \qquad ~
    ~
    \mu_s^{\cC^{(3)},\cR^{(2)}}(\dd \hat \cC^{(3)}, \dd \hat \cR^{(2)})    
    \bigg)
    \cdot 
    \phi^{(3)}( {\hat \cR^{(3)} }   ,    {\color{purple}  {\tilde \cC^{(3)} }  } )
    ~ 
     \mu_s^{v,\cR^{(3)}}(\dd {\hat v  },{\dd \hat  \cR^{(3)} } )
    \\ 
    &
    \quad \cdot  
    \psi_3' \bigg( 
    \int_{\bR^2}    
    \phi^{(2)}(   {\color{purple}  {\tilde \cR^{(2)} }}    ,   \overline\cC^{(2)}   )
    \psi_2 \Big(  \int_{\bR^2}
    \phi^{(1)}(   \overline \cR^{(1)}    ,   \overline\cC^{(1)}   )
    \psi_1( \overline u x ) 
    ~
   \mu_s^{\cC^{(1)},u}(\dd \overline \cC^{(1)}, \dd \overline u)
     ~\Big)
     \mu_s^{\cC^{(2)},\cR^{(1)}}(\dd \overline \cC^{(2)}, \dd \overline \cR^{(1)}) \bigg)~      
     \\ 
    &
    \qquad ~  \mu_s^{\cC^{(3)},\cR^{(2)}}({\color{purple}  \dd \tilde \cC^{(3)}}, {\color{purple} \dd \tilde \cR^{(2)} }) 
    \cdot   
    \psi_2
    \Big(  \int_{\bR^2}
    \phi^{(1)}(   {\color{blue} \cR^{(1)} }   ,   \tilde\cC^{(1)}   )
    \psi_1( \tilde u x ) 
    ~
   \mu_s^{\cC^{(1)},u}(\dd  \tilde \cC^{(1)}, \dd \tilde u)
     ~\Big) 
     \\ 
    &
    \qquad  \qquad \quad 
    , ~  
    \mu_s^{\cC^{(2)},\cR^{(1)}}( {\color{blue}  \dd  \cC^{(2)} },{\color{blue}  \dd  \cR^{(1)} }  )
    \bigg\rangle
    ~
    \pi(  \dd x , \dd y )
    \Bigg)\dd s,
\end{align*}

where the dependency between $\Delta w_{{ i },{j}}^{(2)}$   and  $  w_{ j,{\cdot}}^{(1)}$ in in \eqref{eq: delta w2 eg eq}  is dealt by  the joint distribution $\mu^{\cC^{(2)},\cR^{(1)}} $.
For the combined dynamics  for $\cR^{(2)}_t,\cC^{(2)}_t$, we have

\begin{align*}
    \bE\Big[ &\varphi\big(      \phi^{(2)}(  \cR^{(2)}_t  , \cC^{(2)}_t        ) ~ \big) 
    \Big] = \bE\Big[ \varphi\big(      \phi^{(2)}( \cR^{(2)}_0 ,  \cC^{(2)}_0            ) \big) 
    \Big]
     +
      \int_{0}^t\Bigg( 
    \int_{ X \times Y} \eta   \Big( y - f *\mu_s(x) \Big) 
    \\
      &
    \quad 
    ~  \cdot 
    \bigg\langle  \partial_x \varphi\big(  
        \phi^{(2)}( {\color{blue} \cR^{(2)} } , {\color{blue} \cC^{(2)} }        )  ~ 
    \big)
    \cdot \int_{\bR^2}~
    {  \hat v  } \cdot \psi_4' \bigg(  
    \int_{\bR^2}   
    \phi^{(3)}( {\hat \cR^{(3)} }   ,   \hat\cC^{(3)}   )
      \psi_3 \bigg( 
    \int_{\bR^2}    
    \phi^{(2)}(   \hat \cR^{(2)}    ,   \hat\cC^{(2)}   )
    \\ 
    &
    \qquad ~
    \psi_2 \Big(  \int_{\bR^2}
    \phi^{(1)}(   \hat \cR^{(1)}    ,   \hat\cC^{(1)}   )
    \psi_1( \hat u x ) 
    ~
   \mu_s^{\cC^{(1)},u}(\dd \hat \cC^{(1)}, \dd \hat u)
     ~\Big) 
     \mu_s^{\cC^{(2)},\cR^{(1)}}(\dd \hat \cC^{(2)}, \dd \hat \cR^{(1)}) \bigg)
    \
    \\ 
    &
    \qquad \qquad \qquad ~
    ~
    \mu_s^{\cC^{(3)},\cR^{(2)}}(\dd \hat \cC^{(3)}, \dd \hat \cR^{(2)})    
    \bigg)
    \cdot 
    \phi^{(3)}( {\hat \cR^{(3)} }   ,    {\color{blue} \cC^{(3)} }   )
    ~ 
     \mu_s^{v,\cR^{(3)}}(\dd {\hat v  },{\dd \hat  \cR^{(3)} } )
    \\ 
    &
    \quad \cdot  
    \psi_3' \bigg( 
    \int_{\bR^2}    
    \phi^{(2)}(   {\color{blue} \cR^{(2)} }    ,   \overline\cC^{(2)}   )
    \psi_2 \Big(  \int_{\bR^2}
    \phi^{(1)}(   \overline \cR^{(1)}    ,   \overline\cC^{(1)}   )
    \psi_1( \overline u x ) 
    ~
   \mu_s^{\cC^{(1)},u}(\dd \overline \cC^{(1)}, \dd \overline u)
     ~\Big)
     \mu_s^{\cC^{(2)},\cR^{(1)}}(\dd \overline \cC^{(2)}, \dd \overline \cR^{(1)}) \bigg)~      
     \\ 
    &
    \qquad ~  
    \cdot   
    \psi_2
    \Big(  \int_{\bR^2}
    \phi^{(1)}(   {\color{blue} \cR^{(1)} }   ,   \tilde\cC^{(1)}   )
    \psi_1( \tilde u x ) 
    ~
   \mu_s^{\cC^{(1)},u}(\dd \tilde \cC^{(1)}, \dd \tilde u)
     ~\Big) 
     \\ 
    &
    \qquad  \qquad \quad 
    , ~   \mu_s^{\cC^{(3)},\cR^{(2)}}( {\color{blue}  \dd \cC^{(3)} },{\color{blue}  \dd \cR^{(2)} }  )
    \cdot 
    \mu_s^{\cC^{(2)},\cR^{(1)}}( {\color{blue} \dd  \cC^{(2)} },{\color{blue}  \dd \cR^{(1)} }  )
    \bigg\rangle
    ~
    \pi(  \dd x , \dd y )
    \Bigg)\dd s,
\end{align*}

where the dependency between $\Delta w_{{ i },{j}}^{(2)}$   , $  w_{ {\cdot},i}^{(3)}$ and  $  w_{ j,{\cdot}}^{(1)}$  in  \eqref{eq: delta w2 eg eq} is dealt by $\mu^{\cC^{(3)},\cR^{(2)}} $ and $ \mu^{\cC^{(2)},\cR^{(1)}} $.

\medskip

{ \raggedright     
\textbf{ For $ \phi^{(1)}(  \cR^{(1)}_t    ,   \cC^{(1)}_t    )$: }
}

For the split type dynamics for $\cR^{(1)}_t$, we have

\begin{align*}
    \bE\Big[ &\varphi\big(      \phi^{(1)}(  \cR^{(1)}_t  , \cdot         ) ~ 
    \Big] 
    := 
    \bE\Big[  \varphi\big(   \int_\bR   \phi^{(1)}(  \cR^{(1)}_t    ,   x    ) 
      \mu_t^{\cC^{(1)},u}(\dd x, \cdot )
    ~ \big)  
    \Big]
    = \bE\Big[ \varphi\big(      \phi^{(1)}( \cR^{(1)}_0 ,  \cC^{(1)}_0            ) \big) ~ 
    \Big]
    \\
      &
    +
      \int_{0}^t\Bigg( 
    \int_{ X \times Y} \eta   \Big( y - f *\mu_s(x) \Big) 
    ~  \cdot 
    \bigg\langle  \partial_x \varphi\big(  
    \int_\bR   \phi^{(1)}(  {\color{blue} \cR^{(1)} }   ,   x    ) 
      \mu_s^{\cC^{(1)},u}(\dd x, \cdot )
    \big)
    \\ 
    &
    \qquad
    \cdot \int_{\bR^2}~
    {  \hat v  } \cdot \psi_4' \bigg(  
    \int_{\bR^2}   
    \phi^{(3)}( {\hat \cR^{(3)} }   ,   \hat\cC^{(3)}   )
      \psi_3 \bigg( 
    \int_{\bR^2}    
    \phi^{(2)}(   \hat \cR^{(2)}    ,   \hat\cC^{(2)}   )
    \cdot
     \psi_2 \Big(  \int_{\bR^2}
    \phi^{(1)}(   \hat \cR^{(1)}    ,   \hat\cC^{(1)}   )
    \psi_1( \hat u x ) 
    \\ 
    &
    \qquad ~
    ~ \cdot 
   \mu_s^{\cC^{(1)},u}(\dd \hat \cC^{(1)}, \dd \hat u)
     ~\Big) 
     \mu_s^{\cC^{(2)},\cR^{(1)}}(\dd \hat \cC^{(2)}, \dd \hat \cR^{(1)}) \bigg)
    \
    \mu_s^{\cC^{(3)},\cR^{(2)}}(\dd \hat \cC^{(3)}, \dd \hat \cR^{(2)})    
    \bigg)
    \\ 
    &
    \quad \cdot  
    \int_{\bR^2}   
    \phi^{(3)}( {\hat \cR^{(3)} }   ,   \overline \cC^{(3)}   )
     \psi_3 \bigg( 
    \int_{\bR^2}    
    \phi^{(2)}(   \overline \cR^{(2)}    ,   \overline\cC^{(2)}   )
    \psi_2 \Big(  \int_{\bR^2}
    \phi^{(1)}(   \overline\cR^{(1)}    ,   \overline\cC^{(1)}   )
    \psi_1( \overline u x )  
    ~
   \mu_s^{\cC^{(1)},u}(\dd \overline\cC^{(1)}, \dd \overline u)
     ~\Big) 
     \\ 
    &
    \qquad \qquad \cdot 
     \mu_s^{\cC^{(2)},\cR^{(1)}}(\dd \overline \cC^{(2)}, \dd \overline \cR^{(1)}) \bigg)
     \phi^{(2)}( {\overline \cR^{(2)} }   ,   { \color{blue}  \cC^{(2)} }     ) 
    \cdot 
    ~ 
     \mu_s^{\cC^{(3)},\cR^{(2)}}(\dd \overline \cC^{(3)}, \dd \overline \cR^{(2)})
    \cdot ~
     \mu_s^{v,\cR^{(3)}}(\dd {\hat v  },{\dd \hat  \cR^{(3)} } )     
     \\ 
    &
    \qquad ~  
    \cdot   
    \psi_2' 
    \Big(  \int_{\bR^2}
    \phi^{(1)}(   {\color{blue} \cR^{(1)} }   ,   \tilde\cC^{(1)}   )
    \psi_1( \tilde u x ) 
    ~
   \mu_s^{\cC^{(1)},u}(\dd \tilde \cC^{(1)}, \dd \tilde u)
     ~\Big) 
     \cdot 
     \int_\bR 
     \psi_1( \dddot u  x ) 
     \cdot \mu_s^{\cC^{(1)},u}(\cdot ,{\dddot u }  )
     ~
     \\ 
    &
    \qquad  \qquad \quad 
    , ~  
        \mu_s^{\cC^{(2)},\cR^{(1)}}( {\color{blue}  \dd \cC^{(2)} },{\color{blue}  \dd \cR^{(1)} }  )   
    \bigg\rangle
    ~
    \pi(  \dd x , \dd y )
    \Bigg)\dd s,
\end{align*}
where the dependency between $\Delta w_{{ i },{j}}^{(1)}$   and  $  w_{ {\cdot},i}^{(2)}$ in  \eqref{eq: delta w1 eg eq}   is dealt by the joint distribution $\mu^{\cC^{(2)},\cR^{(1)}} $. 
For the split type dynamics for $\cC^{(1)}_t$, we have 
\begin{align*}
    \bE\Big[ &\varphi\big(      \phi^{(1)}( \cdot  , \cC^{(1)}_t        ) ~ \big)
    \Big] 
     := 
    \bE\Big[  \varphi\big(   \int_\bR   \phi^{(1)}(     x   , \cC^{(1)}_t   ) 
      \mu_t^{\cC^{(2)},\cR^{(1)}}( \cdot ,\dd x  )
    ~ \big) 
    \Big]
    = \bE\Big[ \varphi\big(      \phi^{(1)}( \cdot ,  \cC^{(1)}_0            ) \big) ~ 
    \Big]
    \\
      &
    +
      \int_{0}^t\Bigg( 
    \int_{ X \times Y} \eta   \Big( y - f *\mu_s(x) \Big) 
    ~  \cdot 
    \bigg\langle  
    \partial_x \varphi\big(  
    \int_\bR   \phi^{(1)}( x, {\color{blue} \cC^{(1)} }        ) 
       \mu_s^{\cC^{(2)},\cR^{(1)}}(  \cdot , \dd x  )
    \big)
    \\ 
    &
    \qquad
    \cdot \int_{\bR^2}~     \int_{\bR^2}~\bigg[  
    {  \hat v  } \cdot \psi_4' \bigg(  
    \int_{\bR^2}   
    \phi^{(3)}( {\hat \cR^{(3)} }   ,   \hat\cC^{(3)}   )
      \psi_3 \bigg( 
    \int_{\bR^2}    
    \phi^{(2)}(   \hat \cR^{(2)}    ,   \hat\cC^{(2)}   )
    ~
    \psi_2 \Big(  \int_{\bR^2}
    \phi^{(1)}(   \hat \cR^{(1)}    ,   \hat\cC^{(1)}   )
    \psi_1( \hat u x )
    \\ 
    &
    \qquad  \quad 
    ~ \cdot 
   \mu_s^{\cC^{(1)},u}(\dd \hat \cC^{(1)}, \dd \hat u)
     ~\Big) 
     \mu_s^{\cC^{(2)},\cR^{(1)}}(\dd \hat \cC^{(2)}, \dd \hat \cR^{(1)}) \bigg)
    \cdot 
    \mu_s^{\cC^{(3)},\cR^{(2)}}(\dd \hat \cC^{(3)}, \dd \hat \cR^{(2)})    
    \bigg)
    \\ 
    &
    \quad \cdot  
    \int_{\bR^2}   
    \phi^{(3)}( {\hat \cR^{(3)} }   ,   \overline \cC^{(3)}   )
     \psi_3 \bigg( 
    \int_{\bR^2}    
    \phi^{(2)}(   \overline \cR^{(2)}    ,   \overline\cC^{(2)}   )
    \psi_2 \Big(  \int_{\bR^2}
    \phi^{(1)}(   \overline\cR^{(1)}    ,   \overline\cC^{(1)}   )
    \psi_1( \overline u x )  
    ~
   \mu_s^{\cC^{(1)},u}(\dd \overline\cC^{(1)}, \dd \overline u)
     ~\Big) 
     \\ 
    &
    \qquad \qquad  \cdot 
     \mu_s^{\cC^{(2)},\cR^{(1)}}(\dd \overline \cC^{(2)}, \dd \overline \cR^{(1)}) \bigg)
     \phi^{(2)}( {\overline \cR^{(2)} }   ,   {\color{purple} \dddot  \cC^{(2)} }     ) 
    \cdot 
    ~ 
     \mu_s^{\cC^{(3)},\cR^{(2)}}(\dd \overline \cC^{(3)}, \dd \overline \cR^{(2)})
      \bigg] ~ \cdot 
     \mu_s^{v,\cR^{(3)}}(\dd {\hat v  },{\dd \hat  \cR^{(3)} } )   
     \\ 
    &
    \qquad ~  
    \cdot   
    \psi_2' 
    \Big(  \int_{\bR^2}
    \phi^{(1)}(   {\color{purple} \dddot  \cR^{(1)} }   ,   \tilde\cC^{(1)}   )
    \psi_1( \tilde u x ) 
    ~
   \mu_s^{\cC^{(1)},u}(\dd \tilde \cC^{(1)}, \dd \tilde u)
     ~\Big) 
     \cdot
     \mu_s^{\cC^{(2)},\cR^{(1)}}( { \color{purple} \dd \dddot \cC^{(2)} },{  \color{purple} \dd \dddot\cR^{(1)} }  )
     \\ 
    &
    \qquad  \qquad \quad 
    \cdot \psi_1( {\color{blue} u }  x ) 
    , ~  
          \mu_s^{\cC^{(1)},u}( {\color{blue} \dd\cC^{(1)} },{\color{blue}\dd u }  )   
    \bigg\rangle
    ~
    \pi(  \dd x , \dd y )
    \Bigg)\dd s,
\end{align*}
where the dependency between $\Delta w_{{ i },{j}}^{(1)}$  and  $  u_{ j}$ in  \eqref{eq: delta w1 eg eq}  is dealt by  the joint distribution $\mu^{\cC^{(1)},u} $.
For the combined dynamics  for $\cR^{(1)}_t,\cC^{(1)}_t$, we have

\begin{align*}
    \bE\Big[ &\varphi\big(      \phi^{(1)}(  \cR^{(1)}_t  , \cC^{(1)}_t        ) ~ \big) 
    \Big] 
    = 
    \bE\Big[ \varphi\big(      \phi^{(1)}( \cR^{(1)}_0 ,  \cC^{(1)}_0            ) \big) ~ 
    \Big]
    \\
      &
    +
      \int_{0}^t\Bigg( 
    \int_{ X \times Y} \eta   \Big( y - f *\mu_s(x) \Big) 
    ~  \cdot 
    \bigg\langle  
    \partial_x \varphi\big(  
        \phi^{(1)}( {\color{blue} \cR^{(1)} } , {\color{blue} \cC^{(1)} }        )  
    \big)
     \\ 
    &
    \qquad ~
    \cdot \int_{\bR^2}~
    { \color{purple}  \hat v  } \cdot \psi_4' \bigg(  
    \int_{\bR^2}   
    \phi^{(3)}( {\color{purple} \hat \cR^{(3)} }   ,   \hat\cC^{(3)}   )
      \psi_3 \bigg( 
    \int_{\bR^2}    
    \phi^{(2)}(   \hat \cR^{(2)}    ,   \hat\cC^{(2)}   )
     \psi_2 \Big(  \int_{\bR^2}
    \phi^{(1)}(   \hat \cR^{(1)}    ,   \hat\cC^{(1)}   )
    \psi_1( \hat u x ) 
    ~
    \\ 
    &
    \qquad ~
   \cdot 
   \mu_s^{\cC^{(1)},u}(\dd \hat \cC^{(1)}, \dd \hat u)
     ~\Big)  \ 
     \mu_s^{\cC^{(2)},\cR^{(1)}}(\dd \hat \cC^{(2)}, \dd \hat \cR^{(1)}) \bigg)
    \cdot 
    \mu_s^{\cC^{(3)},\cR^{(2)}}(\dd \hat \cC^{(3)}, \dd \hat \cR^{(2)})    
    \bigg)
    \\ 
    &
    \quad \cdot  
    \int_{\bR^2}   
    \phi^{(3)}( {\color{purple} \hat \cR^{(3)} }   ,   \overline \cC^{(3)}   )
     \psi_3 \bigg( 
    \int_{\bR^2}    
    \phi^{(2)}(   \overline \cR^{(2)}    ,   \overline\cC^{(2)}   )
    \psi_2 \Big(  \int_{\bR^2}
    \phi^{(1)}(   \overline\cR^{(1)}    ,   \overline\cC^{(1)}   )
    \psi_1( \overline u x )  
    ~
   \mu_s^{\cC^{(1)},u}(\dd \overline\cC^{(1)}, \dd \overline u)
     ~\Big) 
     \\ 
    &
    \qquad \qquad   \cdot 
     \mu_s^{\cC^{(2)},\cR^{(1)}}(\dd \overline \cC^{(2)}, \dd \overline \cR^{(1)}) \bigg)
     \phi^{(2)}( {\overline \cR^{(2)} }   ,   { \color{blue}  \cC^{(2)} }     ) 
     \cdot 
    ~ 
     \mu_s^{\cC^{(3)},\cR^{(2)}}(\dd \overline \cC^{(3)}, \dd \overline \cR^{(2)}) 
    \cdot 
     \mu_s^{v,\cR^{(3)}}( {\color{purple} \dd\hat v  },{\color{purple} \dd \hat  \cR^{(3)} } )     
     \\ 
    &
    \qquad   
    \cdot   
    \psi_2' 
    \Big(  \int_{\bR^2}
    \phi^{(1)}(   {\color{blue} \cR^{(1)} }   ,   \tilde\cC^{(1)}   )
    \psi_1( \tilde u x ) 
    ~
   \mu_s^{\cC^{(1)},u}(\dd \tilde \cC^{(1)}, \dd \tilde u)
     ~\Big) 
     \cdot \psi_1( {\color{blue} u }  x ) 
     ~
     \\ 
    &
    \qquad  \qquad \quad 
    , ~  
        \mu_s^{\cC^{(2)},\cR^{(1)}}( {\color{blue}  \dd \cC^{(2)} },{\color{blue}  \dd \cR^{(1)} }  )
    \cdot \mu_s^{\cC^{(1)},u}( {\color{blue}  \dd \cC^{(1)} },{\color{blue}  \dd u }  )   
    \bigg\rangle
    ~
    \pi(  \dd x , \dd y )
    \Bigg)\dd s,
\end{align*}
where the dependency between $\Delta w_{{ i },{j}}^{(1)}$   , $  w_{ {\cdot},i}^{(2)}$ and  $ u_j$ in \eqref{eq: delta w1 eg eq} is dealt by    $\mu^{\cC^{(2)},\cR^{(1)}}$ and $ \mu^{\cC^{(1)},u} $.

{ \raggedright     
\textbf{ For $u_t$: } 
}

\begin{align*}
    \bE\Big[ &\varphi\big(     u_t   \big) 
    \Big] 
    = \bE\Big[ \varphi\big(       u_0            \big)
    \Big]
    \\
      &
    +
      \int_{0}^t\Bigg( 
    \int_{ X \times Y} \eta   \Big( y - f *\mu_s(x) \Big) 
    ~  \cdot 
    \bigg\langle   
    \partial_x\varphi\big(     {\color{blue} u}          \big) 
    \cdot \int_{\bR^2}~
    {  \hat v  } \cdot \psi_4' \bigg(  
    \int_{\bR^2}   
    \phi^{(3)}( {\hat \cR^{(3)} }   ,   \hat\cC^{(3)}   )
    \\ 
    &
    \qquad 
    \psi_3 \bigg( 
    \int_{\bR^2}    
    \phi^{(2)}(   \hat \cR^{(2)}    ,   \hat\cC^{(2)}   )
    \psi_2 \Big(  \int_{\bR^2}
    \phi^{(1)}(   \hat \cR^{(1)}    ,   \hat\cC^{(1)}   )
    \psi_1( \hat u x ) 
    ~
   \mu_s^{\cC^{(1)},u}(\dd \hat \cC^{(1)}, \dd \hat u)
     ~\Big) 
     \mu_s^{\cC^{(2)},\cR^{(1)}}(\dd \hat \cC^{(2)}, \dd \hat \cR^{(1)}) \bigg)
    \
    \\ 
    &
    \qquad \qquad \qquad ~
    ~
    \mu_s^{\cC^{(3)},\cR^{(2)}}(\dd \hat \cC^{(3)}, \dd \hat \cR^{(2)})    
    \bigg)
    \\ 
    &
    \quad \cdot  
    \int_{\bR^2}   
    \phi^{(3)}( {\hat \cR^{(3)} }   ,   \overline \cC^{(3)}   )
     \psi_3'  \bigg( 
    \int_{\bR^2}    
    \phi^{(2)}(   \overline \cR^{(2)}    ,   \overline\cC^{(2)}   )
    \psi_2 \Big(  \int_{\bR^2}
    \phi^{(1)}(   \overline\cR^{(1)}    ,   \overline\cC^{(1)}   )
    \psi_1( \overline u x )  
    ~
   \mu_s^{\cC^{(1)},u}(\dd \overline\cC^{(1)}, \dd \overline u)
     ~\Big) 
     \\ 
    &
    \qquad \qquad \qquad
    \qquad \qquad \qquad
     \mu_s^{\cC^{(2)},\cR^{(1)}}(\dd \overline \cC^{(2)}, \dd \overline \cR^{(1)}) \bigg)
     \\ 
    &
    \qquad \ \ 
    \int_{\bR^2}
     \phi^{(2)}( {\overline \cR^{(2)} }   ,   { \tilde  \cC^{(2)} }     ) 
     \cdot   
    \psi_2' 
    \Big(  \int_{\bR^2}
    \phi^{(1)}(   {\tilde \cR^{(1)} }   ,   \tilde\cC^{(1)}   )
    \psi_1( \tilde u x ) 
    ~
   \mu_s^{\cC^{(1)},u}(\dd \tilde \cC^{(1)}, \dd \tilde u)
     ~\Big)
     \cdot \phi^{(1)}(   {\tilde \cR^{(1)} }   ,   {\color{blue} \cC^{(1)} }   )
     \\
     & \qquad \qquad \qquad
     \cdot \psi_1'( {\color{blue} u }  x ) x
    \cdot 
    ~
    \mu_s^{\cC^{(2)},\cR^{(1)}}( {\tilde \cC^{(2)} },{\tilde \cR^{(1)} }  )
    \cdot 
    ~ 
     \mu_s^{\cC^{(3)},\cR^{(2)}}(\dd \overline \cC^{(3)}, \dd \overline \cR^{(2)})  
     \\ 
    &
    \qquad  \qquad \quad 
    \cdot ~
     \mu_s^{v,\cR^{(3)}}(\dd {\hat v  },{\dd \hat  \cR^{(3)} } )  
    ,  
    ~ 
    \mu_s^{\cC^{(1)},u}( {\color{blue} \dd \cC^{(1)} },{\color{blue} \dd u }  )   
    \bigg\rangle
    ~
    \pi(  \dd x , \dd y )
    \Bigg)\dd s,
\end{align*}
where the dependency between $  u_{ j}$    and   $\Delta w_{ \cdot ,{j}}^{(1)}$ in \eqref{eq: delta  eg eq}  is dealt by the joint distribution $\mu^{\cC^{(1)},u} $.

So that we can see under the set up of $\{u,\cR^{(1)},\cC^{(1)},\cR^{(2)},\cC^{(2)},\cR^{(3)},\cC^{(3)},v   \}_t \sim \mu_t\in\cP(\bR^8) $ with $\mu_t = \mu_t^{v,\cR^{(3)}}  \times 
   \mu_t^{\cC^{(3)},\cR^{(2)}}  \times \mu_t^{\cC^{(2)},\cR^{(1)}}\times   \mu_t^{\cC^{(1)},u}   $ defined in \eqref{eq: 4 marginals}, the  dynamics for measures are now closed, the simplified 5-layer MFNN in \eqref{eq: mlp nn function - 5-layer 1d} admits the mean filed point of view.  The dependency across the weights in different layers are  in a way of rows  follow by columns, and is dealt by the RC ansatz through $\{\phi,\cC,\cR\}^{(\ell)}$ and the corresponding marginals.

   Notice that one can also write down the corresponding Fokker-Planck equations following the steps in \cite{mf-2022-path2} which we do not present here, discussions on the  Fokker-Planck equation are also introduced in \cite{2018-mf-1,2019-mf-lln,lu2020mean,nguyen2023rigorous}.  From the mean-field point of view, the 4 components (each in $\cP(\bR^2)$)  of the measure $\mu_t \in \cP(\bR^8)$  evolve as the training process with the dataset $X\times Y $, and there is more information we can use from the measure evolution which we leave for future work.

\subsubsection{Discussion on other  measure structure}
\label{sec: discuss on other measure structure}

The RC ansatz is not the unique measure structure for the mean-field point of view to MFNN, in \cite[Section 4]{mf-2019-path1},  by freezing the first and the last layers, they show the following 
\begin{align*}
    \mu_t( &\dd v, \dd w^{(3)}, \dd  w^{(2)}, \dd  w^{(1)} , \dd u) 
    \\
    & =
    \mu_t( \dd v,  \cdot ,\cdot , \cdot ,\cdot) 
    \mu_t( v,  \dd w^{(3)} ,\cdot , \cdot ,\cdot) 
    \mu_t( \cdot,  \cdot, \dd w^{(2)} , \cdot ,\cdot) 
    \mu_t( \cdot,  \cdot, \cdot , \dd w^{(1)}   ,u) 
    \mu_t(    \cdot ,\cdot , \cdot ,\cdot, \dd u), 
\end{align*}
where the weights are path-wise connected via the $\gamma$ notation in Definition \ref{def: gamma} such that the second layer $w^{(1)}$ depends on $u$, and the second last layer $w^{(3)}$ depends on $v$, other middle layers are independent. The path-wise connected idea and the "ideal particles" inspired us to develop the RC ansatz, and by choosing a different scalar for the training process ( scalar on $\eta$  for $\{v,w^{(3)},w^{(2)},w^{(1)},u\}$ are $\{N_4,N_4N_3, N_3N_2,N_2N_1,N_1\}$ in \eqref{eq: delta  eg eq}),  the issue discussed in \cite[Remark 3.4]{mf-2019-path1} is amended.

In the great work \cite{nguyen2023rigorous} with rigorous mathematical proof, they observe the layer dependency across different layers and develop the "Neuronal embedding" idea to deal with it, which is of the same nature as the function sets $\{ \phi\}^{\ell}$ in \eqref{eq: emb function}. For the simplified  5-layer MFNN example  in Section \ref{section: anstazr for 5 layers}, their settings would have 4 RVs, while we have 8 RVs, such that for some proper functions $\{g^{(l)}\}_{l=1}^5$,  they have

\begin{align}
    \notag 
    \big\{
    v_{\gamma_4}&, ~
    w^{(3)}_{ \gamma_4,\gamma_3}  ,
     w^{(2)}_{ \gamma_3,\gamma_2}  ,
     w^{(1)}_{ \gamma_2,\gamma_1}  ,
    u_{\gamma_1}
    \big\}_\gamma
    \\
        \label{eq: rigo set up}
    &=
     \big\{
    g^{(5)}(\theta^{(4)}_{\gamma_4}), 
    g^{(4)}(   \theta_{ \gamma_4}^{(4)} ,    \theta_{ \gamma_3}^{(3)}   ),
    g^{(3)}(   \theta_{ \gamma_3}^{(3)}  ,    \theta_{ \gamma_2}^{(2)}     ),
    g^{(2)}(   \theta_{ \gamma_2}^{(2)}  ,    \theta_{ \gamma_1}^{(1)}     ),
    g^{(1)}(\theta_{\gamma_1}^{(1)} )
    \big\}_\gamma
    ,
\end{align}
where $\gamma:=\{\gamma_1,\dots,\gamma_4\}$ with $\gamma_1\in \{1,\dots ,N_1\}$, $\gamma_2\in \{1,\dots ,N_2\}$, $\gamma_3\in \{1,\dots ,N_3\}$ and $\gamma_4\in \{1,\dots ,N_4\}$ with 4 RVs $\{\theta^{(1)},\theta^{(2)},\theta^{(3)},\theta^{(4)} \}\in \bR^4$.

The RC ansatz \eqref{eq: RC set up} has similar  nature as \eqref{eq: rigo set up}, so that 
similar rigorous proof on well-posedness is attainable under proper assumptions.   
By introducing more RVs, the RC ansatz provides more flexibility  for complicate NN architectures( see Section \ref{section: other nn structure}).

For other measure structures, we refer to \cite[Section 9]{nguyen2023rigorous} for a more detailed discussion.

\subsection{ the RC-ansatz for other NN structures}
\label{section: other nn structure}

In this section,  we first provide the following receipt to develop the RC-ansatz using the $\gamma$ notation and $\Gamma$ set: 

\begin{enumerate}
    \item Split any matrix type weights $W\in \bR^{D_1\times D_2}$ with $D_1,D_2>1$ into 2 corresponding RVs $\{\cR,\cC\}$, and viewed the $W$ is developed through some $\{\cR\}_{i=1}^{D_1}$, $\{\cC\}_{j=1}^{D_2}$ and $\phi:\bR^2\rightarrow \bR$. 
    \item The vector like weights $B \in \bR^D$ admits the form of  $\{B\}_{i=1}^{D}$. 
    \item Write down the calculation order using the $\gamma$ notation as in \eqref{eq: mlp nn function - 5-layer 1d}, then split the measure into different part with the $\Gamma$ set. 
\end{enumerate}

We  provide 3 examples: the MLP with bias example,  the Resnet example, and the attention block example to demonstrate the steps to develop proper measure structure, and we make some comments on the weight transfer algorithm.

\subsubsection{MLP with bias}
\label{section: mlp with bias}
We demonstrate a $L$-fully ($L\geq 3$) connected NN, with input $x\in \bR^{D_x}$ and output $f(x)\in \bR^{D_y}$, suppose the width of each layers are the same $N$ (which can be easily extended to non-equal cases), then, for $\ell \in \{1,\dots, L-2\}$, we have

\begin{align}
    U\in \bR^{N \times D_x},~B^U \in \bR^{N},  W^{(\ell)} \in \bR^{N\times N},~B^{(\ell)} \in \bR^{N},     ~ V \in \bR^{ D_y\times N},  ~B^V \in \bR^{D_y},
\end{align}
where for $W^{(\ell)}$, there should be 2 different RVs $\{\cR,\cC\}^{(\ell)}$ with each of $N$ samples to develop $W^{(\ell)}$. 
For $U$, there should be $D_x+1$ different RVs $\{ \cR^{emb}, \{\cC^{emb}_j\}_{j=1}^{D_x}\}$, with $N$ samples of each $\cC^{emb}_j$ RV and   $\cR^{emb}$  to develop   $U$. 
For $V$, there are $D_y+1$ different RVs $\{ \{\cR^{op}_i\}_{i=1}^{D_y},\cC^{op}\}$, with $N$ samples of each $\cR^{op}_i$ RV and   $\cC^{op}$  to develop   $U$. 
The $U,V$ are different from $W^{(\ell)}$ due to data pair $\{x,y\}$, where they are not necessarily identically distributed vectors. So that we have

\begin{align}
    \notag 
  & \Big\{  B^V_{ \gamma_{L+1}},V^L_{\gamma_{L+1},\gamma_{L}},
  B^{(L-2)}_{ \gamma_L},
  W^{(L-2)}_{\gamma_{L},\gamma_{L-1}},
  \dots,
  B^{(\ell)}_{ \gamma_3}
  ,W^{(1)}_{\gamma_3,\gamma_2}    ,
  B^U_{\gamma_2}, U_{\gamma_2,\gamma_1}\Big\}_\gamma 
  =
    \Big\{  B^V_{ \gamma_{L+1}},
    \phi^{(L)}(\cR_{\gamma_{L+1}}^{op},\cC_{\gamma_{L}}^{op}),
    \\
    \label{eq: mlp with bias gamma things}
    & \quad 
  B^{(L-2)}_{ \gamma_L},
  \phi^{(L-1)}(\cR_{\gamma_{L}}^{(L-2)},\cC_{\gamma_{L-1}}^{(L-2)}),
  \dots,
  B^{(\ell)}_{ \gamma_3}, 
   \phi^{(2)}(\cR_{\gamma_{3}}^{(1)},\cC_{\gamma_{2}}^{(1)})
  ,
  B^U_{\gamma_2}, 
  \phi^{(1)}(\cR_{\gamma_{2}}^{emb},\cC_{\gamma_{1}}^{emb})
  \Big\}_\gamma, 
\end{align}
so that the whole RC-ansatz measure structure $\mu \in \cP(\bR^{3L+D_x+D_y-4})$ such that 

\begin{align}
    \notag 
    &\Big\{  B^V_{ \gamma_{L+1}},
    \phi^{(L)}(\cR_{\gamma_{L+1}}^{op},\cC_{\gamma_{L}}^{op}), B^{(L-2)}_{ \gamma_L},
   \cR_{\gamma_{L}}^{(L-2)},\cC_{\gamma_{L-1}}^{(L-2)} ,
  \dots,
  B^{(\ell)}_{ \gamma_3}, 
    \cR_{\gamma_{3}}^{(1)},\cC_{\gamma_{2}}^{(1)}
  ,
  B^U_{\gamma_2}, 
  \cR_{\gamma_{2}}^{emb},\cC_{\gamma_{1}}^{emb}
  \Big\} \sim \mu,
  \\
    \label{eq: measure structure for mlp with bias }
  &\qquad \mu= \mu^{B^V,\cR^{op}}
  \times 
  \mu^{ \cC^{op},B^{L-2},\cR^{L-2}}
  \times   \dots 
  \times   
  \mu^{ \cC^{1},B^{U},\cR^{emb}}
  \times   \mu^{\cC^{emb} },
\end{align}
the measure structure in \eqref{eq: measure structure for mlp with bias } is developed via the corresponding $\Gamma$ set of \eqref{eq: mlp with bias gamma things}, the key idea here is that the RVs are connected through the same $\gamma$. Notice that for $D_x,D_y=1$, the structure above can be simpler by not including the $ \cC^{emb}$ and $ \cR^{op}$.

\subsubsection{Skip connection}
\label{section: skip coonec}

The skip connection architecture is a usual operation in modern NN architecture, the idea to develop the RC-ansatz for the skip connection dynamic is similar to \ref{section: mlp with bias}, we provide the following example to demonstrate the key idea  

\begin{example}
    \label{example: skip connection}
    Consider the output $h_0 \in \bR^N$ of the layer before the skip connection, it then goes through 
    \begin{enumerate}
        \item $h_1 := \psi( W^{(1)} h_0 + B^{(1)})$ for $W^{(1)}\in \bR^{N\times N}, B^{(1)}\in \bR^N$, activation function $\psi:\bR^{N}\rightarrow \bR^N$;
        \item $h_2 := \psi( W^{(2)} h_0 + B^{(2)})$ for $W^{(2)}\in \bR^{N\times N}, B^{(2)}\in \bR^N$, activation function $\psi:\bR^{N}\rightarrow \bR^N$;
        \item $h_3:= F(h_2)$ for some function $F:\bR^N \rightarrow \bR^N$; 
        \item $h_4:= \psi( W^{(3)} h_3 + B^{(3)})$ for $W^{(3)}\in \bR^{N\times N}, B^{(3)}\in \bR^N$, activation function $\psi:\bR^{N}\rightarrow \bR^N$; 
        \item $h_5:= \psi\big( W^{(4)} (h_1+h_4) + B^{(4)} \big)$ for $W^{(4)}\in \bR^{N\times N}, B^{(4)}\in \bR^N$, activation function $\psi:\bR^{N}\rightarrow \bR^N$.
    \end{enumerate}
    From  the calculation above, we have the following weights with $\gamma$ notation in the computation as Example \ref{example: gamma set}
    \begin{align}
      \label{eq: skip connection gamma}
      \Big\{  B^{(4)}_{ \gamma_6}&
  ,W^{(4)}_{\gamma_6,\gamma_5},
  B^{(3)}_{ \gamma_5}
  ,W^{(3)}_{\gamma_5,\gamma_4},
  B^{(2)}_{\gamma_3},W^{(2)}_{\gamma_3,\gamma_2}
  ,
  B^{(1)}_{\gamma_2},W^{(1)}_{\gamma_2,\gamma_1}   
  ,
  B^{(1)}_{\gamma_5},W^{(1)}_{\gamma_5,\gamma_1}  
  \Big\}_\gamma,  
\end{align}
with 
$\gamma=\{\gamma_1,\dots,\gamma_5\}$ with $\Gamma=\big\{ \{\gamma_1\},   \{\gamma_2,\gamma_5\}, \{\gamma_3\}, \{\gamma_4\},  \big\}$, 
which leads to  a measure component  $   \mu^{ \cC^{(4)}, B^{(3)},\cR^{(3)}, \cC^{(2)},B^{(1)},\cR^{(1)}} \in \cP(\bR^6) $.    

\end{example}

Note that in the Example \ref{example: skip connection}, the set $\{\gamma_2,\gamma_5\} \in \Gamma$ comes from the step 5. in $W^{(4)} (h_1+h_4) + B^{(4)}$, since $h_1$ is connected with $W^{(1)}$ and $B^{(1)}$, $h_4$ is connected with $W^{(3)}$ and $B^{(3)}$, while $W^{(1)}$ is also connected with $W^{(2)}$. By further split the matrix weights $W$ into corresponding $\{\cR,\cC\}$, we can specifically shows that $ \cC^{(4)}$ is connected with  $\cR^{(1)},B^{(1)},\cR^{(3)},B^{(3)}$, and $\cR^{(1)}$ is also connected with $\cC^{(2)}$, we thus build the corresponding measure $\mu^{ \cC^{(4)}, B^{(3)},\cR^{(3)}, \cC^{(2)},B^{(1)},\cR^{(1)}} \in \cP(\bR^6) $. 
 
If we apply the settings in \cite{nguyen2023rigorous}, where the $\mu^{ \cC^{(4)}, B^{(3)},\cR^{(3)}, \cC^{(2)},B^{(1)},\cR^{(1)}} \in \cP(\bR^6) $ is replaced by   $\mu^{ \theta^{(2)}, B^{(3)}, B^{(1)}, \theta^{(1)}} \in \cP(\bR^4) $ or  $\mu^{   B^{(3)},   \theta^{(1)},B^{(1)}} \in \cP(\bR^3) $ which may lead to trouble due to less degree of freedom. The RC ansatz thus provides a higher flexibility.

Indeed the measure structure is complicate for more general cases, the correlation results in Figure \ref{fig:4: LLM correlation} admits the analysis, however,  following the receipt to   write down the connection between different weights carefully,   we can develop corresponding RC-ansatz base on $\gamma$ notation and $\Gamma$ set.

\subsubsection{Attention block}
\label{section: attention}

The attention block is one of the most important NN architecture nowadays, we think it would be good to give some analysis. We provide the following example

\begin{example}
    \label{example: attention}
    Consider the output $h_0 \in \bR^{N\times D_x}$ of the layer before the attention block,  given an activation function $\psi:\bR^{N\times (D_x+1)}\rightarrow \bR^{N\times D_x}$, it  goes through 
    \begin{enumerate}
        \item $X := \psi( W^{(1)} h_0 , B^{(1)})$ for $W^{(1)}\in \bR^{N\times N}, B^{(1)}\in \bR^N$ ;
        \item We compute the following attention block, and use $@$ to denote the matrix multiplication      
        \begin{align*} 
            & \quad  h_1 = V @ F\big(  K.T @ Q \big)   
            \quad  \textit{with} \quad  
            \begin{cases}
                Q : =  W^{Q}  X  ,\quad   \bR^{N \times D_x},
                \\
                K : = W^{K} X ,\quad   \bR^{N \times D_x},  
                \\
                V : = W^{V} X  ,\quad   \bR^{N   \times D_x}, 
            \end{cases}
            \\  
            & \textit{where} \quad  K.T @ Q: ~ \bR^{D_x \times D_x},  \quad 
            F(  K.T @ Q): ~ \bR^{D_x \times D_x},  \quad 
            V @ F(  K.T @ Q) : ~ \bR^{N\times D_x}, 
        \end{align*}
        for $ W^{Q},W^{K},W^{V} \in \bR^{N \times N}$ and some function $F:\bR^{D_x \times D_x } \rightarrow \bR^{D_x \times D_x }$. 

        \item  
        $h_2 := \psi( W^{(2)} h_1 , B^{(2)})$ for $W^{(2)}\in \bR^{N\times N}, B^{(2)}\in \bR^N$;
    \end{enumerate}
    
    From  the calculation above, we have the following weights with $\gamma$ notation in the computation 
    \begin{align}
  \label{eq: attention gamma}
      \Big\{  B^{(2)}_{ \gamma_7}&
  ,W^{(2)}_{\gamma_7,\gamma_6},
  W^{V}_{ \gamma_6,\gamma_4}
  ,W^{K}_{\gamma_5,\gamma_3},
   W^{Q}_{\gamma_5,\gamma_2}
  ,
  B^{(1)}_{\gamma_2},W^{(1)}_{\gamma_2,\gamma_1}    
  ,
  B^{(1)}_{\gamma_3},W^{(1)}_{\gamma_3,\gamma_1} 
  ,
  B^{(1)}_{\gamma_4},W^{(1)}_{\gamma_4,\gamma_1} 
  \Big\}_\gamma  
\end{align}
with 
$\gamma=\{\gamma_1,\dots,\gamma_7\}$ with $\Gamma=\big\{ \{\gamma_1\},   \{\gamma_2,\gamma_3,\gamma_4\}, \{\gamma_5\}, \{\gamma_6\}, \{\gamma_7\} \big\}$, 
where  leads to  a  measure component   $   \mu^{  \cC^{V},\cC^{K}, \cC^{Q},B^{(1)},\cR^{(1)}} \in \cP(\bR^5) $. 

\end{example}

Notice that to make the notation clearer, we do not present the bias term in step$(b.)$, one can reach similar dynamics by adding the bias terms. The measure component $   \mu^{  \cC^{V},\cC^{K}, \cC^{Q},B^{(1)},\cR^{(1)}} \in \cP(\bR^5) $  connected to $\{\gamma_2,\gamma_3,\gamma_4\}$ is derived similar to Example \ref{example: skip connection}, where the output $X$ of step 1.   is connected to $W^{Q},W^{K},W^{V}$.   

Once again, based on the Example \ref{example: attention}, one can extend to more general attention block cases using similar dynamics.

\subsection{the RC initialization}
\label{sec: rc ini}

Choosing a proper initialization is important to NN,  for example, the Neural Tangent Kernel approach   introduced in \cite{2018ntkorigin} leads to kernel behaviour in the training of NN, the nice initialization $\mu P$  in \cite{tp4} which leads to nice hyper-parameter transfer application in \cite{tp5}. For the MFNN, \cite[Section 4.2.1]{nguyen2019mfexperiment} provides non-zero mean type normal distribution as initialization.

\begin{algorithm*}[htb!]
	\SetAlgoLined
	\KwIn{ 
        The weight  sets  $\boldsymbol\Theta:=\{W^{(1)},W^{(2)},\dots,W^{(n)}\}$ we need to initialized and will be trained, where the bias terms and other 1-dimension vector type weights are also included in $\boldsymbol\Theta$ and they are set to be a 2-dimension matrix where the added dimension is set  to 1.  
        The measure sets $\boldsymbol\mu:=\{ \mu^{(1)},\dots,\mu^{(2n)} \}$, where each measure $\mu^{(i)} \in \cP(\bR)$ for $i\in\{1,\dots,2n\}$, notice that  $\mu^{(i)}$ can be different. 
        The function sets  $\boldsymbol\Phi:=\{ \phi^{(1)},\dots,\phi^{(n)} \}$, where each function $\phi^{(i)}:\bR^2\rightarrow \bR$ for $i\in\{1,\dots,n\}$, notice that  $\phi^{(i)}$ can be different. }
        $z \gets 1 $  \; 
        \For{$i \gets 1$ to $n$}{
		\eIf{ both the dimensions of $W^{(i)}$ are proportional  to $N$ }{
                get the dimensions $d^{1,i},d^{2,i}$ of $W^{(i)}$ such that $W^{(i)}\in \bR^{d^{1,i}\times d^{2,i}}$      \;
			sample each element of $\cR^{(i)} \in \bR^{d^{1,i}}$ independently from   $\mu^{(z)} \in \cP(\bR)$\;  
                $z \gets z + 1 $  \;
                sample each element of $\cC^{(i)} \in \bR^{d^{2,i}}$ independently from    $\mu^{(z)} \in \cP(\bR)$\;
                $z \gets z + 1 $  \;
                \For{$ j \gets 1$ to $d^{1,i}$}{
                    \For{$ k \gets 1$ to $d^{2,i}$}{
                        $ W^{(i)}_{j,k}\gets \phi^{(i)}(  \cR^{(i)}_j, \cC^{(i)}_k        )$   \;
                }
                }
		}{
            sample each element of $W^{(i)} $  independently from  $\mu^{(z)} \in \cP(\bR)$\; 
            $z \gets z + 1 $  \;
            }
	}
        \KwOut{The initialized weight  sets $\boldsymbol \Theta_0:=\{W^{(1)},W^{(2)},\dots,W^{(n)}\}_0$. }   
	\caption{RC-initialization }
	\label{alg: RC ini}
\end{algorithm*}

As discussed in \cite[Appendix E]{tp4}, the zero mean independent initialization leads to gradient vanish when there are more than 2 hidden layers, e.g, one can check the average-like term nested in \eqref{eq: delta w3 eg eq} goes to 0 for i.i.d $ w^{(1)} _{{\gamma_2},{\gamma_1}} \sim \cN(0,1)$. So that for training the MFNN, we suggest to take an none-zero mean type initialization for i.i.d setting or take a  dependent initialization  type such as \cite[Section 7]{nguyen2023rigorous} and the following "RC-initialization" introduced in Algorithm \ref{alg: RC ini}.

The RC-initialization is a connected type initialization which prevent the  stuck at initialization. This is done by taking proper choices of function sets $\boldsymbol\Phi$ (say the simple add function $\phi(x,y)=x+y$ for $x,y\in \bR$),   with the corresponding RVs $\{\cR,\cC\}$.

From   the simplified 5-layer MFNN analysis in Section \ref{section: anstazr for 5 layers}, one important findings is that the updates of the weights \eqref{eq: delta  eg eq} are connected in row and column sense, so that after training for sufficiently, we expect the   weights are dependent in a sense of rows follow by columns. So that taking RC initialisation shall be very nature and the RC ansatz provide perfect description,  it seems that taking a fully i.i.d initialisation may introduce chaos and slow down the convergence to the "nice" measure.  

However, in our experiments, we find that the RC-initialisation did not outperform and even worse than the i.i.d initialisation. The reason may comes from the fact that the degree of freedom using RC-initialisation is much smaller than the i.i.d type. For example, the $N\times N$ matrix has $2N$ degree of freedom using RC-initialisation but have $N^2$ degree of freedom using i.i.d initialisation. Although at the end of day, we see a strong row and column pattern in the MFNN (e.g. Figure \ref{fig:1: correlation}) and for a trained LLM, introducing higher degree of freedom seems to be helpful instead of adding too much chaos. Maybe there are other ways to properly balance the degree of freedom with chaos by properly change the NN (e.g. adding activation on weights before doing the matrix multiplication) or other proper theory to demonstrate the NN, we shall  leave these mysterious in future work.

\subsubsection{Discussion on the   the feature learning }
\label{section: discussion and conclusions}

The definition of feature learning is introduced in \cite[Definition 3.5]{tp4} which is important to many types of NN especially for LMM as discussed in \cite{tp4,tp6}, the idea of feature learning is connected with the initialisation and  update order of the weights with respect to the width $N$, and thus the output of each layers suffer $\cO(1)$ update. However, it is well-known that the NTK admits a $\cO(1/\sqrt{N})$ for each step update compare to the  the initialisation of $\cO(1)$, thus may lack of the feature learning ability but leads to a kernel behaviour \cite{tp4}. Notice that in the limitation $N\rightarrow \infty$, the distribution of the weights shall stay unchanged in this case, but there exist a significant change of the weights in practice especially for LMMs.     

We would like to address that the MFNN also admit feature learning instead  of kernel regime for a proper choices of the initialisation as discussed in Section \ref{sec: rc ini}. We refer to \cite[Section 9.5]{nguyen2023rigorous} for a more detailed discussion.

\subsection{Sampling methods}
\label{sec: sample methods}

In this section, we provide few sampling method for convince. With a given  sample sets are finite with $N$ identically distributed samples $\{\bx\}_{i=1}^N$ of which $\bx \sim \mu \in \cP(\bR^D)$, we need to sample $M$ different samples  $\{\by \}_{i=1}^M$ from it. \\

\textbf{  Random sampling }\\

We take $M$  different samples from the data set  $\{\bx\}_{i=1}^N$ one by one with randomly choices, each sample $\by_i$ for $i \in \{1,\dots,M\}$  are taken with probability $1/N$ from   $\{\bx\}_{i=1}^N$.  \\

\textbf{ Group sampling }\\

We split the whole data set $\{\bx\}_{i=1}^N$ into  $N_g$ different groups based on some criterion (for example by sorted values of $|\bx|$). The $\by_i$ for $i \in \{1,\dots,M\}$ are sample first with  the corresponding probability ($1/N_g$ if the each groups contains same amount of samples) of the $n^{th}-$ group for $n\in\{1,\dots,N_g\}$ and then  take random sample within the  $n^{th}$  group. \\

\textbf{  Function-Based sampling }\\

We first develop some test function $\phi_j: \cP({\bR^D}) \times\cP({\bR^D}) \rightarrow \bR $   for $j \in \{1,\dots, N_{f}\}$ with $N_{f}$ is the amount of test function.

We show the following  test function $\phi$ as examples:

\begin{align}
    \nonumber 
    \textit{Moments}: \phi(  \mu^{x,N}, \mu^{y,M}   )
    &:= 
     \Big|~ 
     \int_{\bR} |\bx|^p \mu^{x,N}(\dd \bx) - \int_{\bR} |\by|^p \mu^{y,N}(\dd \by)
     ~\Big|
    \\
    &=\Big|~ \frac{1}{N} \sum_{i=1}^N |\bx_i|^p - \frac{1}{M} \sum_{j=1}^M |\by_j|^p ~\Big|, \quad \textit{ for } p\geq 1,
    \\
    \nonumber
    \textit{Indicator}: \phi(  \mu^{x,N}, \mu^{y,M}   )
    &:= 
     \Big|~ 
     \int_{\bR}  \1\{|\bx_i|^p \geq q \} \mu^{x,N}(\dd \bx) - \int_{\bR}  \1\{|\by_i|^p \geq q \}\mu^{y,N}(\dd \by)
     ~\Big|
    \\
    &= \Big|~ \frac{1}{N} \sum_{i=1}^N \1 \{|\bx_i|^p \geq q \} - \frac{1}{M} \sum_{j=1}^M  \1\{|\by_j|^p \geq q \} ~\Big|, \quad \textit{ for } q\in \bR^+,
    \\
    \textit{where} \quad 
    \mu^{x,N} &:= \frac{1}{N} \sum_{i=1}^N \delta_{\bx_i},
    \quad 
    \mu^{y,N} := \frac{1}{M} \sum_{i=1}^M \delta_{\by_i}.
\end{align} 

We can then measure the differences between the empirical measure of the  data set $\mu^{x,N}$ and the the empirical measure of the  sample set $\mu^{y,M}$ via some weighted summation:

\begin{align}
    \cL_{mea}( \mu^{x,N}, \mu^{y,M}    ):= \sum_{j=1}^{N_{f}} w_{j}\phi_{j}(   \mu^{x,N}, \mu^{y,M}  ),
\end{align}
where $\{w\}_{j=1}^{N_{f}}$ are some positive constants.

The receipts of optimized sampling is now clear: 
\begin{enumerate}
    \item We samples $N_y$ different sets of $\big\{ \{\by^{i}_j\}_{j=1}^M\big\}_{i=1}^{N_y}$, so that we have  $N_y$ different $\mu^{i,y,M}$.
    \item we compute the loss function value between $\mu^{x,N}$ and $\mu^{i,y,M}$, and take the corresponding sample of :
    \begin{align}
        \mu^{\star,y,M} = \operatorname*{argmin}_{  \mu \in  \{ \mu^{i,y,M}  \}_{i=1}^{N_y}   }   \cL_{mea}( \mu^{x,N}, \mu     ). 
    \end{align}
\end{enumerate}
For example, we can choose different moments as test function and select the best sample $\mu^{\star,y,M} $  with minimal moment compare to $\mu^{x,N}$.

\end{document}